%% file: paper.tex
\documentclass[10pt,twocolumn]{article}

\usepackage{cvpr}
\usepackage{times}
\usepackage{graphicx}
\usepackage{amsmath}
\usepackage{amssymb}

\usepackage{color}
\usepackage[backend=bibtex,style=ieee,backref=true,natbib=true,url=false,doi=false]{biblatex}
\usepackage{siunitx}
\usepackage{pgfplotstable}
\usepackage{subcaption}
\usepackage{booktabs}
\usepackage{pdflscape}
\usepackage{afterpage}
\usepackage{tikz}

\usetikzlibrary{arrows,positioning,calc,tikzmark,decorations.pathreplacing}

\input{ACsettings}
\input{pgftablehighlight}
\usepgfplotslibrary{colorbrewer}
\usepackage[breaklinks=true,letterpaper=true,colorlinks,bookmarks=false]{hyperref}

\cvprfinalcopy 

\def\cvprPaperID{449} 

\begin{document}

\title{Deep Roots:\\
Improving CNN Efficiency with Hierarchical Filter Groups}

\author{Yani Ioannou$^1$ \and Duncan Robertson$^2$ \and Roberto Cipolla$^1$ \and Antonio Criminisi$^2$\\$^1$University of Cambridge, $^2$Microsoft Research}

\maketitle

\begin{abstract}
We propose a new method for creating computationally efficient and compact convolutional neural networks (CNNs) using a novel sparse connection structure that resembles a tree root. This allows a significant reduction in computational cost and number of parameters compared to state-of-the-art deep CNNs, without compromising accuracy, by exploiting the sparsity of inter-layer filter dependencies.
We validate our approach by using it to train more efficient variants of state-of-the-art CNN architectures, evaluated on the CIFAR10 and ILSVRC datasets. Our results show similar or higher accuracy than the baseline architectures with much less computation, as measured by CPU and GPU timings. For example, for ResNet 50, our model has 40\% fewer parameters, 45\% fewer floating point operations, and is 31\% (12\%) faster on a CPU (GPU). For the deeper ResNet 200 our model has 25\% fewer floating point operations and 44\% fewer parameters, while maintaining state-of-the-art accuracy. For GoogLeNet, our model has 7\% fewer parameters and is 21\% (16\%) faster on a CPU (GPU).
\end{abstract}

\section{Introduction}
This paper describes a new method for creating computationally efficient and compact convolutional neural networks (CNNs) using a novel sparse connection structure that resembles a tree root. This allows a significant reduction in computational cost and number of parameters compared to state-of-the-art deep CNNs without compromising accuracy.

It has been shown that a large proportion of the learned weights in deep networks are redundant~\citep{Denil2013predicting}, a property that has been widely exploited to make neural networks smaller and more computationally efficient~\citep{Szegedy2014going,Denton2014efficient}). It is unsurprising then that regularization is a critical part of training such networks using large datasets~\citep{Krizhevsky2012}. Without regularization deep networks are susceptible to over-fitting. Regularization may be achieved by weight decay or dropout~\cite{Hinton2012}. Furthermore, a carefully designed sparse network connection structure can also have a regularizing effect. Convolutional Neural Networks (CNNs)~\citep{Fuk80,Lecun1998} embody this idea, using a sparse convolutional connection structure to exploit the locality of natural image structure. In consequence, they are easier to train.

With few exceptions, state-of-the-art CNNs for image recognition are largely monolithic, with each filter operating on the feature maps of all filters on a previous layer. Interestingly, this is in stark contrast to what we understand of biological neural networks, where we see ``highly evolved arrangements of smaller, specialized networks which are interconnected in very specific ways''~\citep{minsky1988perceptrons}.

Recently, learning a low-rank basis for filters was found to improve generalization while reducing the computational complexity and model size of a CNN with only full rank filters~\citep{Ioannou2016}. However, this work addressed only the spatial extents of the convolutional filters (\ie $h$ and $w$ in Fig.~\ref{fig:normalconv}). In this work we will show that a similar idea can be applied to the channel extents -- \ie filter inter-connectivity -- by using \emph{filter groups}~\citep{Krizhevsky2012}. We show that simple alterations to state-of-the-art CNN architectures can drastically reduce computational cost and model size without compromising accuracy.

\section{Related Work}
\label{previouswork}
Most previous work on reducing the computational complexity of CNNs has focused on approximating convolutional filters in the spatial (as opposed to the channel) domain, either by using low-rank approximations~\citep{mamalet2012simplifying,journals/corr/JaderbergVZ14, journals/pami/SironiTRLF15, journals/corr/LebedevGROL14, Ioannou2016}, or Fourier transform based convolution~\cite{mathieu2013fast, rippel2015spectral}. More general methods have used reduced precision number representations~\cite{1502.02551v1} or compression of previously trained models~\cite{Chen2015,Kim2016}. Here we explore methods that reduce the computational impact of the large number of filter channels within state-of-the art networks. Specifically, we consider decreasing the number of incoming connections to nodes.

\begin{figure}[tb]
\begin{subfigure}[b]{\linewidth}
\centering
\includegraphics[width=0.95\linewidth, page=1]{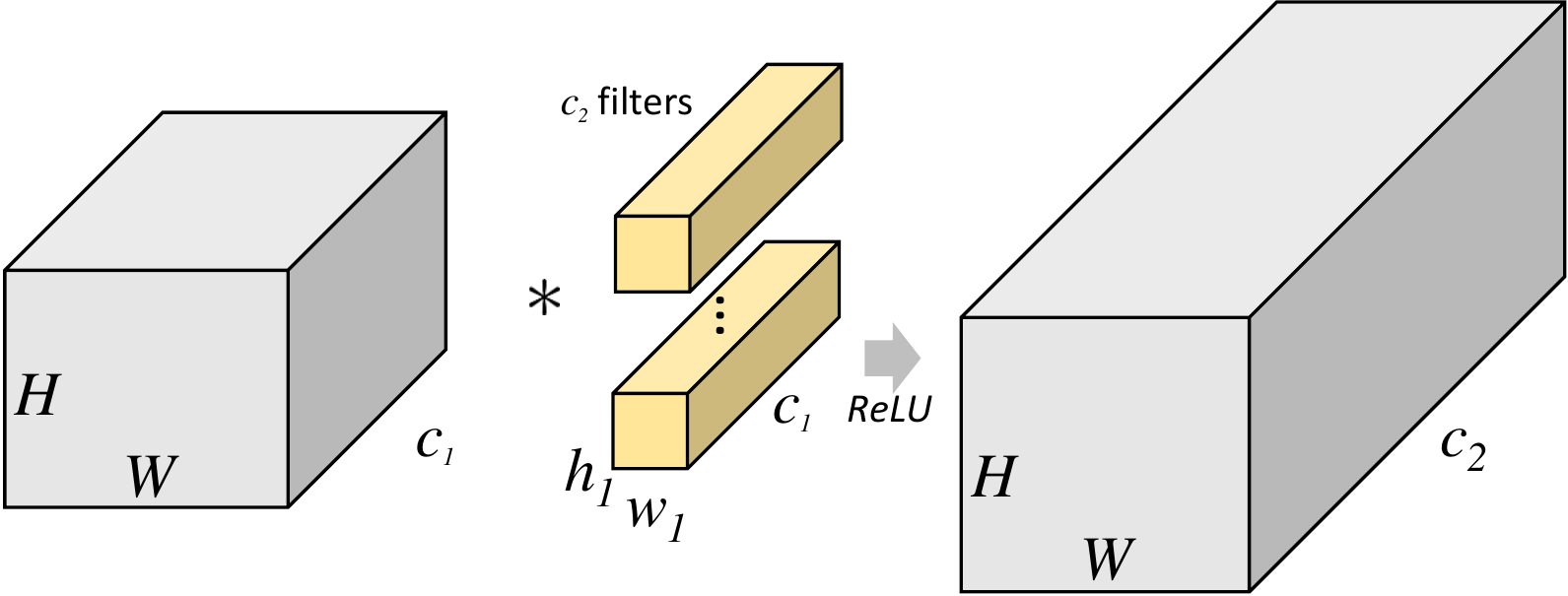}
    \caption{Convolution.}
   \label{fig:normalconv}
\end{subfigure}\\
\begin{subfigure}[b]{\linewidth}
\centering
\includegraphics[width=0.95\linewidth, page=2]{figs/groupfig}
   \caption{Convolution with filter groups.}
   \label{fig:groupedconv}
\end{subfigure}
\caption{\textbf{Filter Groups.} (a) Convolutional filters (yellow) typically have the same channel dimension $c_1$ as the input feature maps (gray) on which they operate. However, (b) with filter grouping, $g$ independent groups of $c_2/g$ filters operate on a fraction $c_1/g$ of the input feature map channels, reducing filter dimensions from $h$$\times$$w$$\times$$c_1$ to $h$$\times$$w$$\times$$c_1/g$. This change does not affect the dimensions of the input and output feature maps but significantly reduces computational complexity and the number of model parameters.}
\label{fig:groupconfig}
\end{figure}

\input{alexnetmaplot.tex}
\paragraph{AlexNet Filter Groups.} Amongst the seminal contributions made by \citet{Krizhevsky2012}~is the use of `filter groups' in the convolutional layers of a CNN (see Fig. \ref{fig:groupconfig}). While their use of filter groups was necessitated by the practical need to sub-divide the work of training a large network across multiple GPUs, the side effects are somewhat surprising. Specifically, the authors observe that independent filter groups learn a separation of responsibility (colour features vs. texture features) that is consistent over different random initializations. Also surprising, and not explicitly stated in~\citep{Krizhevsky2012}, is the fact that the AlexNet network has approximately 57\% fewer connection weights than the corresponding network without filter groups. This is due to the reduction in the input channel dimension of the grouped convolution filters (see Fig.~\ref{fig:alexnetplots}).
Despite the large difference in the number of parameters between the models, both achieve comparable accuracy on ILSVRC -- in fact the smaller grouped network gets $\approx1$\% lower top-5 validation error. This paper builds upon these findings and extends them to state-of-the-art networks.

\paragraph{Low-dimensional Embeddings.}
\citet{Lin2013NiN} proposed a method to reduce the dimensionality of convolutional feature maps. 
By using relatively cheap `1$\times$1' convolutional layers (i.e. layers comprising $d$ filters of size $1\times 1 \times c$, where $d<c$), they learn to map feature maps into lower-dimensional spaces, \ie to new feature maps with fewer channels. Subsequent spatial filters operating on this lower dimensional input space require significantly less computation. This method is used in most state of the art networks for image classification to reduce computation~\citep{Szegedy2014going,He2015}. Our method is complementary.

\paragraph{GoogLeNet.} In contrast to much other work, \citet{Szegedy2014going} propose a CNN architecture that is highly optimized for computational efficiency. GoogLeNet uses, as a basic building block, a mixture of low-dimensional embeddings~\citep{Lin2013NiN} and heterogeneously sized spatial filters -- collectively an `inception' module. 
There are two distinct forms of convolutional layers in the inception module, low-dimensional embeddings (1$\times$1) and spatial (3$\times$3, 5$\times $5). GoogLeNet keeps large, expensive spatial convolutions (\ie 5$\times$5) to a minimum by using few of these filters, using more 3$\times$3 convolutions, and even more 1$\times$1 filters. The motivation is that most of the convolutional filters respond to localized patterns in a small receptive field, with few requiring a larger receptive field. The number of filters in each successive inception module increases slowly with decreasing feature map size, in order to maintain computational performance. GoogLeNet is by far the most efficient state-of-the-art network for ILSVRC, achieving near state-of-the-art accuracy with the lowest computation/model size. However, we will show that even such an efficient and optimized network architecture benefits from our method.

\paragraph{Low-Rank Approximations.}
Various authors have suggested approximating learned convolutional filters using tensor decomposition~\citep{journals/corr/JaderbergVZ14,journals/corr/LebedevGROL14,Kim2016}. For example, \citet{journals/corr/JaderbergVZ14} propose approximating the convolutional filters in a trained network with representations that are low-rank both in the spatial and the channel domains. This approach significantly decreases computational complexity, albeit at the expense of a small amount of accuracy. In this paper we are not approximating an existing model's weights but creating a new network architecture with explicit structural sparsity, which is then trained from scratch.

\paragraph{Learning a Basis for Filters}
\begin{figure}[tb]
\centering
\includegraphics[width=\linewidth, page=3]{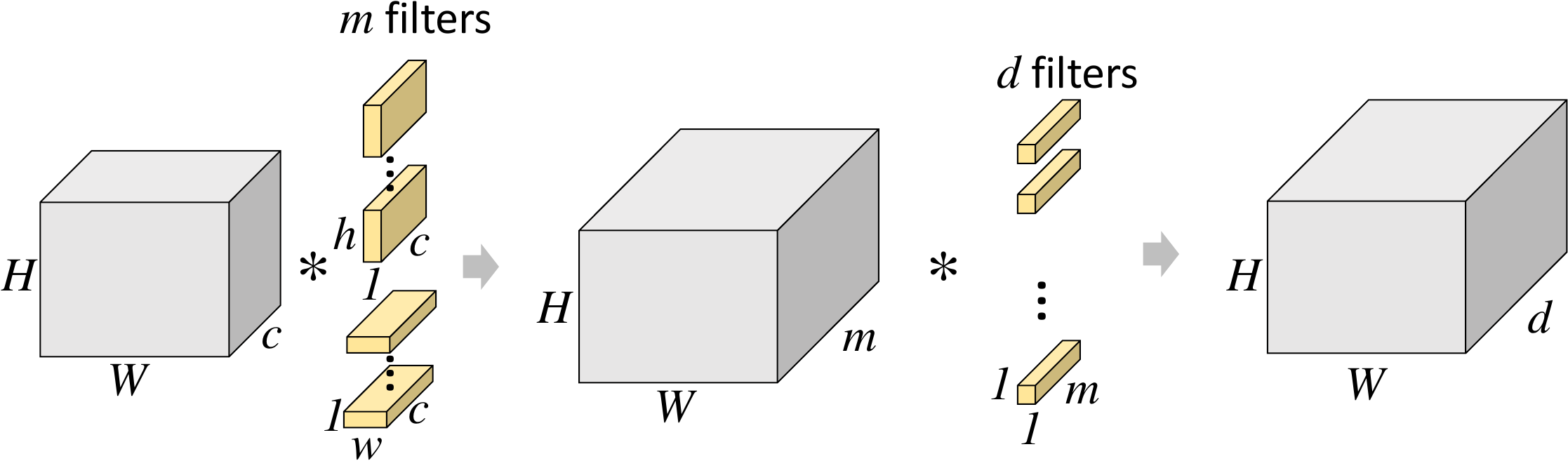}
\caption{\textbf{Learning a (Spatial) Basis for Filters.} Learning a linear combination of mostly small, heterogeneously sized spatial filters~\citep{Ioannou2016}. Note that all filters operate on all $c$ channels of the input feature map.}
\label{fig:spatialbasis}
\end{figure}
Our approach is connected with that of \citet{Ioannou2016} who showed that replacing 3$\times$3$\times$c filters with linear combinations of filters with smaller spatial extent (\eg 1$\times$3$\times c$, 3$\times$1$\times c$ filters, see Fig.~\ref{fig:spatialbasis}) could reduce the model size and computational complexity of state-of-the-art CNNs, while maintaining or even increasing accuracy. However, that work did not address the channel extent of the filters.

\section{Root Architectures}
\label{method}
In this section we present the main contribution of our work: the use of novel sparsely connected architectures resembling tree roots -- to decrease computational complexity and model size compared to state-of-the-art deep networks for image recognition.

\paragraph{Learning a Basis for Filter Dependencies}
It is unlikely that every filter (or neuron) in a deep neural network needs to depend on the output of all the filters in the previous layer. In fact, reducing filter co-dependence in deep networks has been shown to benefit generalization. For example, \citet{Hinton2012} introduced {\em dropout} for regularization of deep networks. When training a network layer with dropout, a random subset of neurons is excluded from both the forward and backward pass for each mini-batch.  Furthermore, \citet{Cogswell2016} observe a correlation between the covariance of hidden unit activations and overfitting. To explicitly reduce the covariance of hidden activations, they train networks with a loss function, based on the covariance matrix of the activations in a hidden layer. 

Instead of using a modified loss, regularization penalty, or randomized network connectivity during training to prevent co-adaption of features, we take a much more direct approach. We use filter groups (see Fig.~\ref{fig:groupconfig}) to force the network to learn filters with only limited dependence on previous layers. Each of the filters in the filter groups is smaller in the channel extent, since it operates on only a subset of the channels of the input feature map. 

\begin{figure}[tb]
\centering
\begin{subfigure}[b]{\linewidth}
\centering
\includegraphics[width=\linewidth, page=4]{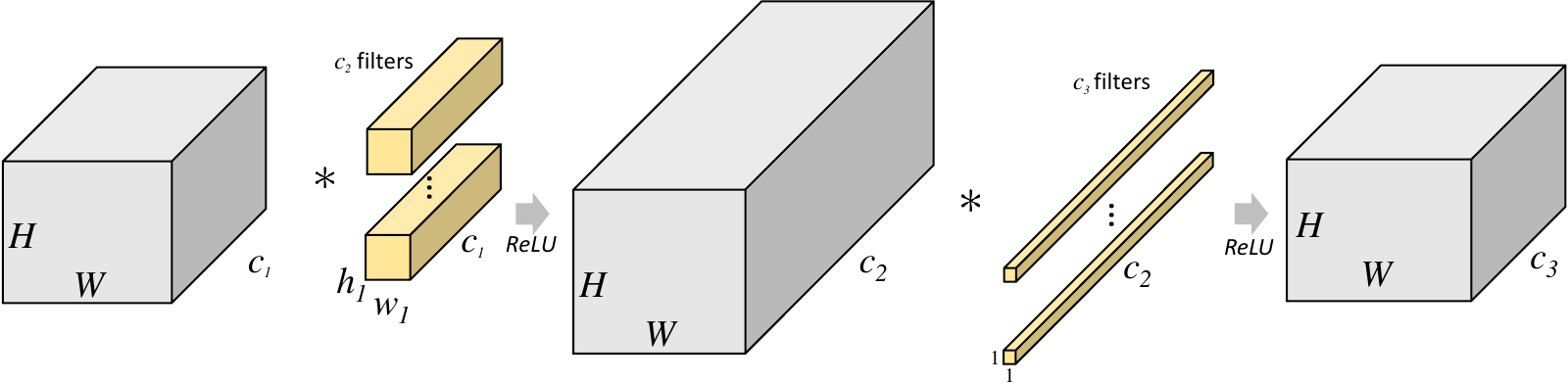}
   \caption{Convolution with $d$ filters of shape $h\times w\times c$.}
   \label{fig:normalresnet}
\end{subfigure}\\
\begin{subfigure}[b]{\linewidth}
\includegraphics[width=\linewidth, page=5]{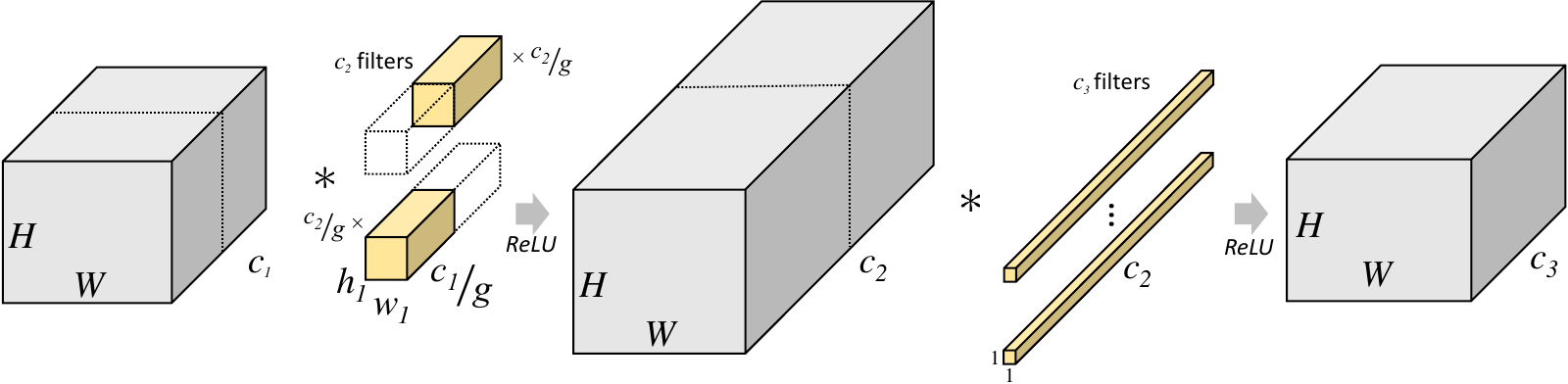}
   \caption{Root-2 Module: Convolution with $d$ filters in $g = 2$ filter groups, of shape $h\times w\times c/2$.}
   \label{fig:rootresnet2}
\end{subfigure}
\begin{subfigure}[b]{\linewidth}
\includegraphics[width=\linewidth, page=6]{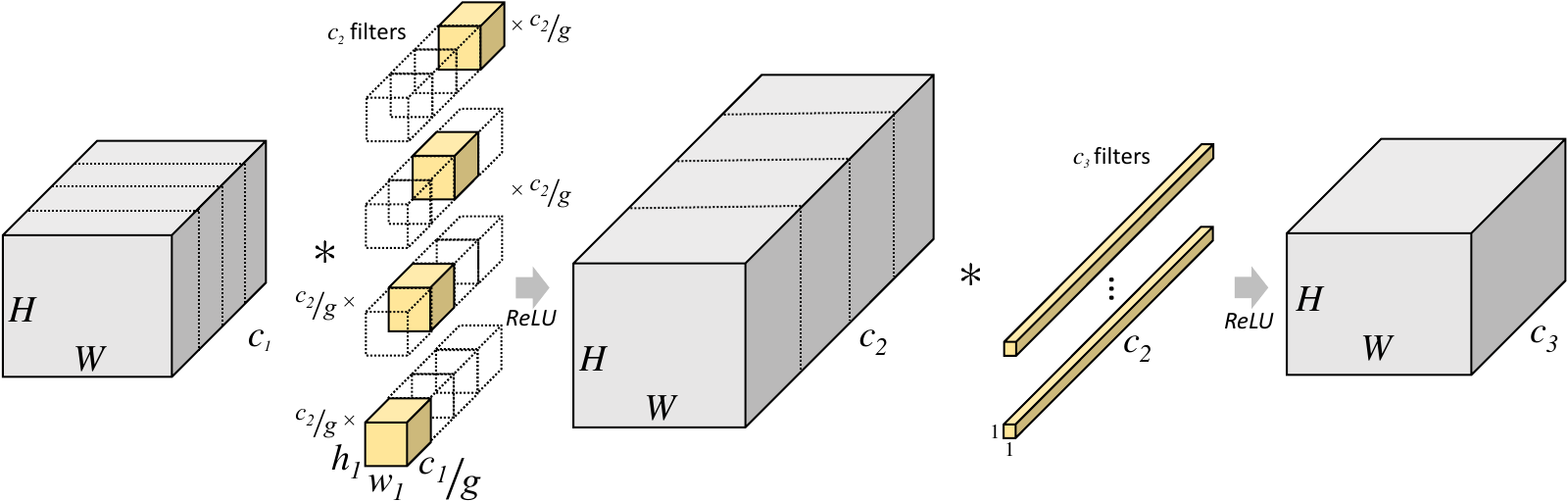}
   \caption{Root-4 Module: Convolution with $d$ filters in $g = 4$ filter groups, of shape $h\times w\times c/4$.}
   \label{fig:rootresnet4}
\end{subfigure}
\caption{\textbf{Root Modules.} Root modules (b), (c) compared to a typical set of convolutional layers (a) found in ResNet and other modern architectures. Grey blocks represent the feature maps over which a layer's filters operate, while colored blocks represent the filters of each layer. 
}
\label{fig:rootmodule}
\end{figure}
This reduced connectivity also reduces computational complexity and model size since the size of filters in filter groups are reduced drastically, as is evident in Fig.~\ref{fig:rootmodule}. Unlike methods for increasing the efficiency of deep networks by approximating pre-trained existing networks (see \S\ref{previouswork}), our models are trained from random initialization using stochastic gradient descent. This means that our method can also speed up training and, since we are not merely approximating an existing model's weights, the accuracy of the existing model is not an upper bound on accuracy of the modified model.

\paragraph{Root Module}
The basic element of our network architecture, a \emph{root module}, is shown in Fig.~\ref{fig:rootmodule}. A root module has a given number of filter groups, the more filter groups, the fewer the number of connections to the previous layer's outputs. Each spatial convolutional layer is followed by a low-dimensional embedding (1$\times$1 convolution). Like in \citep{Ioannou2016}, this configuration  learns a linear combination of the basis filters (filter groups), implicitly representing a filter of full channel depth, but with limited filter dependence.

\section{Results}
Here we present image classification results obtained by replacing spatial convolutional layers within existing state-of-the-art network architectures with root modules (described in \S\ref{method}) .

\subsection{Improving Network in Network on CIFAR-10}
Network in Network (NiN)~\cite{Lin2013NiN} is a near state-of-the-art network for CIFAR-10~\cite{CIFAR10}. It is composed of 3 spatial (5$\times$5, 3$\times$3) convolutional layers with a large number of filters (192), interspersed with pairs of low-dimensional embedding (1$\times$1) layers. As a baseline, we replicated the standard NiN network architecture as described by \citet{Lin2013NiN} but used state-of-the-art training methods. We trained using random 32$\times$32 cropped and mirrored images from 4-pixel zero-padded  mean-subtracted images, as in~\citep{goodfellow2013maxout, He2015}. We also used the initialization of \citet{He2015b} and batch normalization~\citep{Ioffe2015}. With this configuration, ZCA whitening was not required to reproduce validation accuracies obtained in~\citep{Lin2013NiN}. We also did not use dropout, having found it to have little effect, presumably due to our use of batch normalization, as suggested by~\citet{Ioffe2015}.

\begin{table}[tbp]
\caption[Network in Network Architectures]{\textbf{Network-in-Network}. Filter groups in each convolutional layer.}
\label{table:ninconfig}
\centering
\begin{tabular}{@{}lm{1.5em}m{1.5em}m{1.5em}m{1.5em}m{1.5em}m{1.5em}m{1.5em}m{1.5em}m{1.5em}@{}}
\toprule
    Model & \multicolumn{3}{c}{conv1} & \multicolumn{3}{c}{conv2} & \multicolumn{3}{c}{conv3} \\
     & \textit{\footnotesize a} & \textit{\footnotesize b} & \textit{\footnotesize c} & \textit{\footnotesize a} & \textit{\footnotesize b} & \textit{\footnotesize c} & \textit{\footnotesize a} & \textit{\footnotesize b} & \textit{\footnotesize c} \\
     & \textit{\footnotesize5$\times$5} & \textit{\footnotesize1$\times$1} & \textit{\footnotesize1$\times$1} & \textit{\footnotesize5$\times$5} & \textit{\footnotesize1$\times$1} & \textit{\footnotesize1$\times$1} & \textit{\footnotesize3$\times$3} & \textit{\footnotesize1$\times$1} & \textit{\footnotesize1$\times$1} \\
    Orig. & 1 & 1 & 1 & 1 & 1 & 1 & 1 & 1 & 1\\
    \midrule
    root-2 & 1 & 1 & 1 & 2 & 1 & 1 & 1 & 1 & 1\\
    root-4 & 1 & 1 & 1 & 4 & 1 & 1 & 2 & 1 & 1\\
    root-8 & 1 & 1 & 1 & 8 & 1 & 1 & 4 & 1 & 1\\
    root-16 & 1 & 1 & 1 & 16 & 1 & 1 & 8 & 1 & 1\\
    \bottomrule
\end{tabular}
\end{table}
\input{cifarninmatable}
To assess the efficacy of our method, we replaced the spatial convolutional layers of the original NiN network with root modules (as described in \S\ref{method}). We preserved the original number of filters per layer but subdivided them into groups as shown in Table~\ref{table:ninconfig}. We considered the first of the pair of existing 1$\times$1 layers to be part of our root modules. 
We did not group filters in the first convolutional layer -- since it operates on the three-channel image space, it is of limited computational impact compared to other layers.
\input{cifarninmaplot}
Results are shown in Table~\ref{table:nincifarresults} and Fig.~\ref{fig:nincifarplotsconvonly} for various network architectures\footnote{Here (and subsequently unless stated otherwise) timings are per image for a forward pass computed on a large batch. Networks were implemented using Caffe (with CuDNN and MKL) and run on an Nvidia Titan Z GPU and 2 10-core Intel Xeon E5-2680 v2 CPUs.}. Compared to the baseline architecture, the root variants achieve a significant reduction in computation and model size without a significant reduction in accuracy. For example, the root-8 architecture gives equivalent accuracy with only 46\% of the floating point operations (FLOPS), 33\% of the model parameters of the original network, and approximately 37\% and 23\% faster CPU and GPU timings (see \S\ref{gpuexplanation} for an explanation of the GPU timing disparity).

\newcommand{\covarlabels}[1]{%
\vspace{0.75em}
\begin{tikzpicture}[remember picture]
\node [anchor=south west, inner sep=0pt] (c)
    {
        #1
    };
    \path[use as bounding box] (c.south west) rectangle (c.north east);
    \node [anchor=south west, xshift=-0.5em, yshift=-0.5em, rotate=45] at (c.north west) {\footnotesize 0};
    \node [anchor=south east, xshift=\linewidth, yshift=-0.2em] at (c.north west) {\footnotesize 192};
    \node [anchor=south west, xshift=0.25em, yshift=-1.05\linewidth, rotate=90] at (c.north west) {\footnotesize 192};
    \node [anchor=south, xshift=0.5\linewidth] at (c.north west) {\footnotesize\texttt{conv3a}};
    \node [anchor=south, xshift=0.2em, yshift=-0.5\linewidth, rotate=90] at (c.north west) {\footnotesize \texttt{conv2c}};
\end{tikzpicture}%
}

\begin{figure}[tb]
\centering
\begin{subfigure}[b]{0.48\linewidth}
\centering
    \covarlabels{\includegraphics[width=\linewidth]{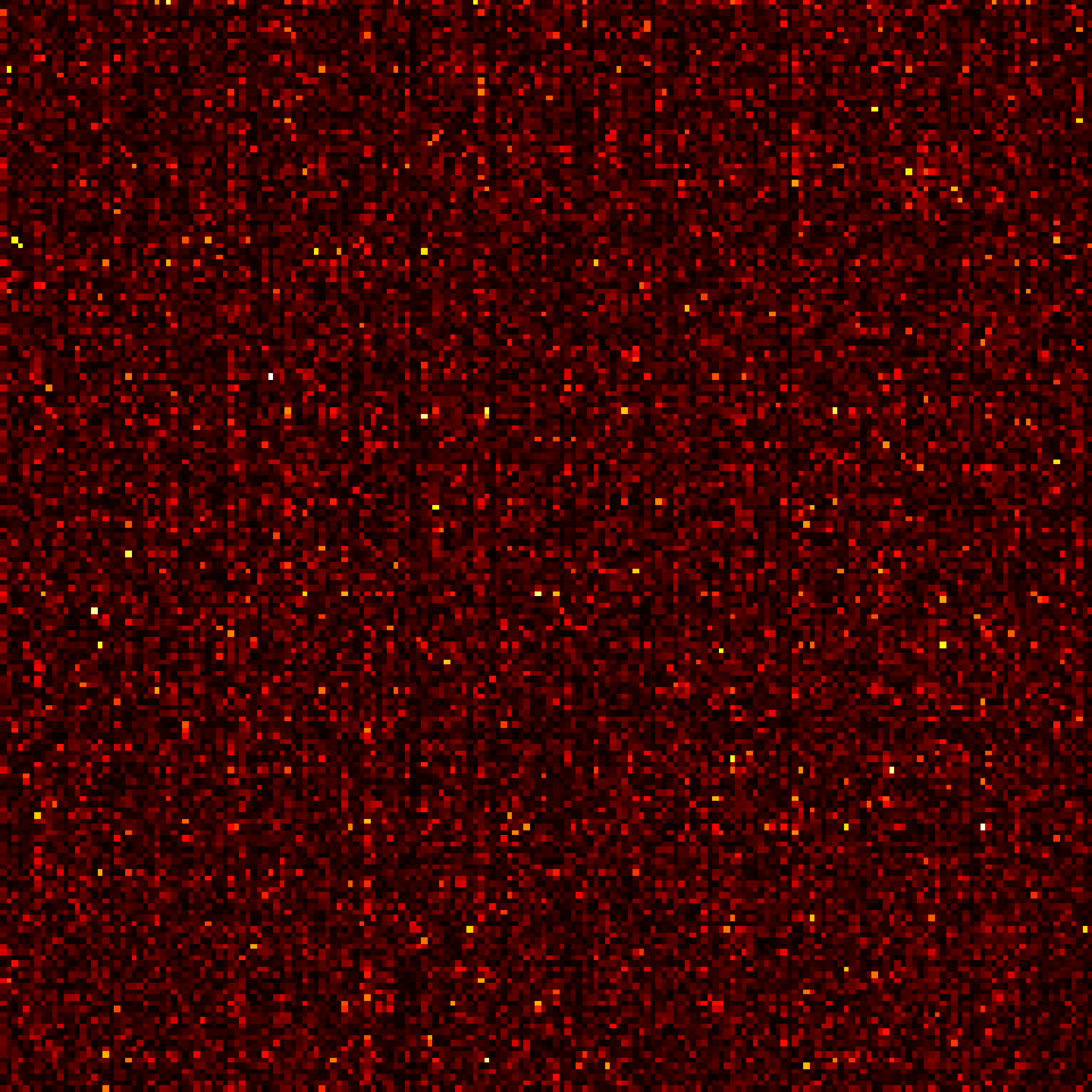}}
    \caption{\textbf{Standard}}
    \label{fig:normalcovartest}
\end{subfigure}
~
\begin{subfigure}[b]{0.48\linewidth}
\centering
    \covarlabels{\includegraphics[width=\linewidth]{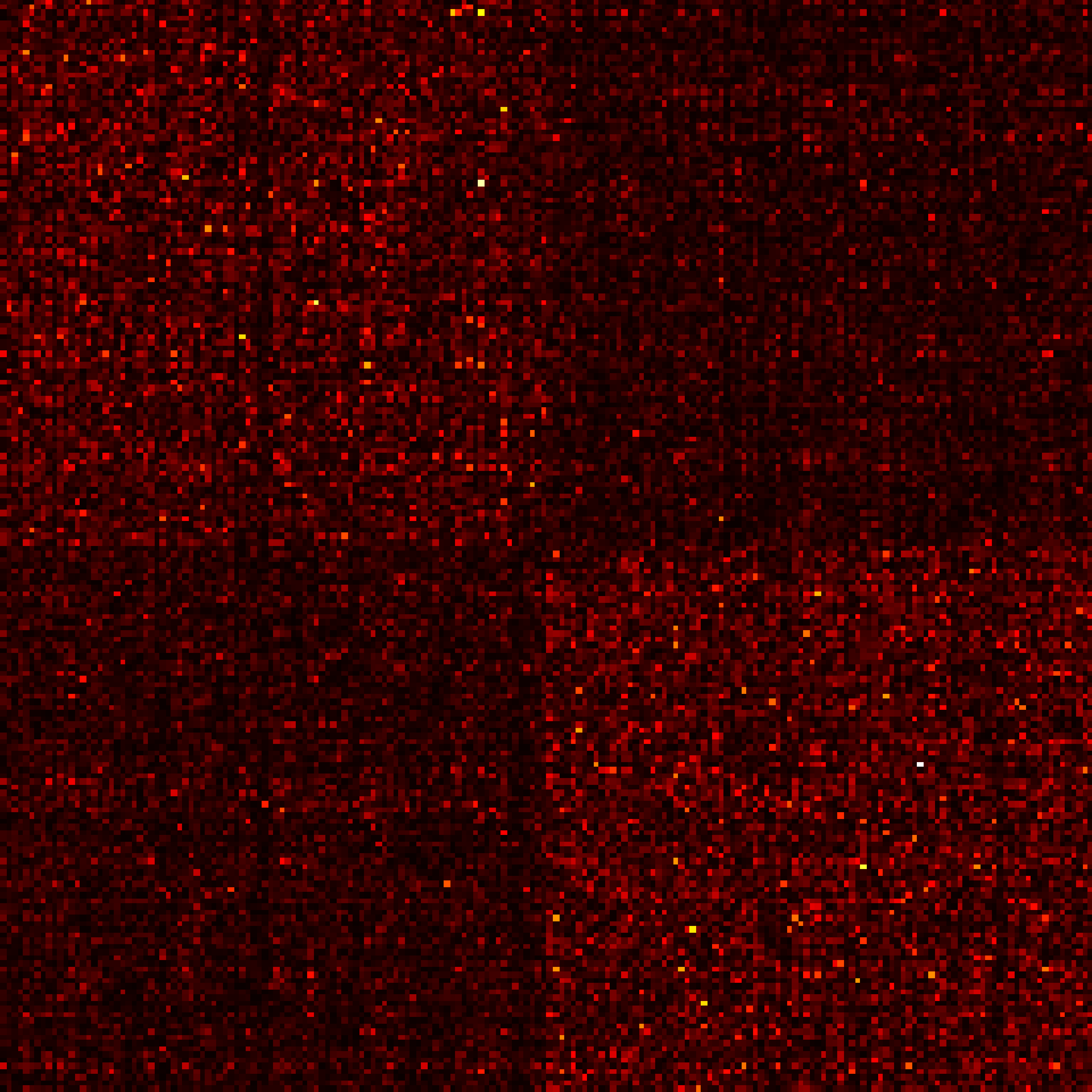}}
    \caption{\textbf{Root-4:} 2 filter groups}
    \label{fig:root4}
\end{subfigure}
~
\begin{subfigure}[b]{0.48\linewidth}
\centering
    \covarlabels{\includegraphics[width=\linewidth]{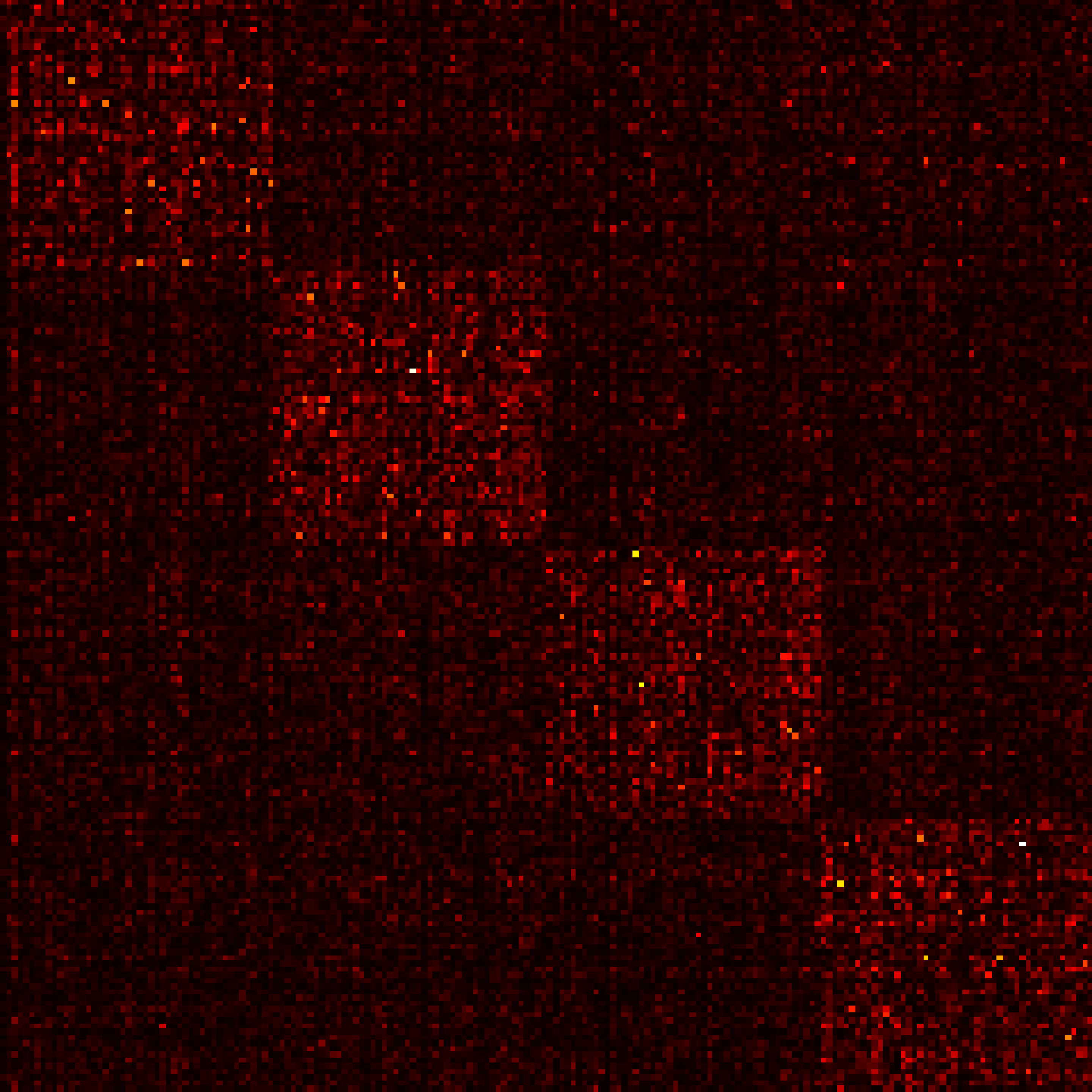}}
    \caption{\textbf{Root-8:} 4 filter groups}
    \label{fig:root8corr}
\end{subfigure}
~
\begin{subfigure}[b]{0.48\linewidth}
\centering
    \covarlabels{\includegraphics[width=\linewidth]{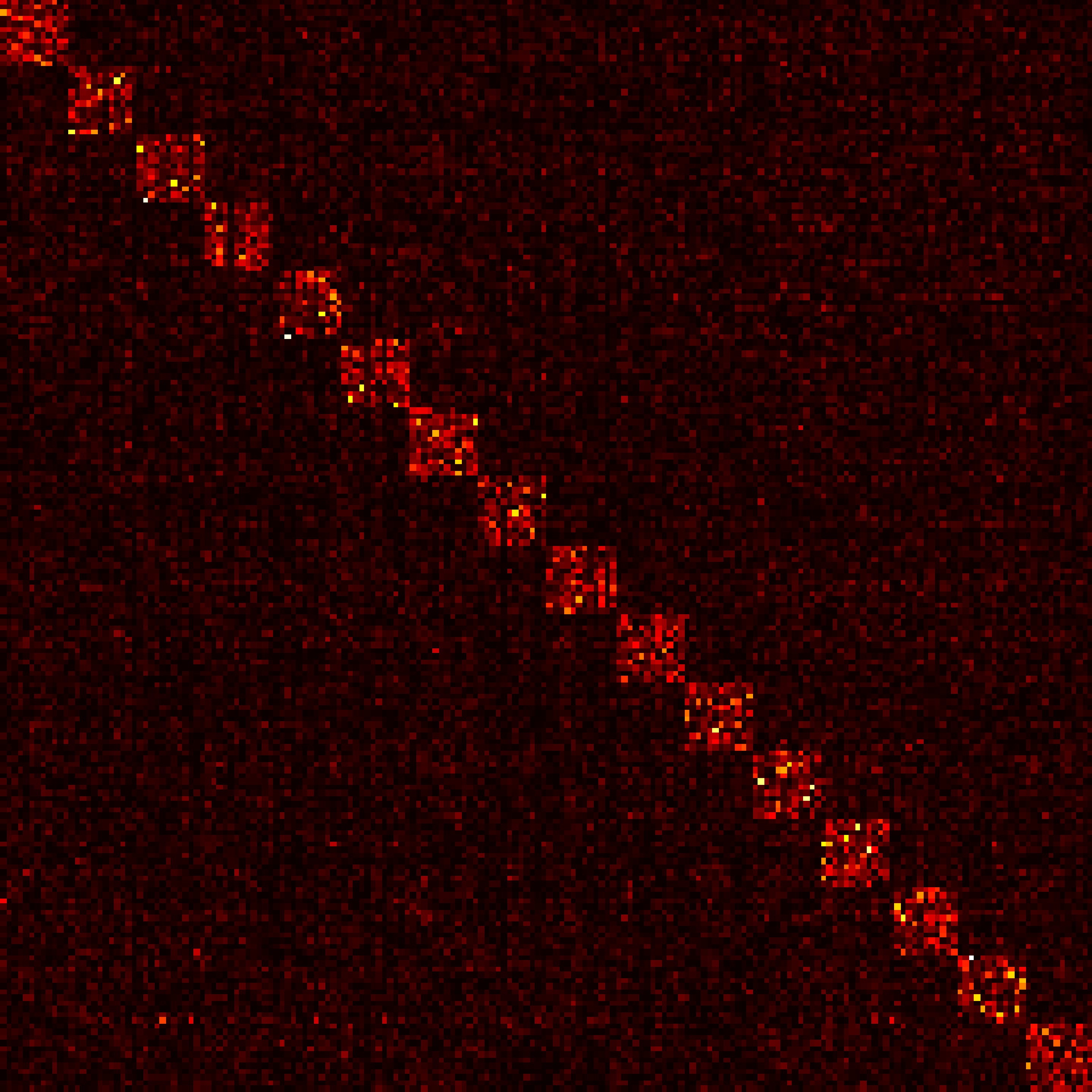}}
    \caption{\textbf{Root-32:} 16 filter groups}
    \label{fig:root32corr}
\end{subfigure}
\caption{\textbf{Inter-layer Filter Correlation.} The block-diagonal sparsity learned by a root-module is visible in the correlation of filters on layers \texttt{conv3a} and \texttt{conv2c} in the NiN network.}
\label{fig:covar}
\end{figure}
Figure~\ref{fig:covar} shows the inter-layer correlation between the adjacent filter layers \texttt{conv2c} and \texttt{conv3a} in the network architectures outlined in Table~\ref{table:ninconfig} as evaluated on the CIFAR test set. The block-diagonalization enforced by the filter group structure (as illustrated in Fig.~\ref{fig:groupconfig}) is visible, more so with larger number of filter groups. This shows that the network learns an organization of filters such that the sparsely distributed strong filter relations, visible in \ref{fig:normalcovartest} as brighter pixels, are grouped into a denser block-diagonal structure, leaving a visibly darker, low-correlated background. See \S\ref{interlayercovar} for more images, and an explanation of their derivation.

\subsection{Grouping Degree with Network Depth}
\begin{figure*}[tb]
\begin{subfigure}[c]{\linewidth}
\centering
\begin{tikzpicture}
    \begin{scope}[]
    \matrix[column sep=0em]{
    \node (1a) {
        \raisebox{-0.5\height}{\includegraphics[height=0.08\linewidth, page=15]{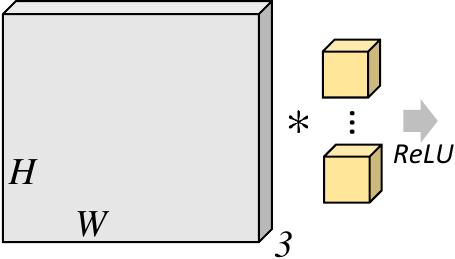}}
    };&
    \node (1b) {
        \raisebox{-0.5\height}{\includegraphics[height=0.11\linewidth, page=17]{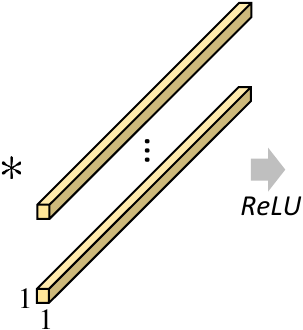}}
    };&
    \node (1c) {
        \raisebox{-0.5\height}{\includegraphics[height=0.11\linewidth, page=17]{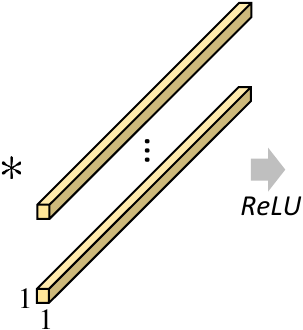}}
    };&
    \node (2a) {
        \raisebox{-0.5\height}{\includegraphics[height=0.09\linewidth, page=16]{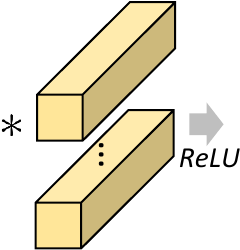}}
    };&
    \node (2b) {
        \raisebox{-0.5\height}{\includegraphics[height=0.11\linewidth, page=17]{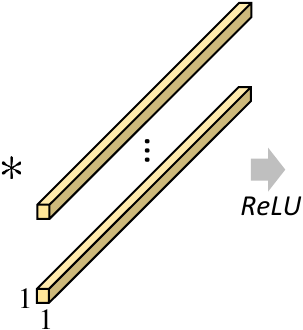}}
    };&
    \node (2c) {
        \raisebox{-0.5\height}{\includegraphics[height=0.11\linewidth, page=17]{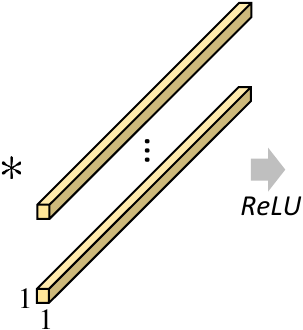}}
    };&
    \node (3a) {
        \raisebox{-0.5\height}{\includegraphics[height=0.09\linewidth, page=16]{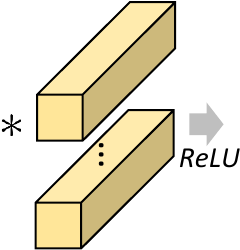}}
    };&
    \node (3b) {
        \raisebox{-0.5\height}{\includegraphics[height=0.11\linewidth, page=17]{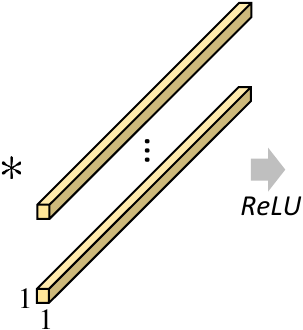}}
    };&
    \node (3c) {
        \raisebox{-0.5\height}{\includegraphics[height=0.11\linewidth, page=17]{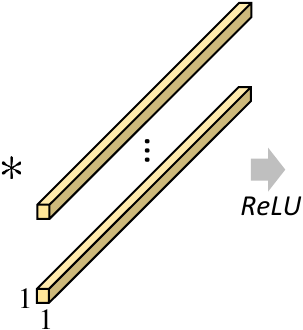}}
    };&
    \node (7) {
        {\Large $\cdots$}
    };\\
    \draw node{{\footnotesize \textit{input image}} \hspace{0.7em} {\footnotesize \textit{conv1a}}};&
    \draw node{\footnotesize \textit{conv1b}};&
    \draw node{\footnotesize \textit{conv1c}};&
    \draw node{\footnotesize \textit{conv2a}};&
    \draw node{\footnotesize \textit{conv2b}};&
    \draw node{\footnotesize \textit{conv2c}};&
    \draw node{\footnotesize \textit{conv3a}};&
    \draw node{\footnotesize \textit{conv3b}};&
    \draw node{\footnotesize \textit{conv3c}};\\
    };
    \end{scope}
\end{tikzpicture}
    \caption{Standard}
    \label{fig:standardtopology}
\end{subfigure}
~
\begin{subfigure}[c]{\linewidth}
\begin{tikzpicture}
    \begin{scope}[]
    \matrix[column sep=0em]{
    \node (1a) {
        \includegraphics[height=0.08\linewidth, page=15]{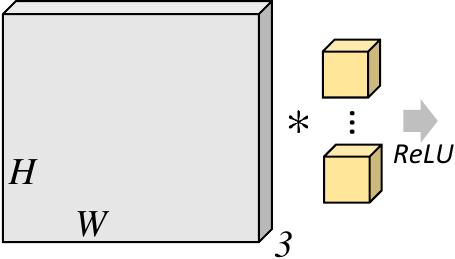}
    };&
    \node (1b) {
        \includegraphics[height=0.11\linewidth, page=17]{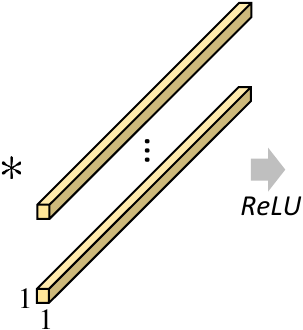}
    };& 
    \node (1c) {
        \includegraphics[height=0.11\linewidth, page=17]{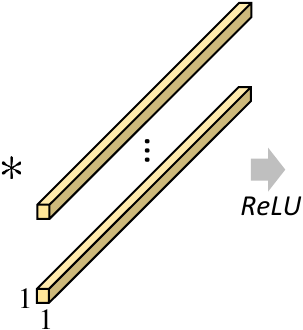}
    };& 
    \node (2a) {
        \includegraphics[height=0.15\linewidth, page=19]{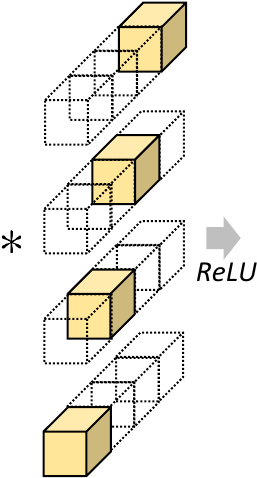}
    };&
    \node (2b) {
        \includegraphics[height=0.11\linewidth, page=17]{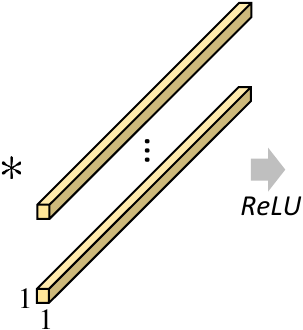}
    };&
    \node (2c) {
        \includegraphics[height=0.11\linewidth, page=17]{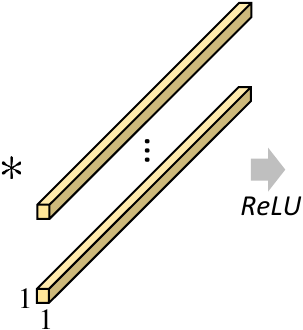}
    };&
    \node (3a) {
        \includegraphics[height=0.09\linewidth, page=18]{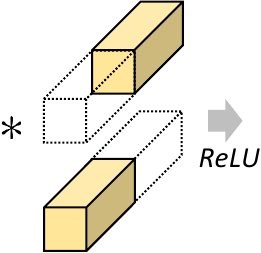}
    };&
    \node (3b) {
        \includegraphics[height=0.11\linewidth, page=17]{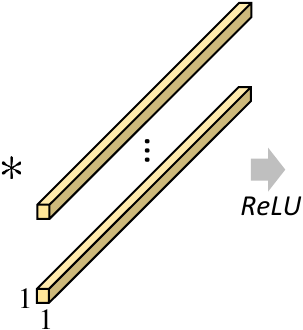}
    };&
    \node (3c) {
        \includegraphics[height=0.11\linewidth, page=17]{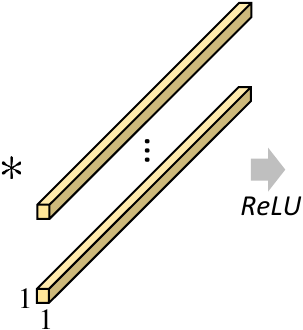}
    };&
    \node (4) {
        {\Large $\cdots$}
    };\\
    };
    \draw[decorate,decoration={brace,mirror},](2a.south west) -- node[below=3pt] {\small root-4 module} ++(3.5, 0);
    \draw[decorate,decoration={brace,mirror},yshift=-2em](3a.south west) + (0, -0.5) -- node[below=3pt] {\small root-2 module} ++(3.5, -0.5);
    \end{scope}
\end{tikzpicture}
    \caption{Root-4 Architecture}
    \label{fig:root4topology}
\end{subfigure}
\caption{\textbf{Network-in-Network Root Architecture.} The Root-4 architecture as compared to the original architecture for all the convolutional layers. Colored blocks represent the filters of each layer. Here we don't show the intermediate feature maps over which a layer's filters operate, or the final fully connected layer, out of space considerations (see Fig.\ref{fig:rootmodule}). The decreasing degree of grouping in successive root modules means that our network architectures somewhat resemble plant roots, hence the name root.
}
\label{fig:networktopology}
\end{figure*}

An interesting question concerns how the degree of grouping in our root modules should be varied as a function of depth in the network. For the NiN-like architectures described earlier, we might consider having the degree of grouping: (1) decrease with depth after the first convolutional layer, \eg 1--8--4 (`root'); (2) remain constant with depth after the first convolutional layer, \eg 1--4--4 (`column'); or (3) increase with depth, \eg 1--4--8 (`tree').

To determine which approach is best, we created variants of the NiN architecture with different degrees of grouping per layer. Results are shown in Fig.~\ref{fig:nincifarplotsconvonly} (numerical results are included in \S\ref{nincifarfull}). The results show that the so-called root topology (illustrated in Fig.~\ref{fig:networktopology}) gives the best performance, providing the smallest reduction in accuracy for a given reduction in model size and computational complexity. Similar experiments with deeper network architectures have delivered similar results and so we have reported results for root topologies. This aligns with the intuition of deep networks for image recognition subsuming the deformable parts model. If we assume that filter responses identify parts (or more elemental features), then there should be more filter dependence with depth, as more parts (filter responses) are assembled into complex concepts.

\subsection{Improving Residual Networks on ILSVRC}
Residual networks (ResNets)~\citep{He2015} are the state-of-the art network for ILSVRC. ResNets are more computationally efficient than the VGG architecture~\cite{Simonyan2014verydeep} on which they are based, due to the use of low-dimensional embeddings~\citep{Lin2013NiN}. ResNets are also more accurate and quicker to converge due to the use of identity mappings.

\subsubsection{ResNet 50}
\label{resnet50results}
As a baseline, we used the `ResNet 50' model~\citep{He2015} (the largest residual network model to fit onto 8 GPUs with Caffe). ResNet 50 has 50 convolutional layers, of which one-third are spatial convolutions (non-1$\times$1). We did not use any training augmentation aside from random cropping and mirroring. 
For training, we used the initialization scheme described by \citep{He2015b} modified for compound layers~\citep{Ioannou2016} and batch normalization~\citep{Ioffe2015}.
\begin{table}[tbp]
\caption{\textbf{ResNet 50}. Filter groups in each conv.\ layer.}
\label{table:resnet50config}
\centering
\begin{tabular}{@{}lcccccccccccc@{}}
\toprule
    Model & conv1 & \multicolumn{2}{c}{res2\{a--c\}} & \multicolumn{2}{c}{res3\{a--d\}} & \multicolumn{2}{c}{res4\{a--f\}} & \multicolumn{2}{c}{res5\{a--c\}} \\
     & \textit{\footnotesize7$\times$7} & \textit{\footnotesize1$\times$1} & \textit{\footnotesize3$\times$3} & \textit{\footnotesize1$\times$1} & \textit{\footnotesize3$\times$3} & \textit{\footnotesize1$\times$1} & \textit{\footnotesize3$\times$3} & \textit{\footnotesize1$\times$1} & \textit{\footnotesize3$\times$3} \\
    Orig. & 1 & 1 & 1 & 1 &  1 & 1 &  1 & 1 & 1 \\
    \midrule
    root-2 & 1 & 1 & 2 & 1 &  1 & 1 &  1 & 1 & 1 \\
    root-4 & 1 & 1 & 4 & 1 &  2 & 1 &  1 & 1 & 1 \\
    root-8 & 1 & 1 & 8 & 1 &  4 & 1 &  2 & 1 & 1 \\
    root-16 & 1 & 1 & 16 & 1 &  8 & 1 &  4 & 1 & 2 \\
    root-32 & 1 & 1 & 32 & 1 & 16 & 1 &  8 & 1 & 4 \\
    root-64 & 1 & 1 & 64 & 1 & 32 & 1 & 16 & 1 & 8 \\
    \bottomrule
\end{tabular}
\end{table}
To assess the efficacy of our method, we replaced the spatial convolutional layers of the original network with root modules (as described in \S\ref{method}). We preserved the original number of filters per layer but subdivided them into groups as shown in Table~\ref{table:resnet50config}. We considered the first of the existing 1$\times$1 layers subsequent to each spatial convolution to be part of our root modules. 

\input{resnet50matable.tex}
\input{resnet50maplot.tex}
Results are shown in Table \ref{table:resnet50imagenetresults} and Fig.~\ref{fig:resnet50plots} for various network architectures. Compared to the baseline architecture, the root variants achieve a significant reduction in computation and model size without a significant reduction in accuracy. For example, the best result (root-16) exceeds the baseline accuracy by 0.2\% while reducing the model size by 27\% and floating-point operations (multiply-add) by 37\%. CPU timings were 23\% faster, while GPU timings were 13\% faster. With a drop in accuracy of only 0.1\% however, the root-64 model reduces the model size by 40\%, and reduces the floating point operations by 45\%. CPU timings were 31\% faster, while GPU timings were 12\% faster. 

\subsubsection{ResNet 200}
\label{resnet200results}
\input{resnet200matable.tex}
To show that the method applies to deeper architectures, we also applied our method to ResNet 200, the deepest network for ILSVRC 2012. To provide a baseline we used code implementing full training augmentation to achieve state-of-the-art results\footnote{\url{https://github.com/facebook/fb.resnet.torch}}. Table \ref{table:resnet200imagenetresults} shows the results of these experiments, top-1 and top-5 error are for center cropped images. The models trained with roots have comparable error to the baseline network, with fewer parameters and less computation. The root-32 model has 25\% fewer FLOPS and 44\% fewer parameters than ResNet 200.

\subsection{Improving GoogLeNet on ILSVRC}
\label{googlenet50results}
\input{googlenetmatable.tex}

\begin{table}[tbp]
\caption[GoogLeNet Architectures]{\textbf{GoogLeNet}. Filter groups in each convolutional layer and Inception module (\textit{incp}.)}
\label{table:googlenetconfig}
\centering
\begin{tabular}{@{}lm{1.6em}m{1.1em}m{1.1em}m{1.1em}m{1.1em}m{1.1em}m{1.1em}m{1.1em}m{1.1em}m{1.1em}m{1.1em}m{1.1em}@{}}
\toprule
    Model & conv1 & \multicolumn{2}{c}{conv2} & \multicolumn{3}{c}{incp.~3\{a,b\}} & \multicolumn{3}{c}{incp.~4\{a--e\}} & \multicolumn{3}{c}{incp.~5\{a,b\}} \\
     & \textit{\footnotesize7$\times$7} & \textit{\footnotesize1$\times$1} & \textit{\footnotesize3$\times$3} & \textit{\footnotesize1$\times$1} & \textit{\footnotesize3$\times$3} & \textit{\footnotesize5$\times$5} & \textit{\footnotesize1$\times$1} & \textit{\footnotesize3$\times$3} & \textit{\footnotesize5$\times$5} & \textit{\footnotesize1$\times$1} & \textit{\footnotesize3$\times$3} & \textit{\footnotesize5$\times$5} \\
    Orig. & 1 &  1 &  1 & 1 & 1 & 1 & 1 & 1 & 1 & 1 & 1 & 1\\ 
    \midrule
    root-2 & 1 &  1 &  2 & 1 & 1 & 1 & 1 & 1 & 1 & 1 & 1 & 1\\
    root-4 & 1 &  1 &  4 & 1 & 2 & 2 & 1 & 1 & 1 & 1 & 1 & 1\\
    root-8 & 1 &  1 &  8 & 1 & 4 & 4 & 1 & 2 & 2 & 1 & 1 & 1\\
    root-16 & 1 &  1 & 16 & 1 & 8 & 8 & 1 & 4 & 4 & 1 & 2 & 2\\
    \bottomrule
\end{tabular}
\end{table}

We replicated the network as described by~\citet{Szegedy2014going}, with the exception of not using any training augmentation aside from random crops and mirroring (as supported by Caffe~\cite{Jia2014}). To train we used the initialization of \citep{He2015b} modified for compound layers~\citep{Ioannou2016} and batch normalization without the scale and bias~\citep{Ioffe2015}. At test time we only evaluate the center crop image.

While preserving the original number of filters per layer, we trained networks with various degrees of filter grouping, as described in Table~\ref{table:googlenetconfig}.  While the inception architecture is relatively complex, for simplicity, we always use the same number of groups within each of the groups of different filter sizes, despite them having different cardinality. For all of the networks, we only grouped filters within each of the `spatial' convolutions (3$\times$3, 5$\times$5).
\input{googlenetmaplot.tex}

As shown in Table \ref{table:googlenetimagenetresults}, and plotted in Fig.~\ref{fig:googlenet50plots}, our method shows significant reduction in computational complexity -- as measured in FLOPS (multiply-adds), CPU and GPU timings -- and model size, as measured in the number of floating point parameters. For many of the configurations the top-5 accuracy remains within 0.5\% of the baseline model. 
The highest accuracy result, is 0.1\% off the top-5 accuracy of the baseline model, but has a 0.1\% higher top-1 accuracy -- within the error bounds resulting from training with different random initializations. While maintaining the same accuracy, this network has 9\% faster CPU and GPU timings. However, a model with only 0.3\% lower top-5 accuracy than the baseline has much higher gains in computational efficiency -- 44\% fewer floating point operations (multiply-add), 7\% fewer model parameters, 21\% faster CPU and 16\% faster GPU timings.

While these results may seem modest compared to the results for ResNet, GoogLeNet is by far the smallest and fastest near state-of-the-art model ILSVRC model. We believe that more experimentation in using different cardinalities of filter grouping in the heterogeneously-sized filter groups within each inception module will improve results further.

\section{GPU Implementation}
\label{gpuexplanation}
Our experiments show that our method can achieve a significant reduction in CPU and GPU runtimes for state-of-the-art CNNs without compromising accuracy. However, the reductions in GPU runtime were smaller than might have been expected based on theoretical predictions of computational complexity (FLOPs). We believe this is largely a consequence of the optimization of Caffe for existing network architectures (particularly AlexNet and GoogLeNet) that do not use a high degree of filter grouping.

Caffe presently parallelizes over filter groups by using multiple CUDA streams to run multiple CuBLAS matrix multiplications simultaneously. However, with a large degree of filter grouping, and hence more, smaller matrix multiplications, the overhead associated with calling CuBLAS from the host can take approximately as long as the matrix computation itself. To avoid this overhead, CuBLAS provides batched methods (\eg \texttt{cublasXgemmBatched}), where many small matrix multiplications can be batched together in one call. \citet{Jhurani2015} explore in depth the problem of using GPUs to accelerate the multiplication of very small matrices (smaller than 16$\times$16), and show it is possible to achieve high throughput with large batches, by implementing a more efficient interface than that used in the CuBLAS batched calls.
We have modified Caffe to use CuBLAS batched calls, and achieved significant speedups for our  root-like network architectures compared to vanilla Caffe without CuDNN, e.g. a 25\% speed up on our root-16 modified version of the GoogleNet architecture. However, our optimized implementation still is not as fast as Caffe with CuDNN (which was used to generate the results in this paper), presumably because of other unrelated optimizations in the (proprietary) CuDNN library. Therefore we suggest that direct integration of CuBLAS-style batching into CuDNN could improve the performance of filter groups significantly.

\section{Future Work} 
In this paper we focused on using homogeneous filter groups (with a uniform division of filters in each group), however this may not be optimal. Heterogeneous filter groups may reflect better the filter co-dependencies found in deep networks. Learning a combined spatial~\citep{Ioannou2016} and channel basis, may also improve efficiency further.

\section{Conclusion}
We explored the effect of using complex hierarchical arrangements of filter groups in CNNs and show that imposing a structured decrease in the degree of filter grouping with depth -- a `root' (inverse tree) topology -- can allow us to obtain more efficient variants of state-of-the-art networks without compromising accuracy. Our method appears to be complementary to existing methods, such as low-dimensional embeddings, and can be used more efficiently to train deep networks than methods that only approximate a pre-trained model's weights.

We validated our method by using it to create more efficient variants of state-of-the-art Network-in-network, GoogLeNet, and ResNet architectures, which were evaluated on the CIFAR10 and ILSVRC datasets. Our results show similar accuracy with the baseline architecture with fewer parameters and much less compute (as measured by CPU and GPU timings). For Network-in-Network on CIFAR10, our model has 33\% of the parameters of the original network, and approximately 37\% (23\%) faster CPU (GPU) timings. For ResNet 50, our model has 27\% fewer parameters, and was 24\% (11\%) faster on a CPU (GPU). 
Even for the most efficient of the near state-of-the-art ILSVRC network, GoogLeNet, our model uses 7\% fewer parameters and is 21\% (16\%) faster on a CPU (GPU).

\renewcommand\bibname{\refname}
{\small
\printbibliography
}
\cleardoublepage
\appendix
\section{Appendices}
\subsection{Full Network-in-Network Results}
\label{nincifarfull}
\begin{table}[htb]
\caption[Network-in-Network CIFAR10]{\textbf{Network-in-Network CIFAR10}}
\label{table:nincifarresultsfull}
\centering
\pgfplotstableread[col sep=comma]{data/nincifar.csv}\data
\pgfplotstableread[col sep=comma]{data/nincifar_root_s.csv}\codata
\pgfplotstableread[col sep=comma]{data/nincifar_tree_s.csv}\tdatatable
\pgfplotstableread[col sep=comma]{data/nincifar_col_s.csv}\cdatatable
\pgfplotstablevertcat{\data}{\codata}
\pgfplotstablevertcat{\data}{\tdatatable}
\pgfplotstablevertcat{\data}{\cdatatable}
\pgfplotstabletypeset[
    every head row/.style={before row=\toprule,after row=\midrule},
    every last row/.style={after row=\bottomrule},
    every row no 0/.style={after row=\midrule}, 
    every nth row={4}{after row=\midrule}, 
    fixed zerofill,     
    columns={full name, ma, param, accuracy, cpu, gpu},
    columns/full name/.style={
        column name=Model,
        string type
    },
    columns/ma/.style={
        column name=FLOPS {\small $\times 10^{8}$},
        preproc/expr={{##1/1e8}}
    },
    columns/param/.style={
        column name=Param. {\small $\times 10^{5}$},
        preproc/expr={{##1/1e5}}
    },
    columns/accuracy/.style={
        column name=Accuracy,
        precision=4
    },
    columns/gpu/.style={
        column name=GPU (ms),
        precision=3
    },
    columns/cpu/.style={
        column name=CPU (ms),
        precision=1
    },
    column type/.add={@{}lp{3em}p{3em}p{4em}p{3em}p{3em}p{3em}@{}}{},
    highlight col max ={\data}{accuracy},
    highlight col min ={\data}{param}, 
    highlight col min ={\data}{ma}, 
    col sep=comma]{\data}
\end{table}

Table \ref{table:nincifarresultsfull} shows the full results for the Network-in-Network experiments on CIFAR10 on various hierarchical network topologies.

\subsection{Inter-Layer Covariance}
\label{interlayercovar}
To show the relationships between filters between adjacent convolutional layers, as illustrated in Fig.~1
, we calculate the covariance of the responses from two adjacent featuremaps, the outputs of convolutional layers with $c_1$ and $c_2$ filters.

Let $X_i = [\mathbf{x}_{i,1}; \mathbf{x}_{i,2}; \ldots ; \mathbf{x}_{i,N}]$ be the matrix of $N$ samples $\mathbf{x}_{i,n}$ from the $c_i$ dimensional featuremap for layer $i$. We consider each pixel across the two featuremaps to be a sample, and thus each vector $\mathbf{x}_{i,n}$ is a single pixel filter response of dimension $c_i$. If two featuremaps have different spatial dimensions, due to pooling, we up-sample the smaller featuremap (with nearest neighbor interpolation) such that there are the same number of pixels (and thus samples) in each featuremap.

Given two samples $X_1, X_2$ with zero mean (\ie mean subtracted) for two adjacent featuremaps, we calculate the inter-layer covariance,

\begin{eqnarray}
    cov(X_1, X_2) &=& E\left[X_1\,X_2^\textrm{T}\right],\\
    &=& \frac{1}{N-1} X_1\,X_2^\textrm{T}.
\end{eqnarray}

\renewcommand{\covarlabels}[5]{%
\begin{tikzpicture}[anchor=south west]
    \node [inner sep=0pt] (c)
    {
        #5
    };
    \ifx\covarwidth\undefined
    \newlength{\covarwidth}
    \newlength{\covarheight}
    \fi
    \settowidth{\covarwidth}{#5}
    \settoheight{\covarheight}{#5}
    \path[use as bounding box] (c.south west) rectangle (c.north east);
    \node [anchor=south west, xshift=-0.5em, yshift=-0.5em, rotate=45] at (c.north west) {\footnotesize 0};
    \node [anchor=south east, xshift=\covarwidth, yshift=-0.2em] at (c.north west) {\footnotesize #4};
    \node [anchor=south west, xshift=0.25em, yshift=-1.05\covarheight, rotate=90] at (c.north west) {\footnotesize #2};
    \node [anchor=south, xshift=0.5\covarwidth] at (c.north west) {\footnotesize\texttt{#3}};
    \node [anchor=south, xshift=0.2em, yshift=-0.5\covarheight, rotate=90] at (c.north west) {\footnotesize \texttt{#1}};
\end{tikzpicture}%
}

\begin{figure}[tbp]
\centering
\begin{subfigure}[b]{0.45\linewidth}
\centering
    \covarlabels{conv2c}{192}{conv3a}{192}{\includegraphics[width=\linewidth]{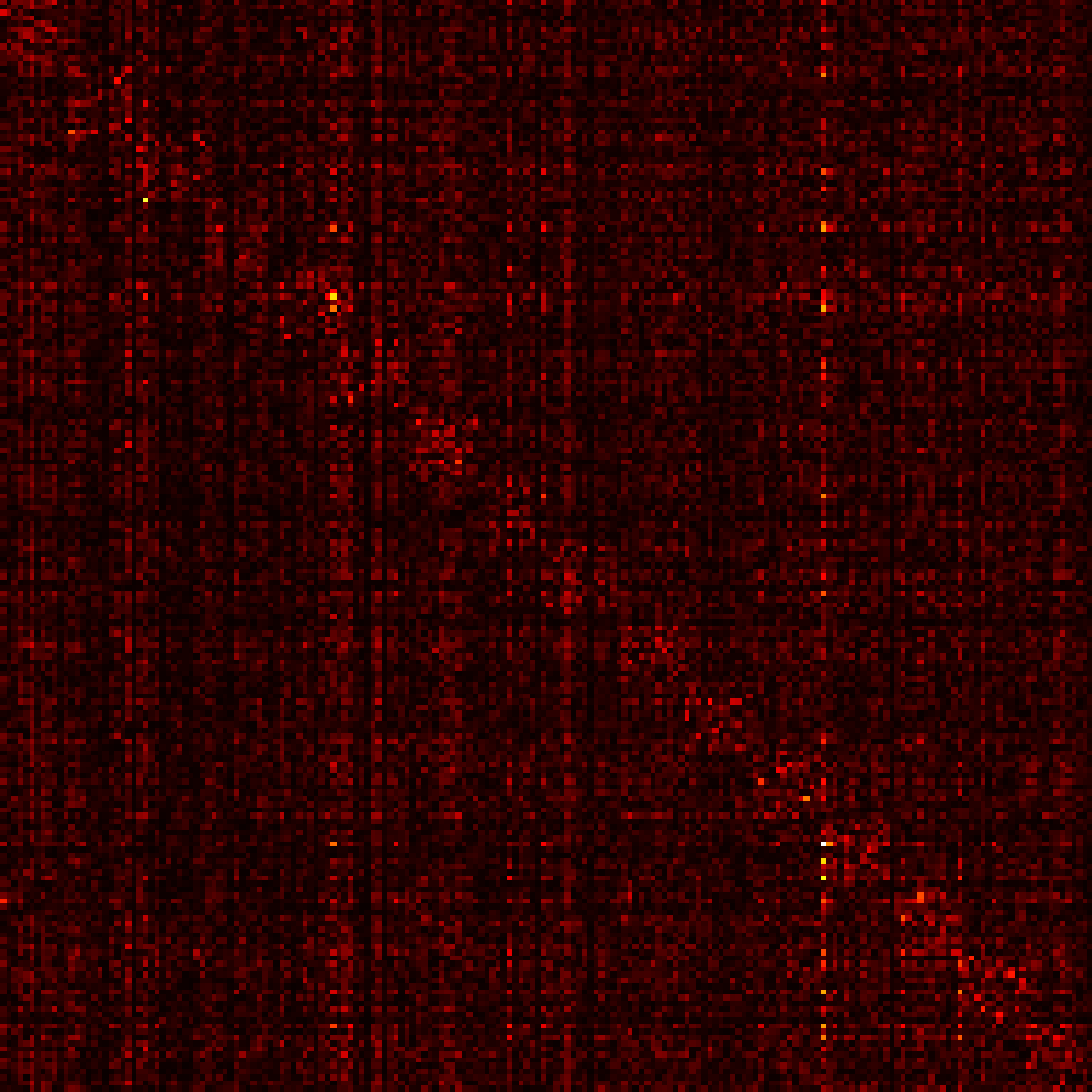}}
    \caption{Non-whitened responses}
    \label{fig:notwhitened}
\end{subfigure}
~
\begin{subfigure}[b]{0.45\linewidth}
\centering
    \covarlabels{conv2c}{192}{conv3a}{192}{\includegraphics[width=\linewidth]{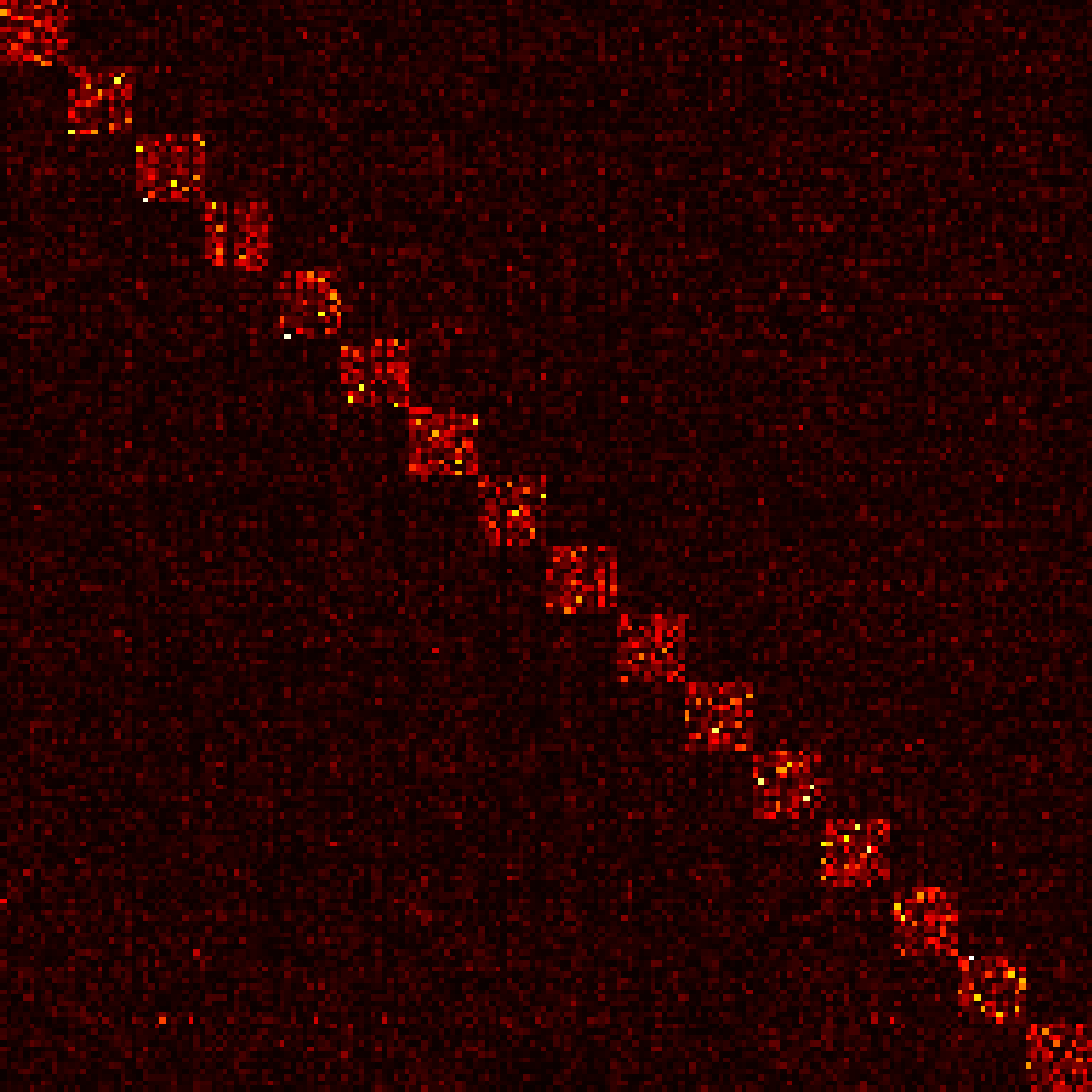}}
    \caption{Whitened responses}
    \label{fig:whitened}
\end{subfigure}
\caption{Covariance for between two layers in the root-32 Network-in-Network model with and without whitened responses}
\label{fig:whitevsnot}
\end{figure}

While this shows the covariance between layers, it is conflated with the inherent covariances within $X_1$ and $X_2$ from the data (as shown in Fig.~\ref{fig:notwhitened}). We can more clearly show the covariance between layers by first whitening (using ZCA~\cite{CIFAR10}) the samples in $X_1$ and $X_2$. For a covariance matrix,

\begin{equation}
    cov(X, X) = \frac{1}{N-1} XX^\textrm{T},
\end{equation}

The ZCA whitening transformation is given by,
\begin{equation}
    W = \sqrt{N-1}\left(XX^\textrm{T}\right)^{-\frac{1}{2}}.
\end{equation}

Since the covariance matrix is symmetric, it is easily diagonalizable (\ie PCA),
\begin{eqnarray}
    cov(X, X) &=& \frac{1}{N-1} XX^\textrm{T},\\
    &=&\frac{1}{N-1} PDP^\textrm{T},\\
\end{eqnarray}
where $P$ is a orthogonal matrix and $D$ a diagonal matrix. This diagonalization allows a simplified calculation of the whitening transformation (see the derivation in Appendix A of \cite{CIFAR10}),

\begin{equation}
    W = \sqrt{N-1}PD^{-\frac{1}{2}}P^\textrm{T},
\end{equation}
where $D^{-\frac{1}{2}}$ is simply D with an element-wise power of $-\frac{1}{2}$.

The covariance between the whitened featuremap responses is then,
\begin{equation}
    cov(W_1 X_1, W_2 X_2) = E\left[(W_1X_1)\,(W_2X_2)^\textrm{T}\right].
\end{equation}
\afterpage{%
\begin{landscape}
\begin{figure}[p]
\begin{subfigure}[c]{\paperwidth}
\centering
    \covarlabels{conv1a}{192}{conv1a}{192}{\includegraphics[width=0.11\textwidth]{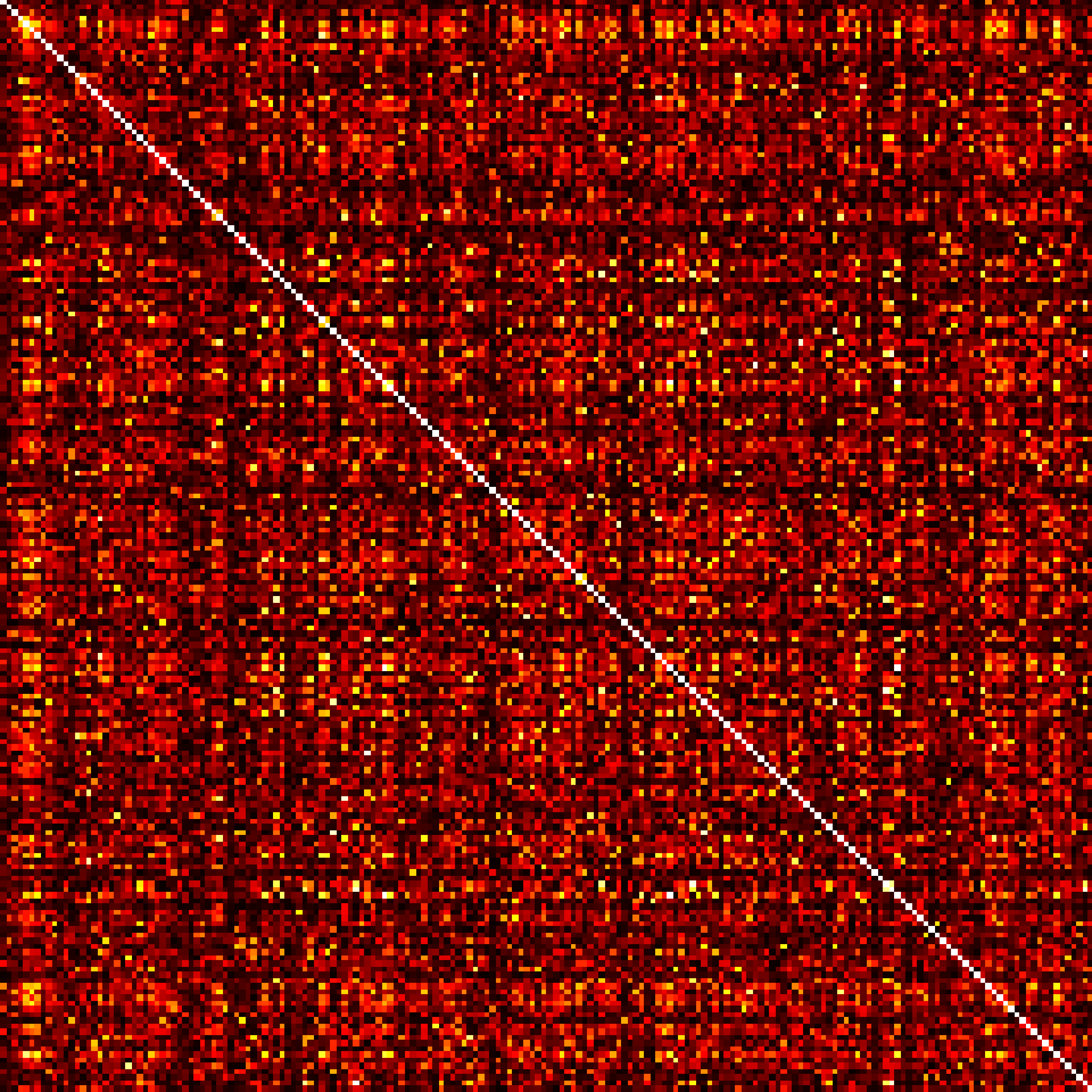}}
~
    \covarlabels{conv1b}{160}{conv1b}{160}{\includegraphics[width=0.11\textwidth]{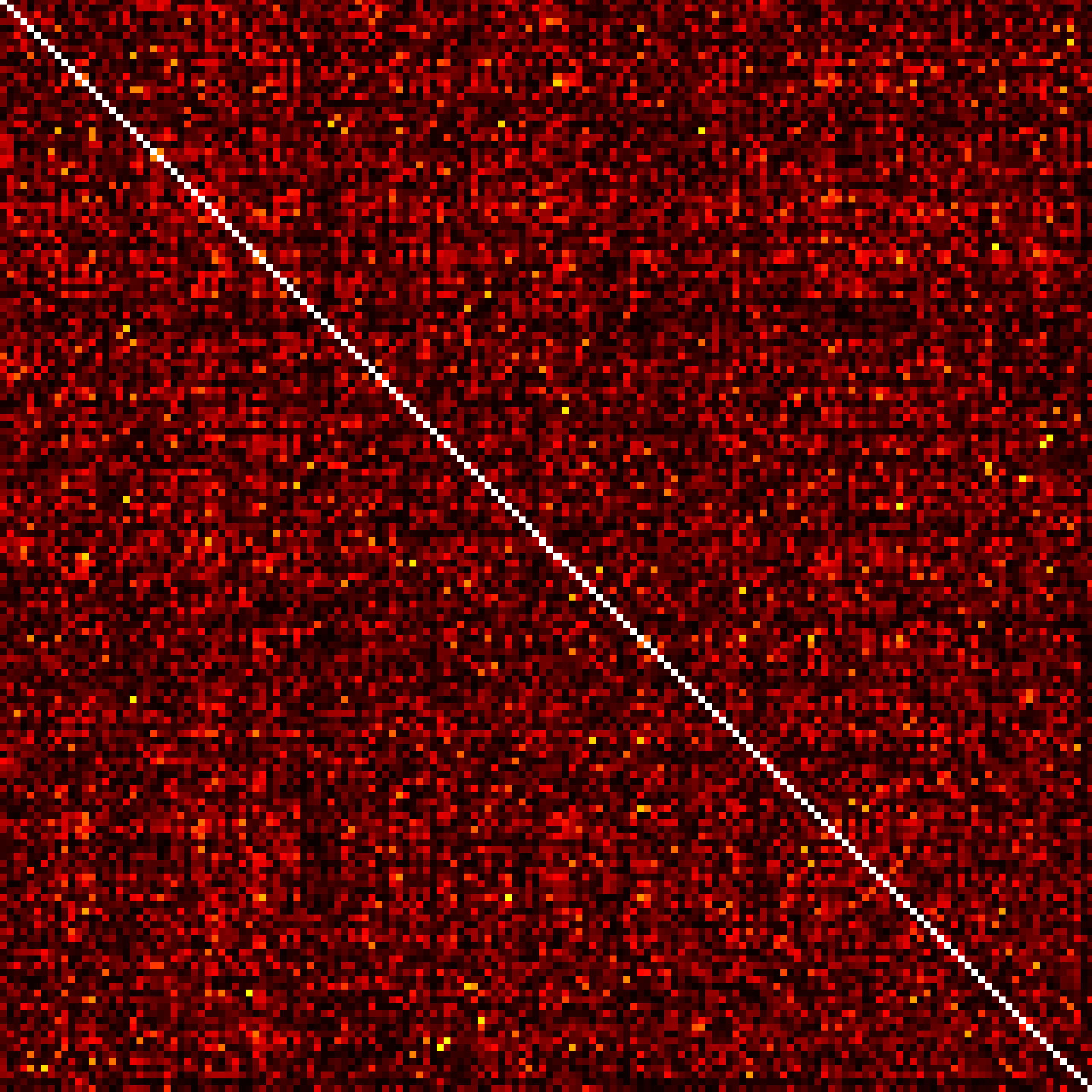}}
~
    \covarlabels{conv1c}{96}{conv1c}{96}{\includegraphics[width=0.11\textwidth]{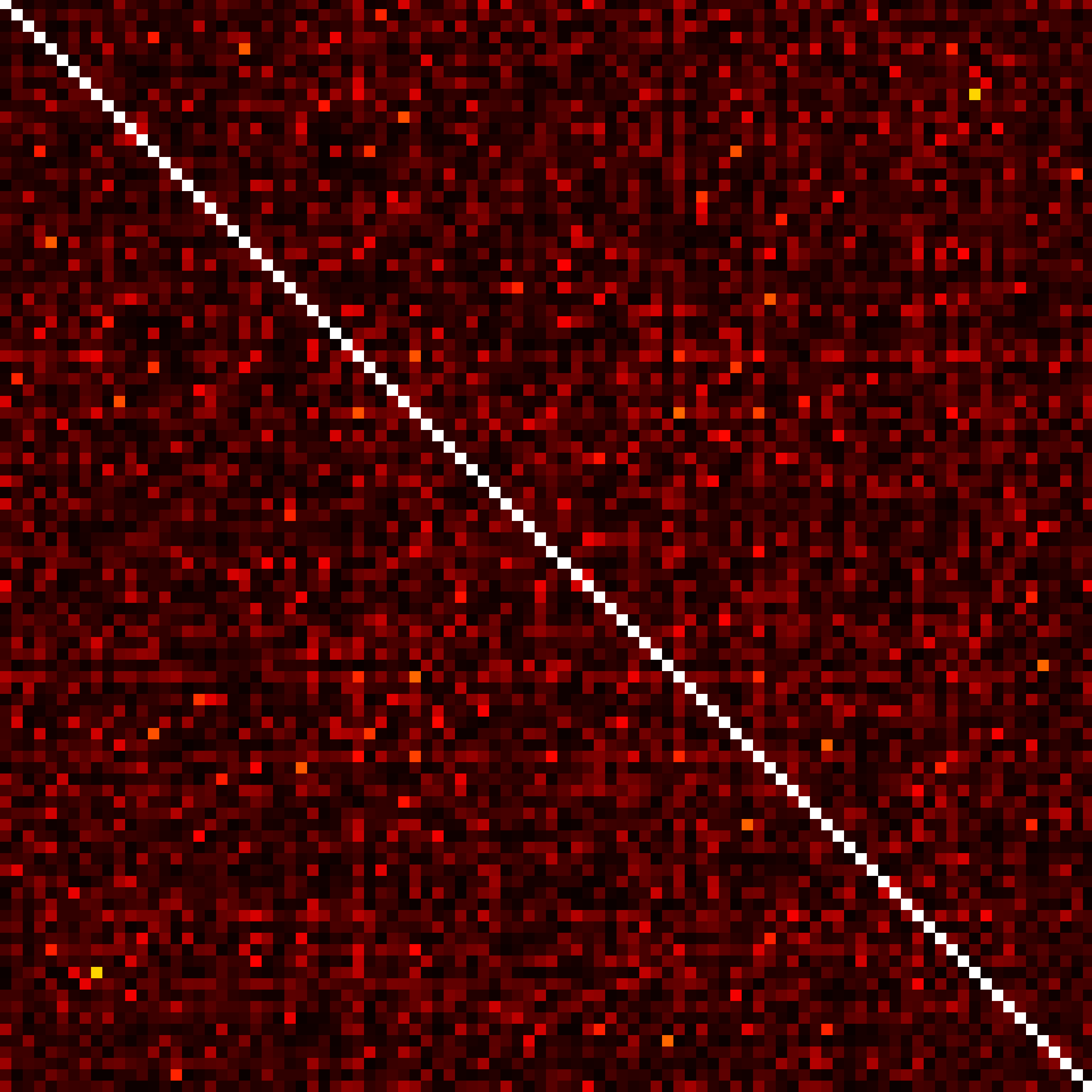}}
~
    \covarlabels{conv2a}{192}{conv2a}{192}{\includegraphics[width=0.11\textwidth]{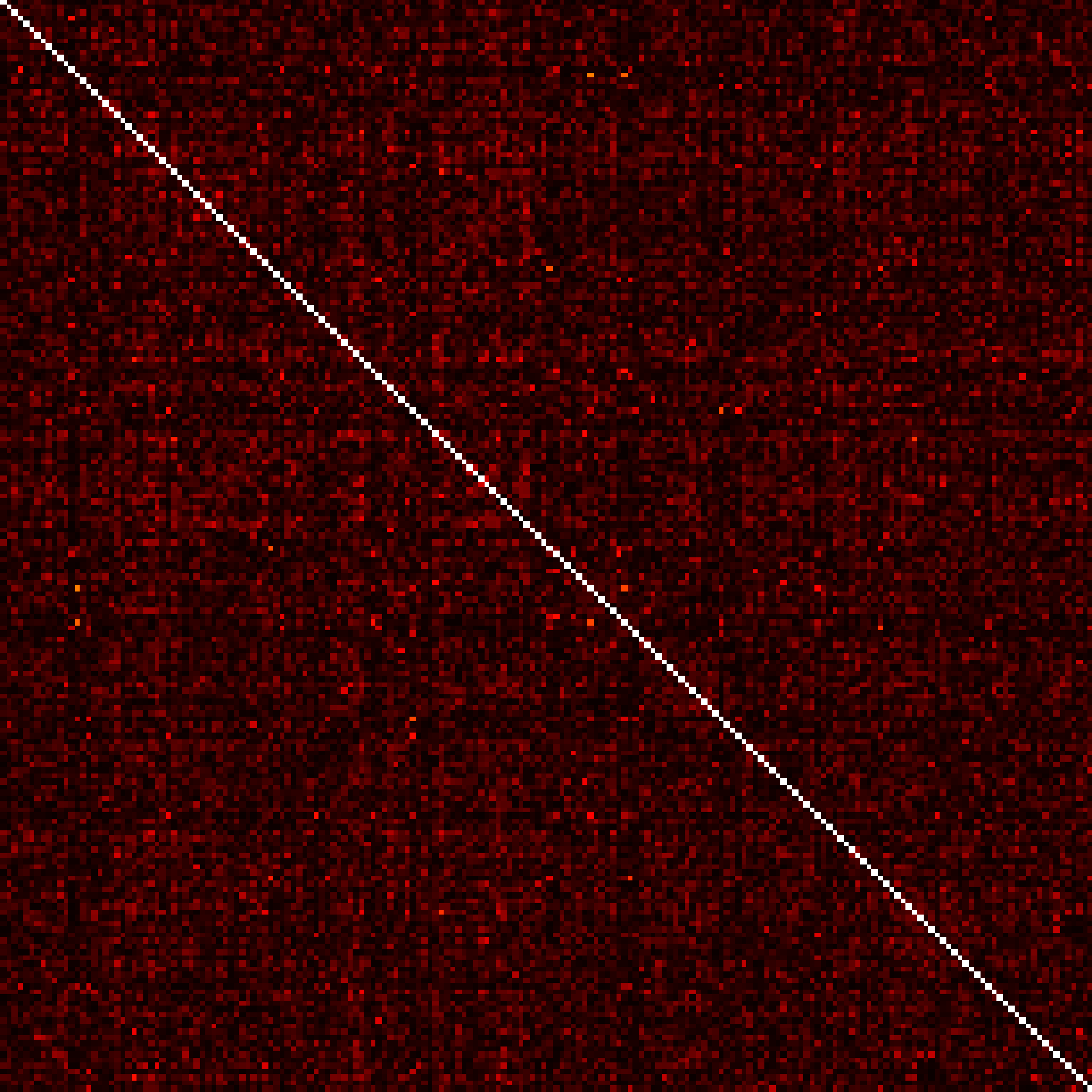}}
~
    \covarlabels{conv2b}{192}{conv2b}{192}{\includegraphics[width=0.11\textwidth]{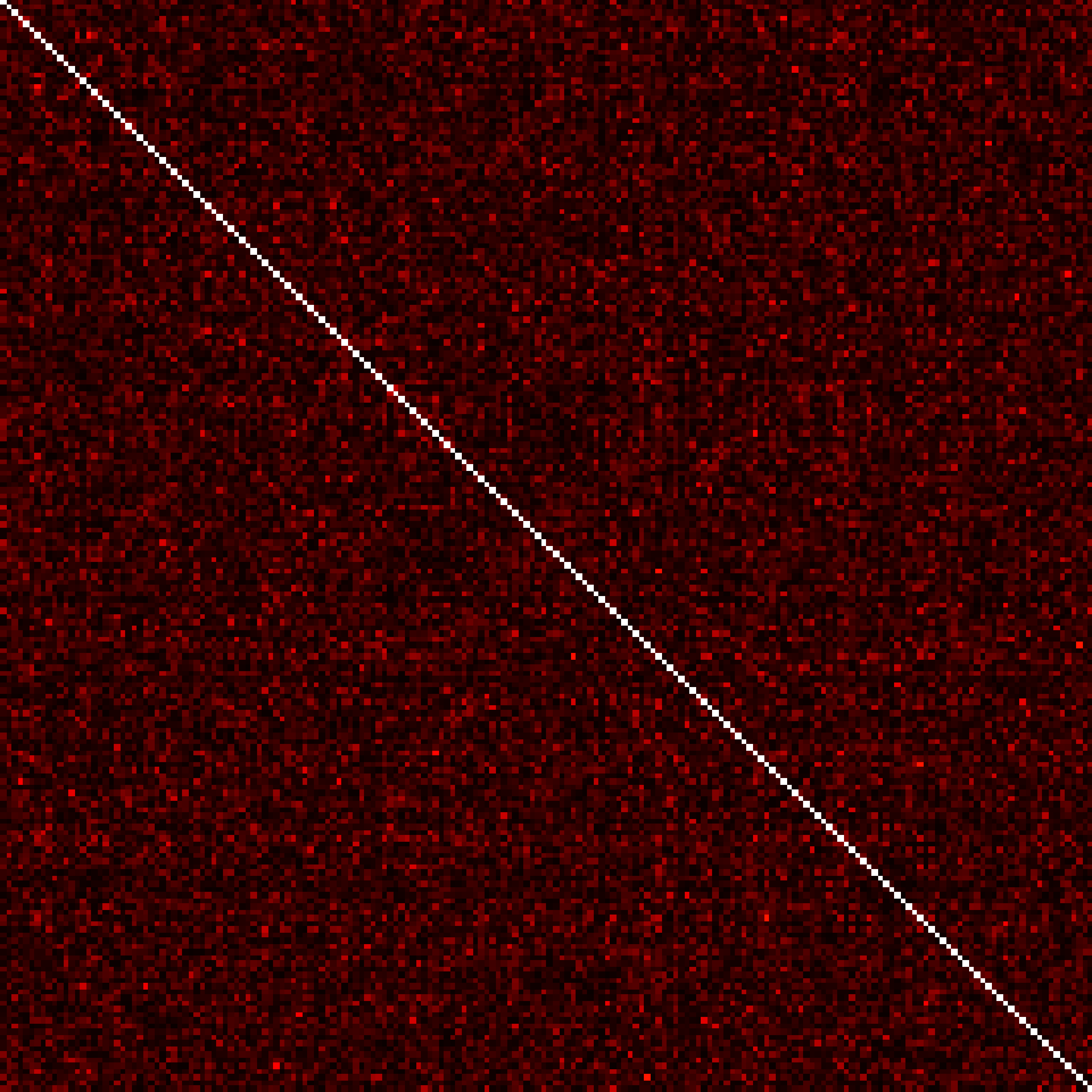}}
~
    \covarlabels{conv2c}{192}{conv2c}{192}{\includegraphics[width=0.11\textwidth]{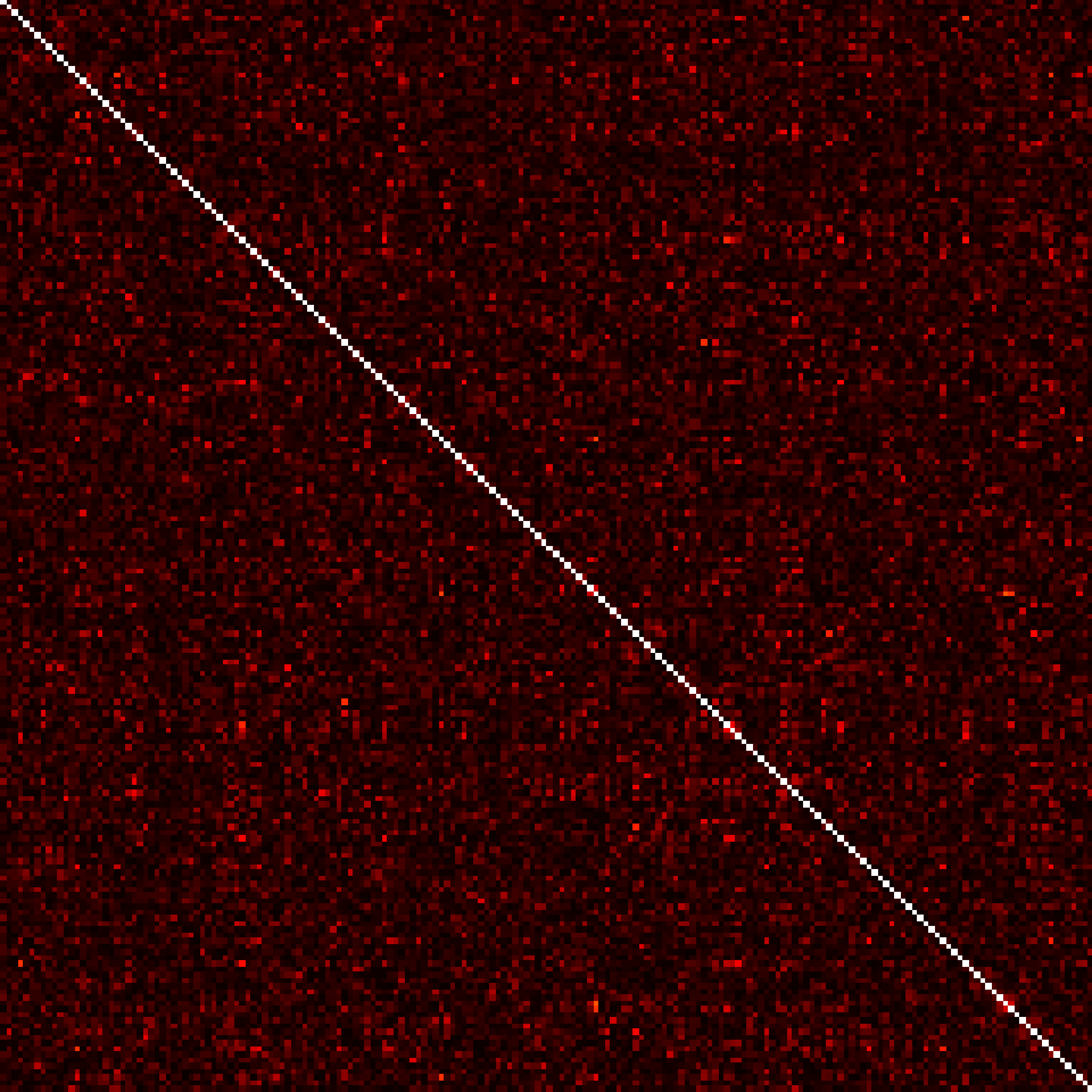}}
~
    \covarlabels{conv3a}{192}{conv3a}{192}{\includegraphics[width=0.11\textwidth]{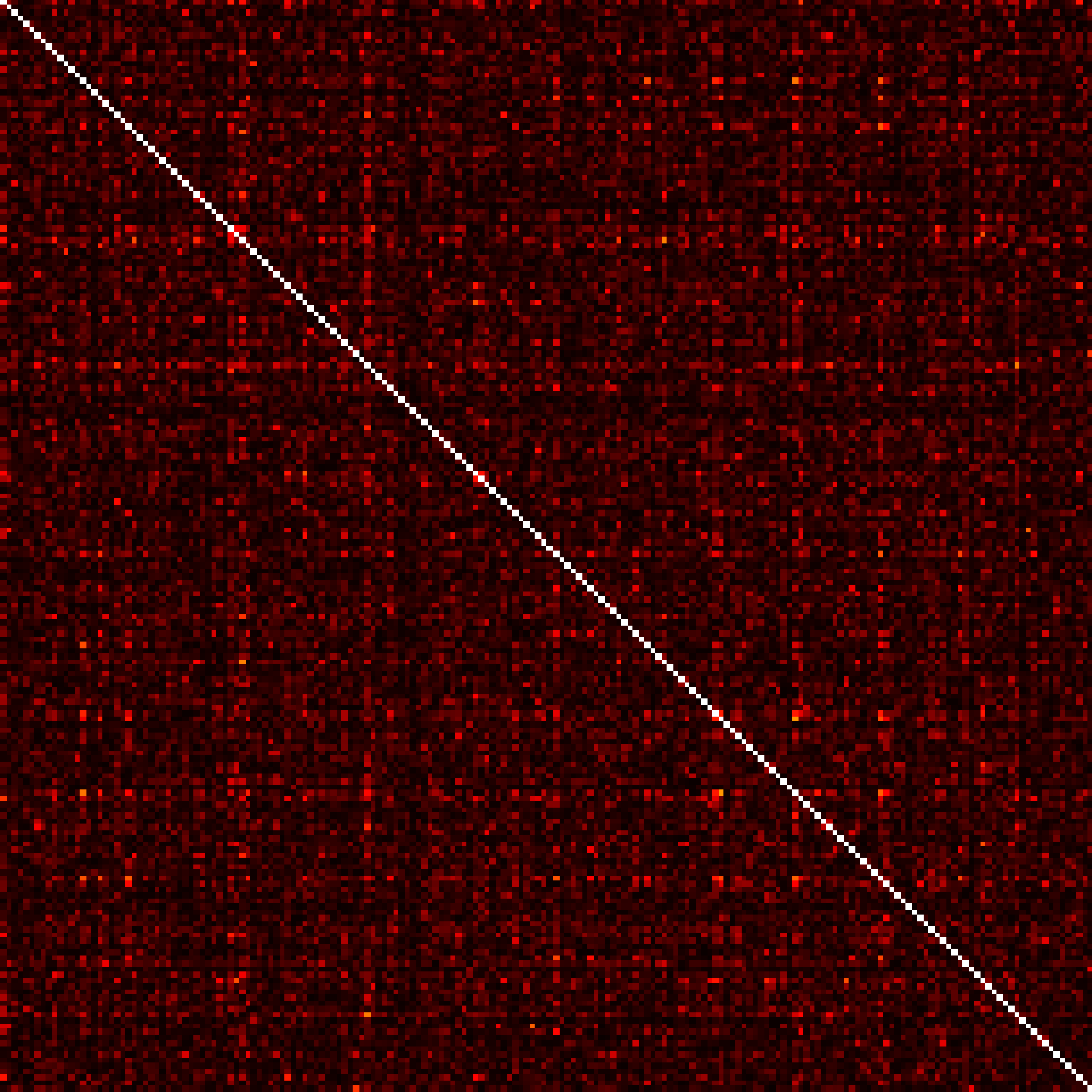}}
~
    \covarlabels{conv3b}{192}{conv3b}{192}{\includegraphics[width=0.11\textwidth]{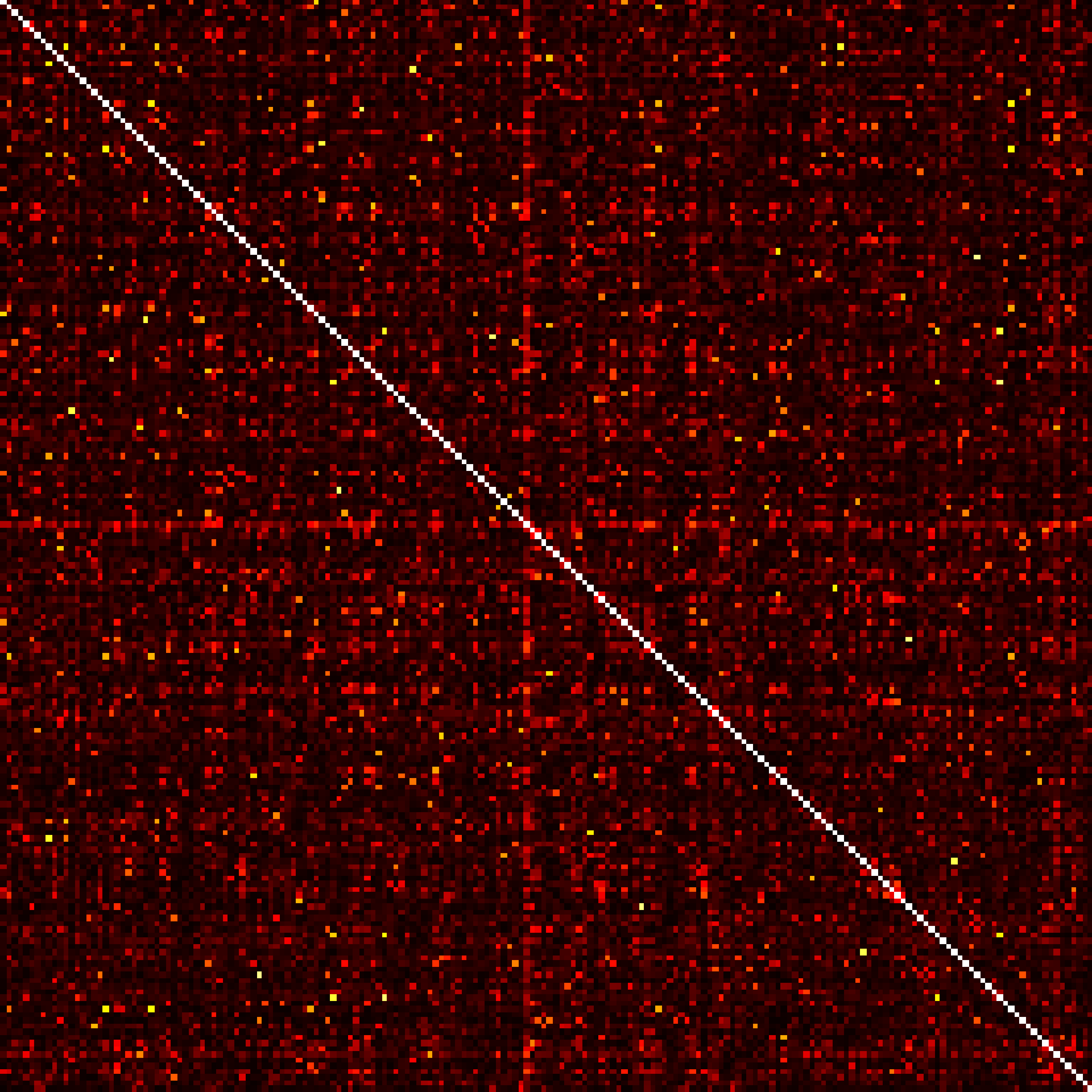}}
\label{fig:corrroot1}
\caption{\textbf{Network-in-Network.}}
\vspace*{0.6em}
\end{subfigure}
~
\begin{subfigure}[c]{\paperwidth}
\centering
    \covarlabels{conv1a}{192}{conv1a}{192}{\includegraphics[width=0.11\textwidth]{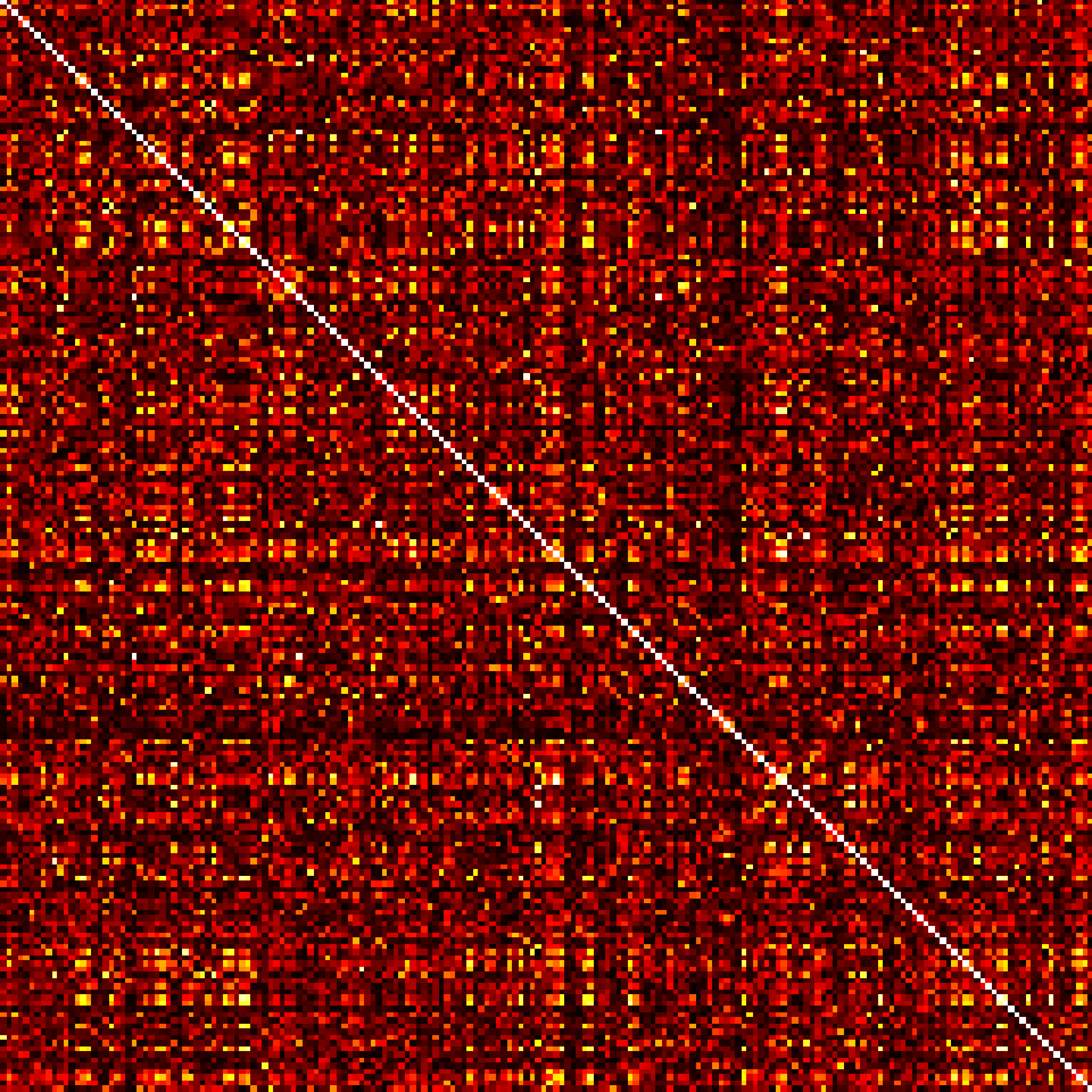}}
~
    \covarlabels{conv1b}{160}{conv1b}{160}{\includegraphics[width=0.11\textwidth]{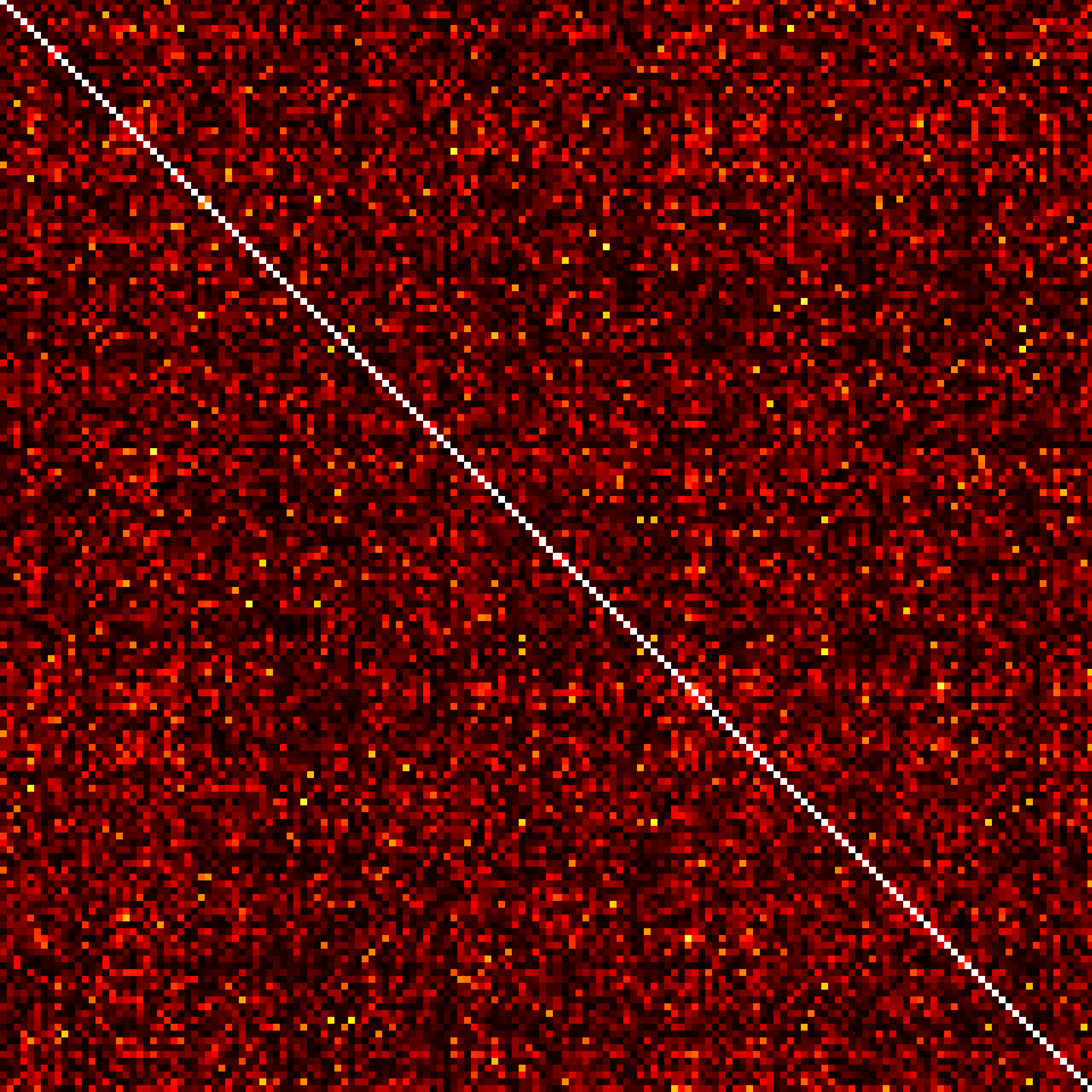}}
~
    \covarlabels{conv1c}{96}{conv1c}{96}{\includegraphics[width=0.11\textwidth]{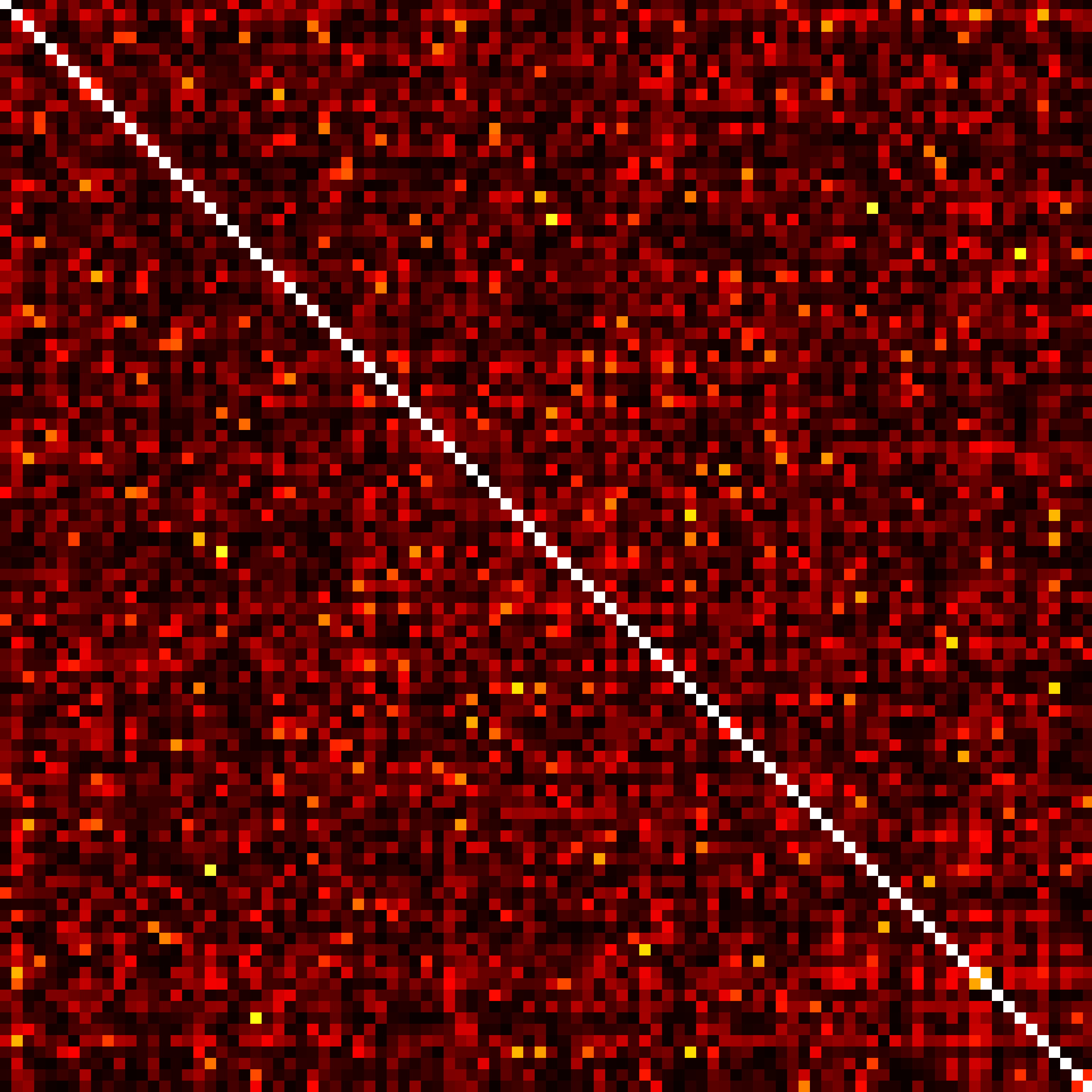}}
~
    \covarlabels{conv2a}{192}{conv2a}{192}{\includegraphics[width=0.11\textwidth]{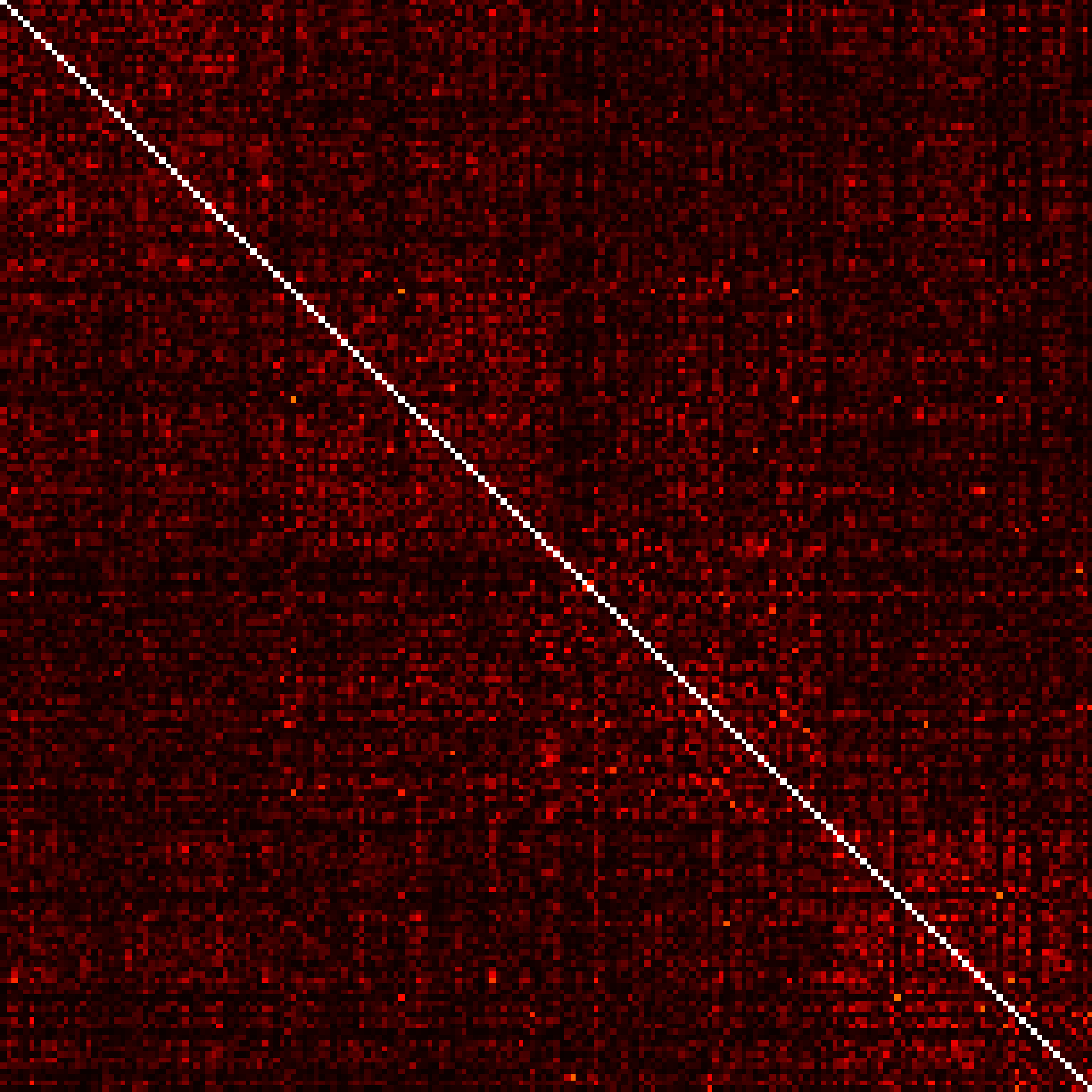}}
~
    \covarlabels{conv2b}{192}{conv2b}{192}{\includegraphics[width=0.11\textwidth]{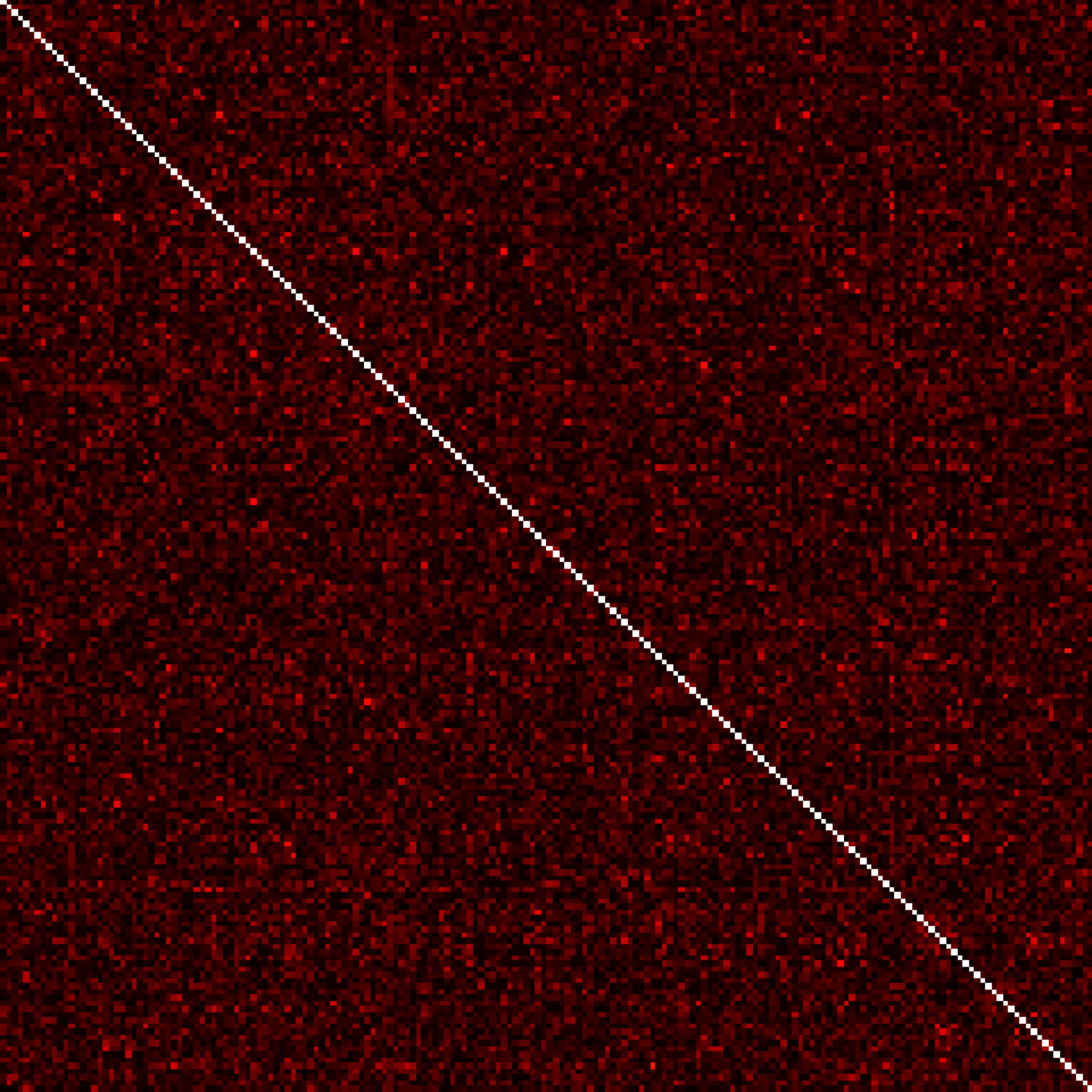}}
~
    \covarlabels{conv2c}{192}{conv2c}{192}{\includegraphics[width=0.11\textwidth]{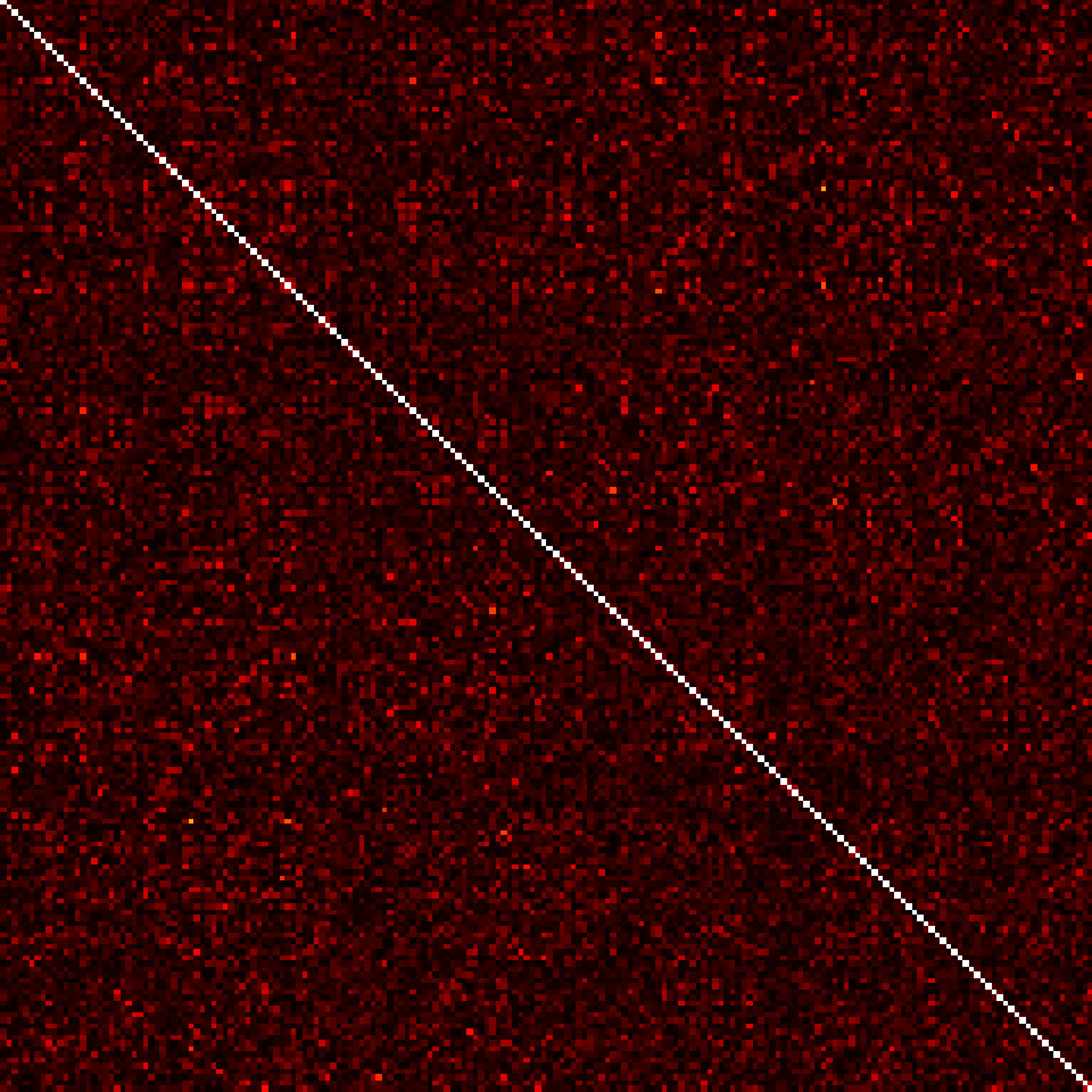}}
~
    \covarlabels{conv3a}{192}{conv3a}{192}{\includegraphics[width=0.11\textwidth]{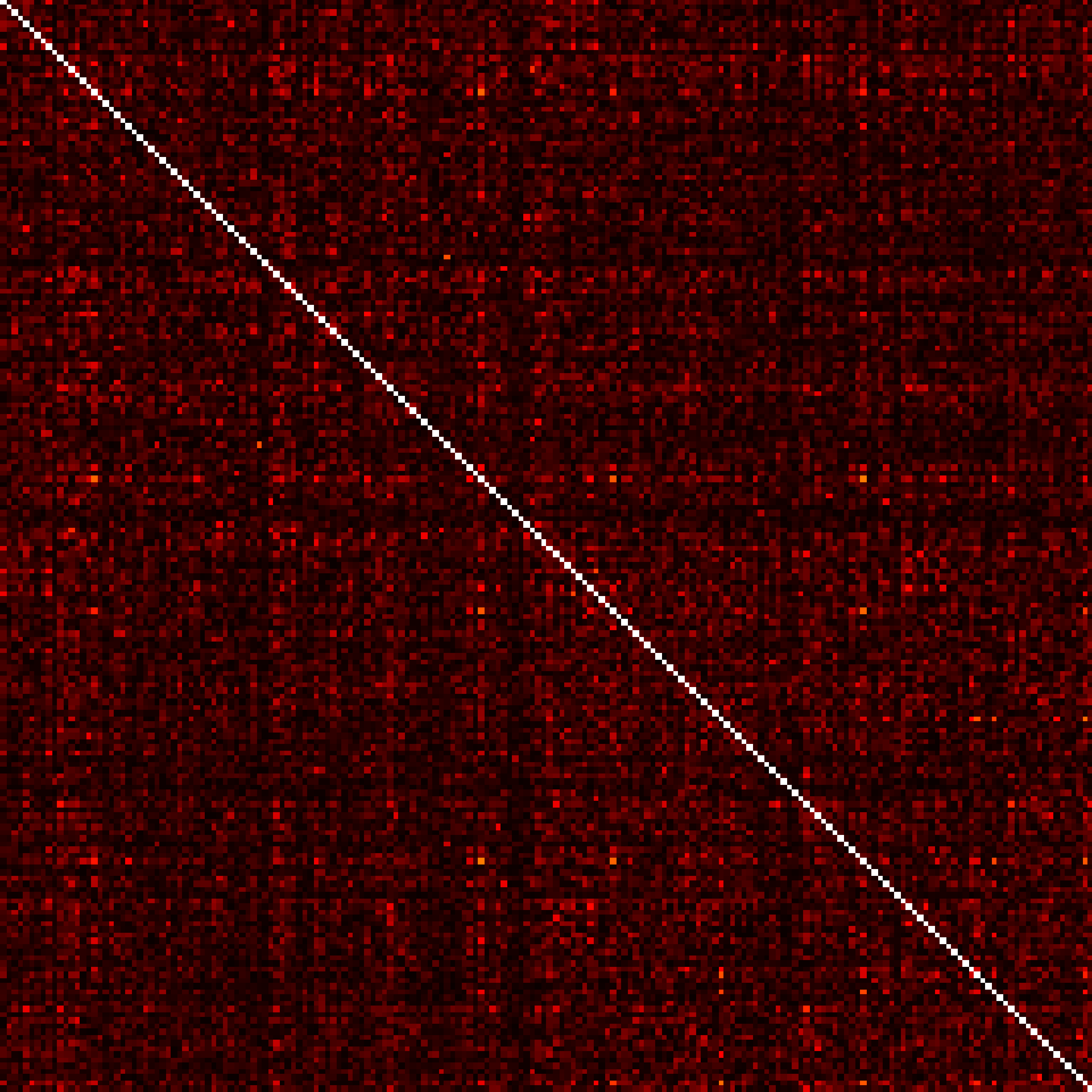}}
~
    \covarlabels{conv3b}{192}{conv3b}{192}{\includegraphics[width=0.11\textwidth]{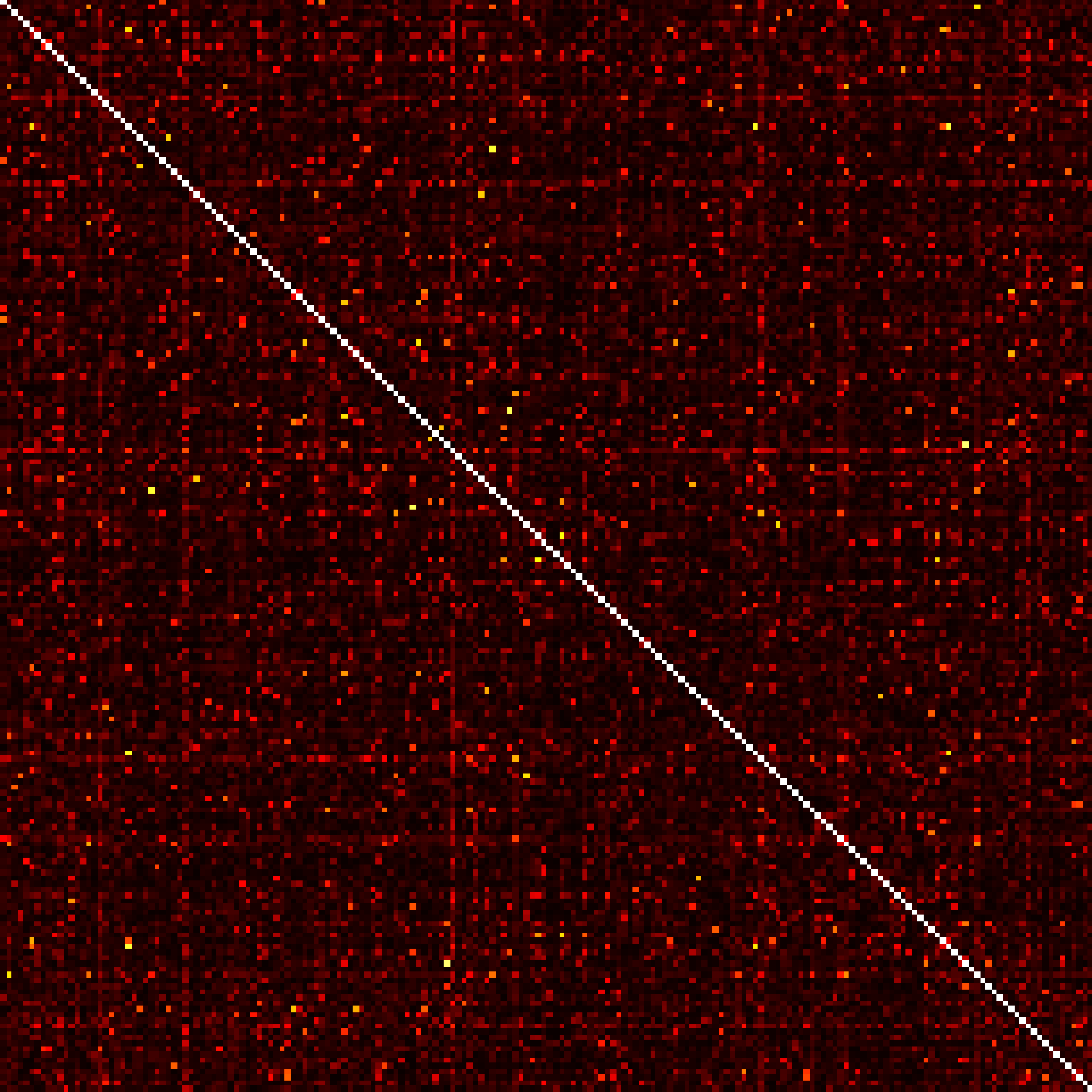}}
\caption{\textbf{Root-4.}}
\vspace*{0.6em}
\label{fig:corrroot4}
\end{subfigure}
~
\begin{subfigure}[c]{\paperwidth}
\centering
    \covarlabels{conv1a}{192}{conv1a}{192}{\includegraphics[width=0.11\textwidth]{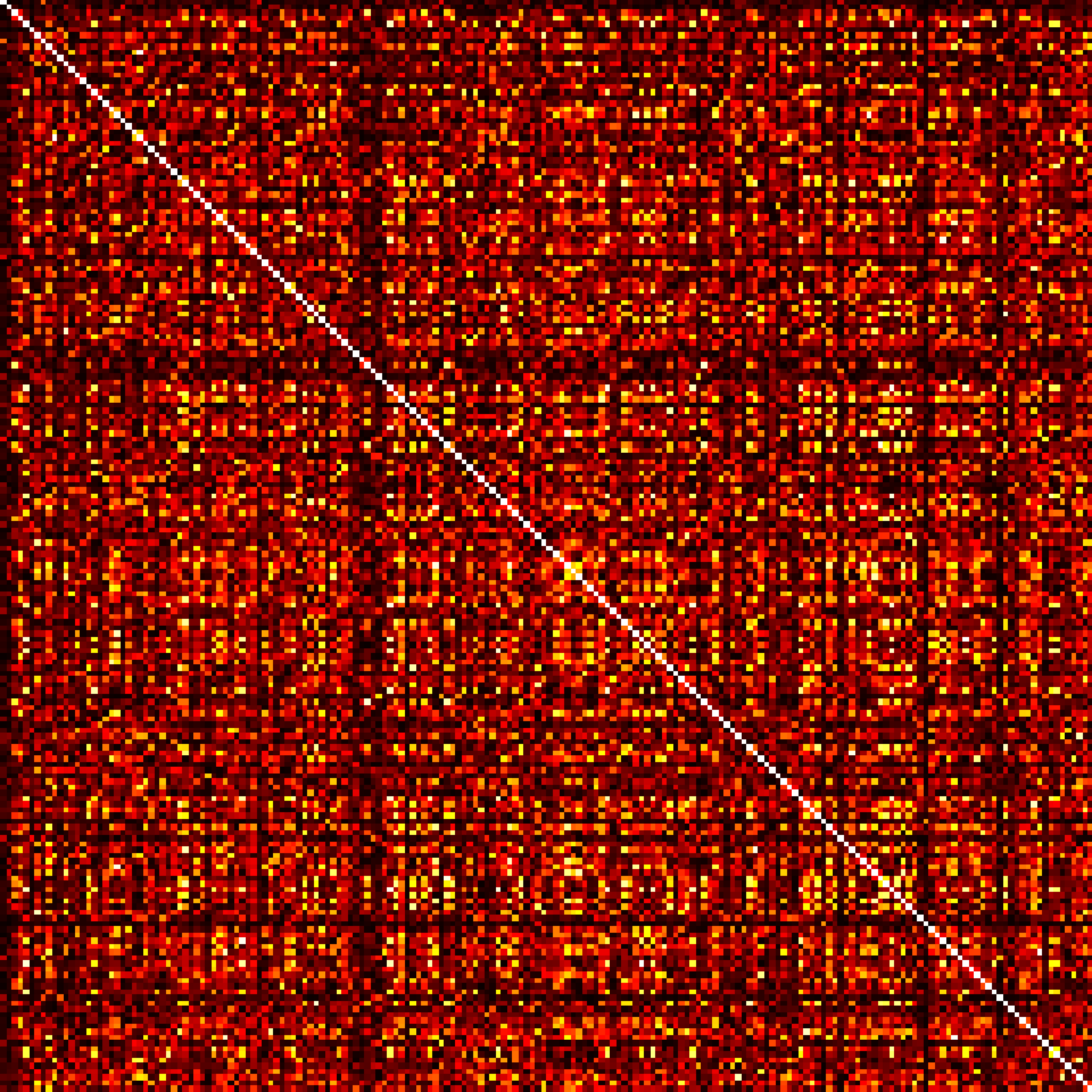}}
~
    \covarlabels{conv1b}{160}{conv1b}{160}{\includegraphics[width=0.11\textwidth]{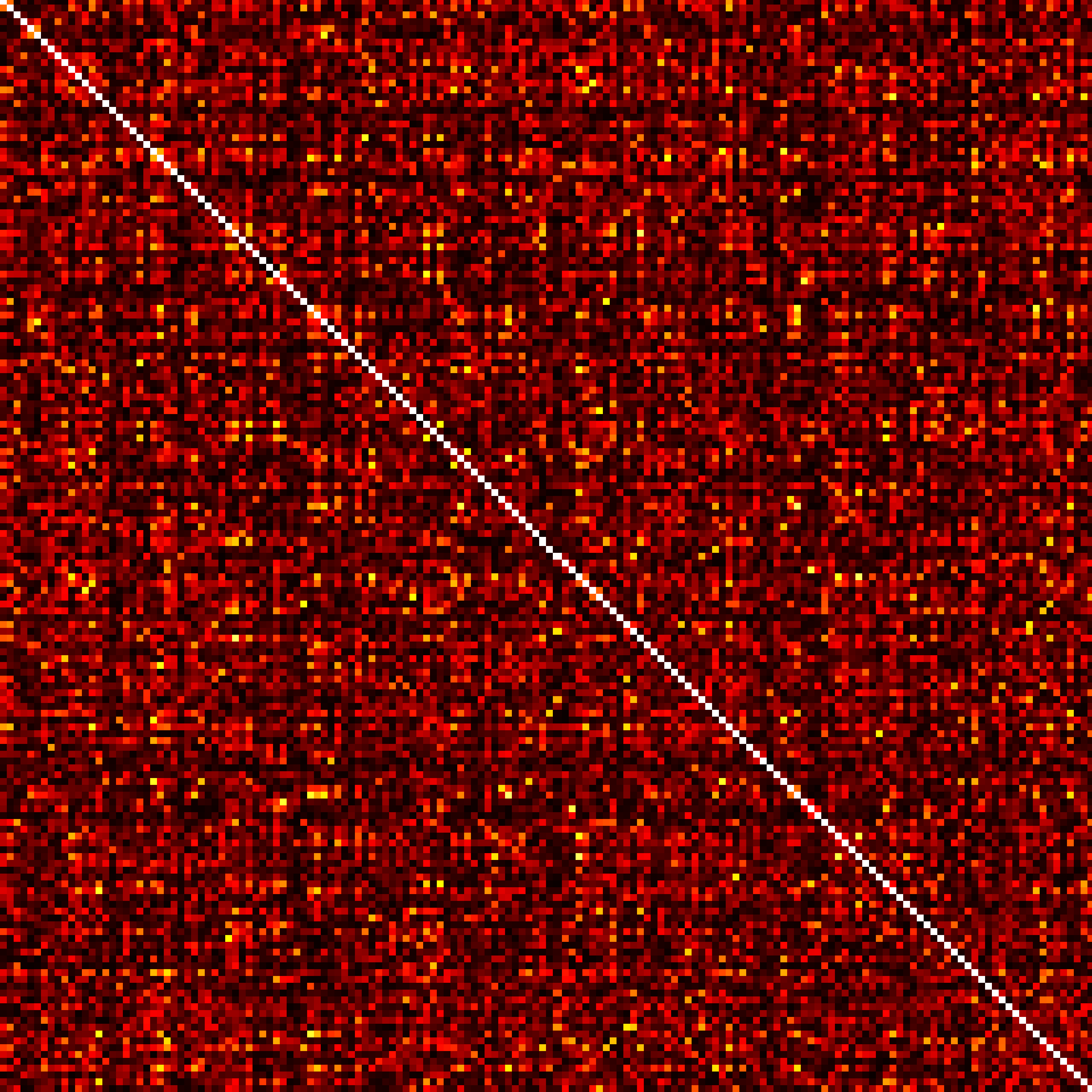}}
~
    \covarlabels{conv1c}{96}{conv1c}{96}{\includegraphics[width=0.11\textwidth]{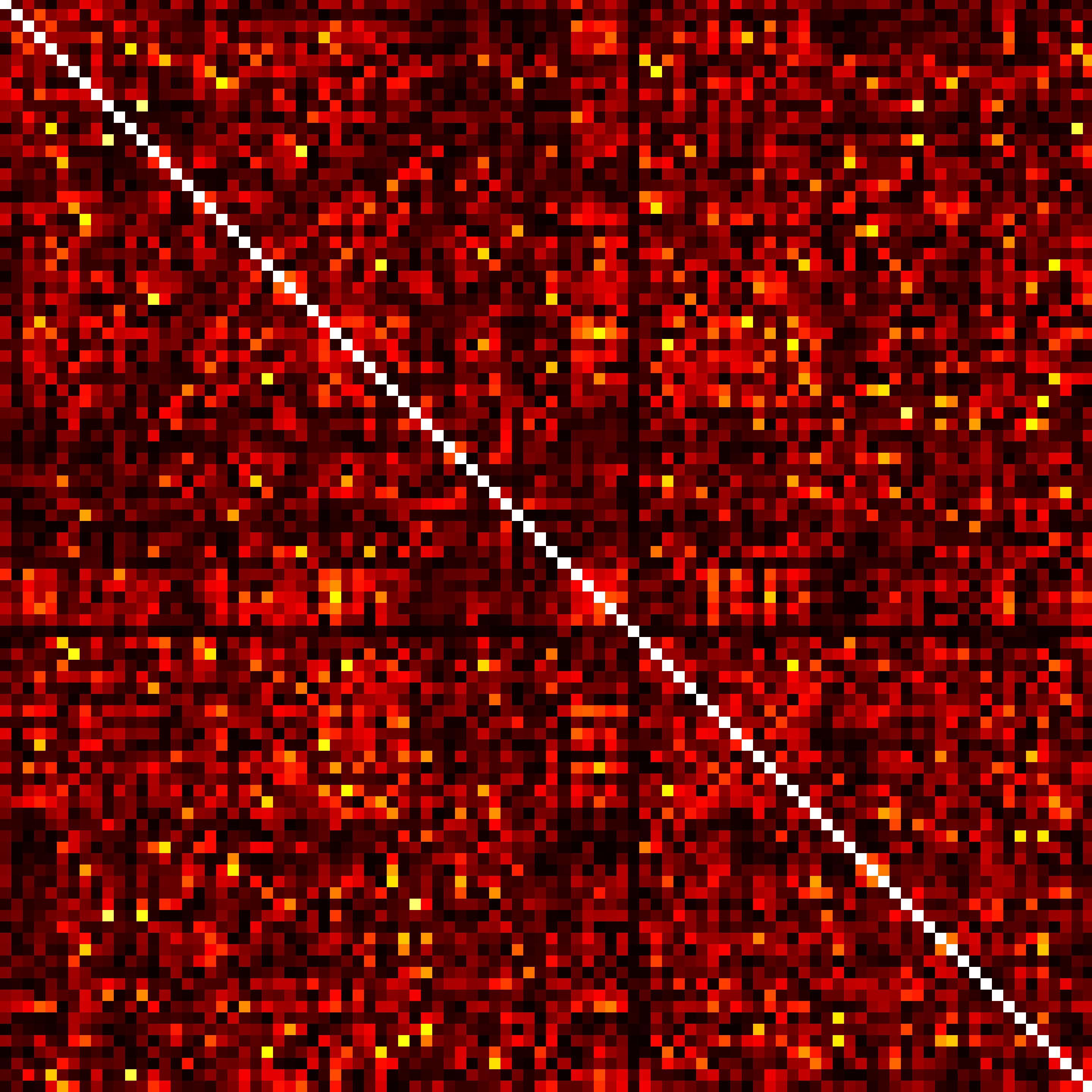}}
~
    \covarlabels{conv2a}{192}{conv2a}{192}{\includegraphics[width=0.11\textwidth]{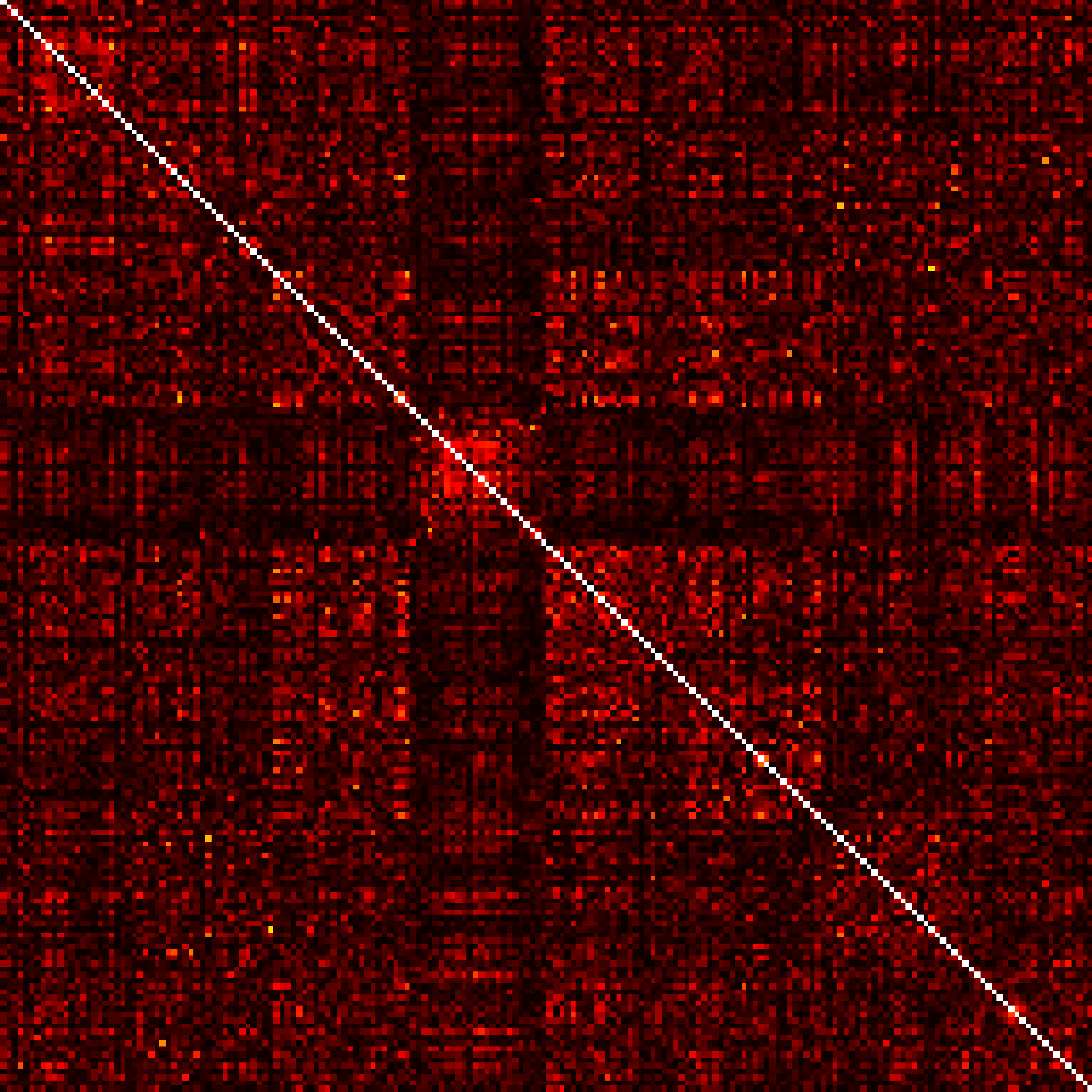}}
~
    \covarlabels{conv2b}{192}{conv2b}{192}{\includegraphics[width=0.11\textwidth]{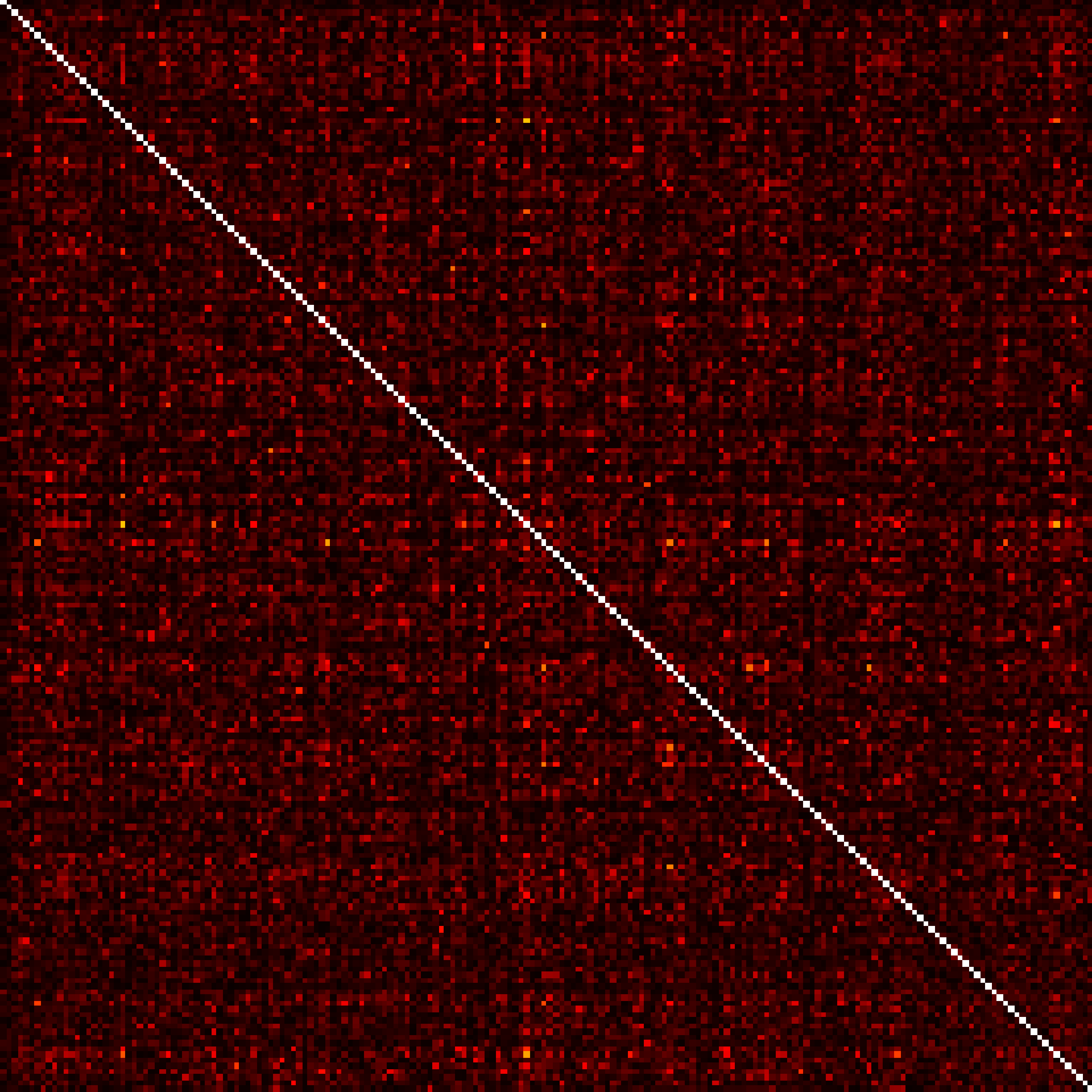}}
~
    \covarlabels{conv2c}{192}{conv2c}{192}{\includegraphics[width=0.11\textwidth]{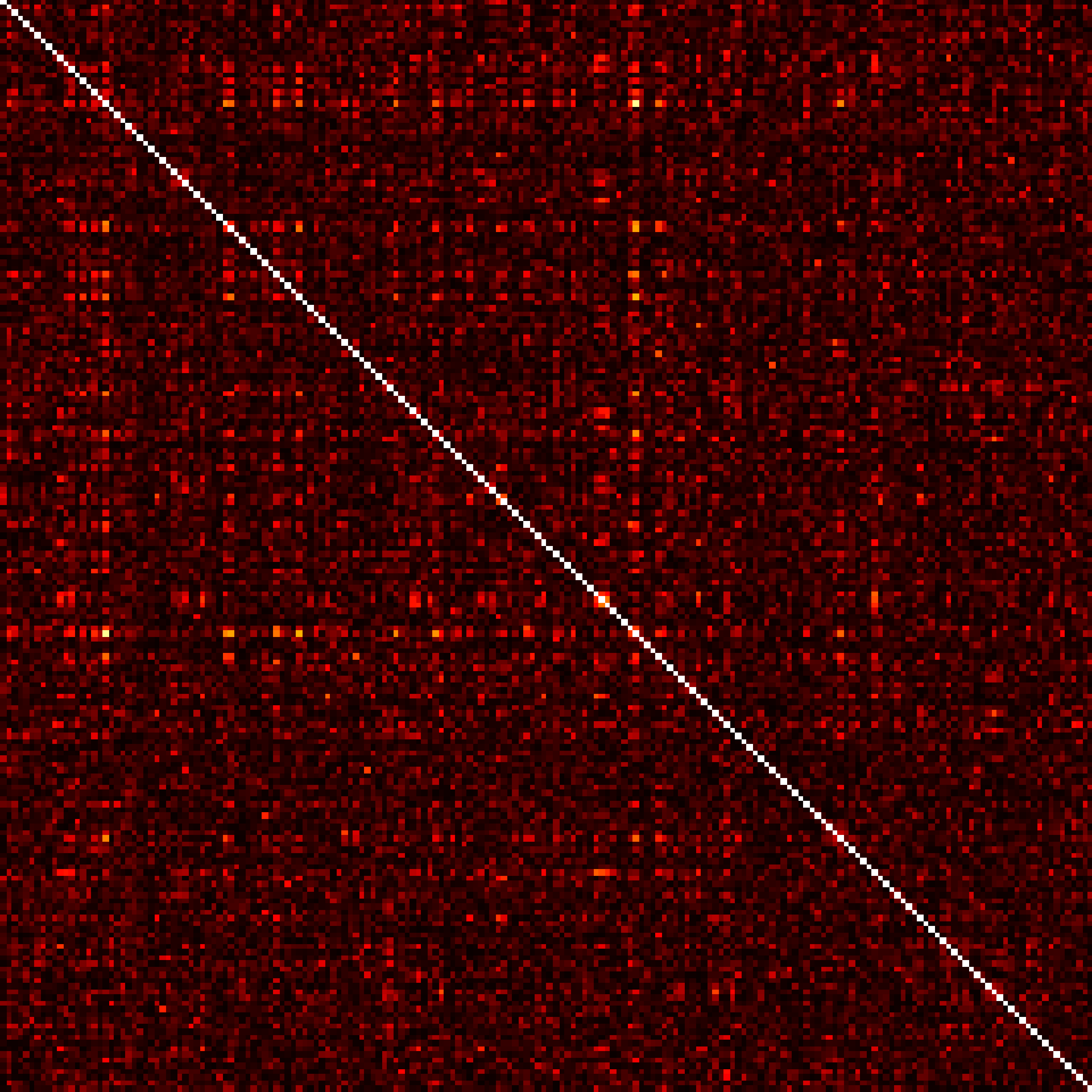}}
~
    \covarlabels{conv3a}{192}{conv3a}{192}{\includegraphics[width=0.11\textwidth]{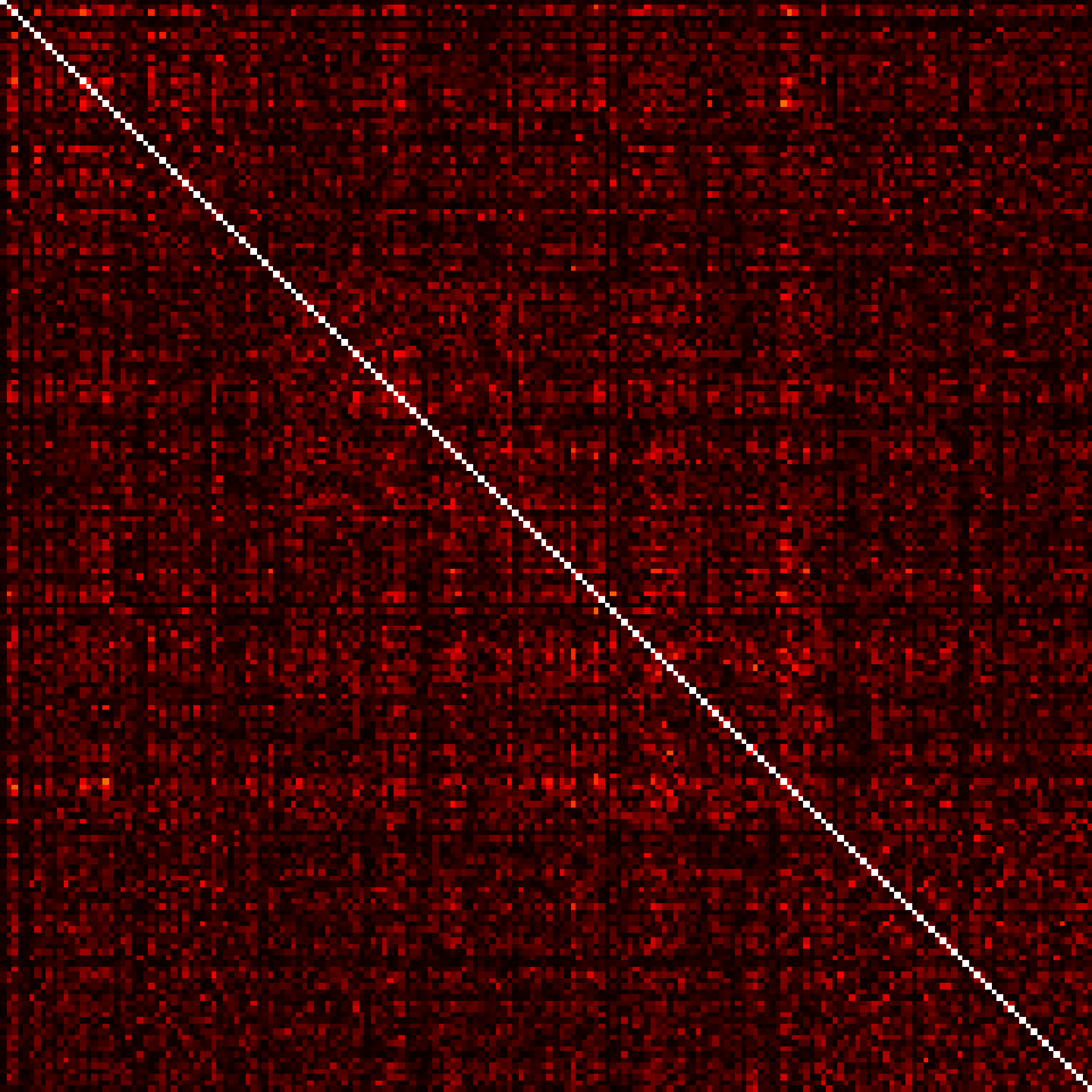}}
~
    \covarlabels{conv3b}{192}{conv3b}{192}{\includegraphics[width=0.11\textwidth]{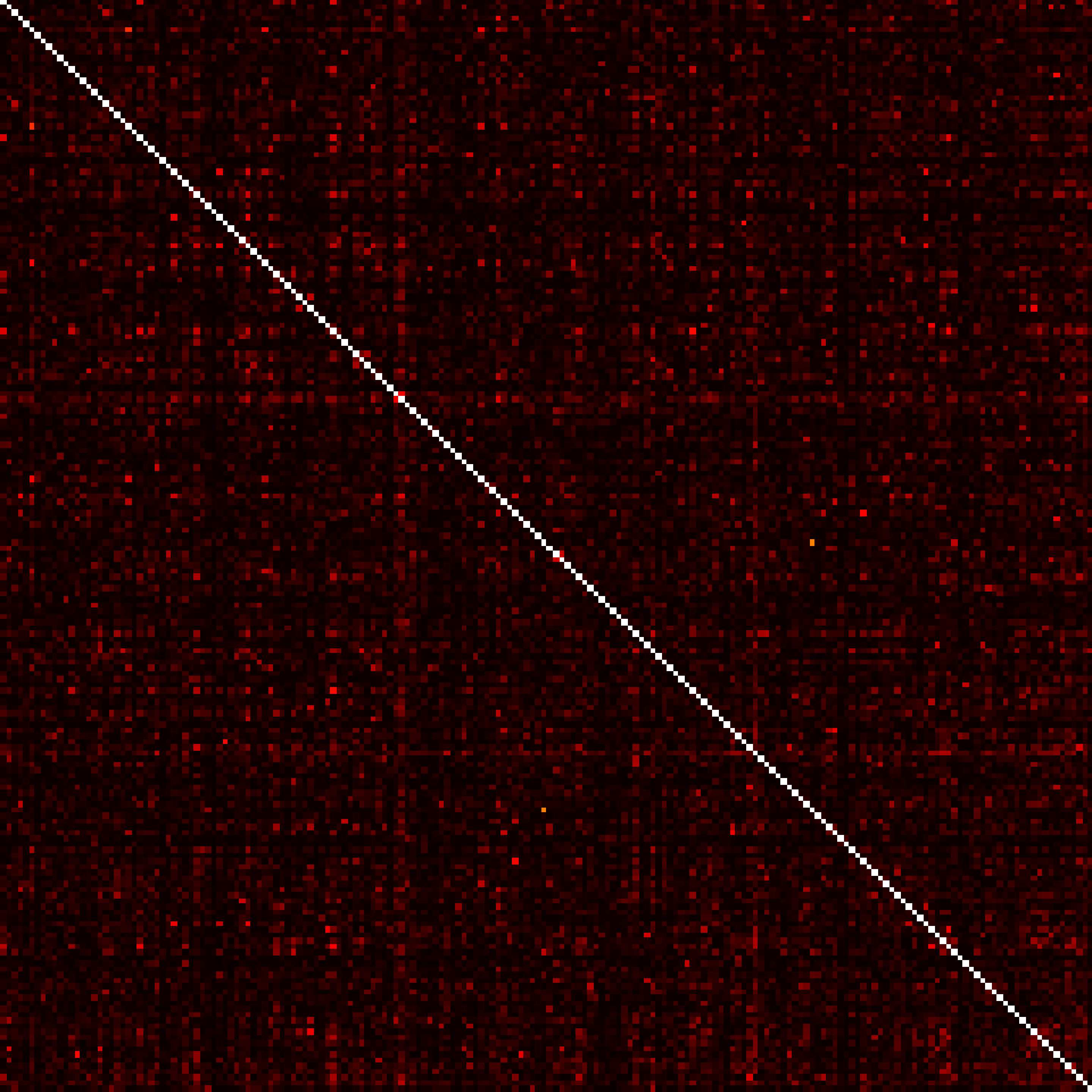}}
\caption{\textbf{Root-8.}}
\vspace*{0.6em}
\label{fig:corrroot8}
\end{subfigure}
~
\begin{subfigure}[c]{\paperwidth}
\centering
    \covarlabels{conv1a}{192}{conv1a}{192}{\includegraphics[width=0.11\textwidth]{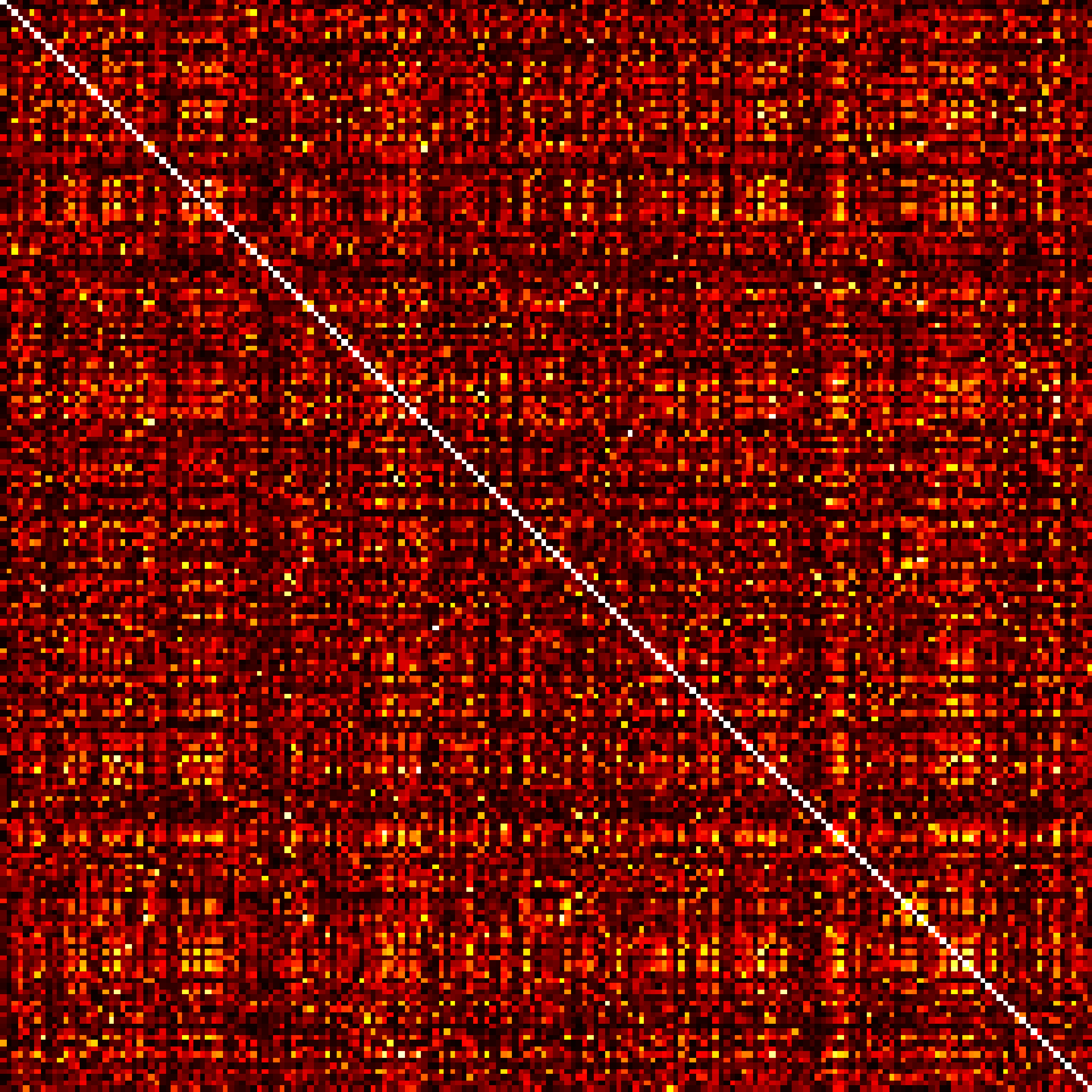}}
~
    \covarlabels{conv1b}{160}{conv1b}{160}{\includegraphics[width=0.11\textwidth]{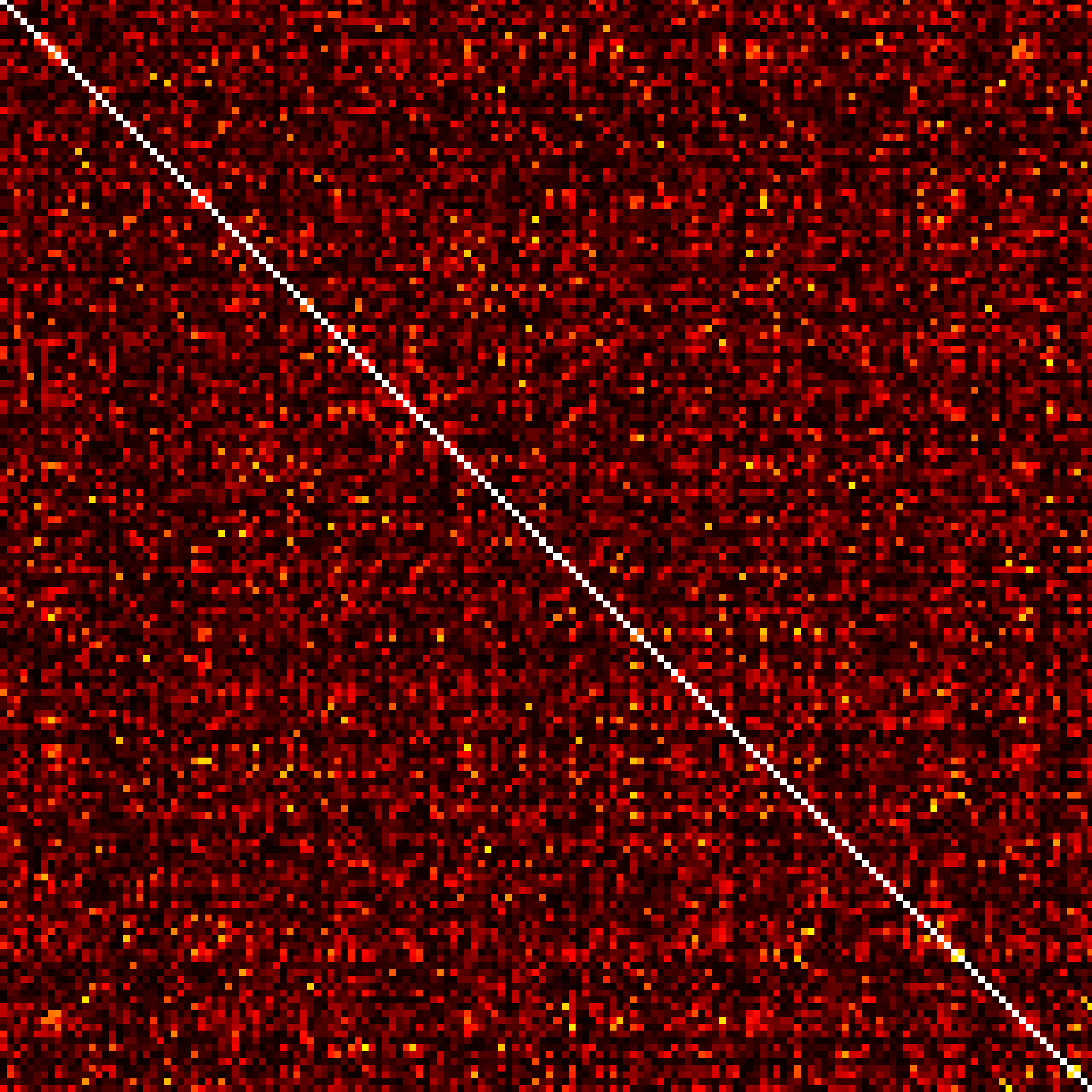}}
~
    \covarlabels{conv1c}{96}{conv1c}{96}{\includegraphics[width=0.11\textwidth]{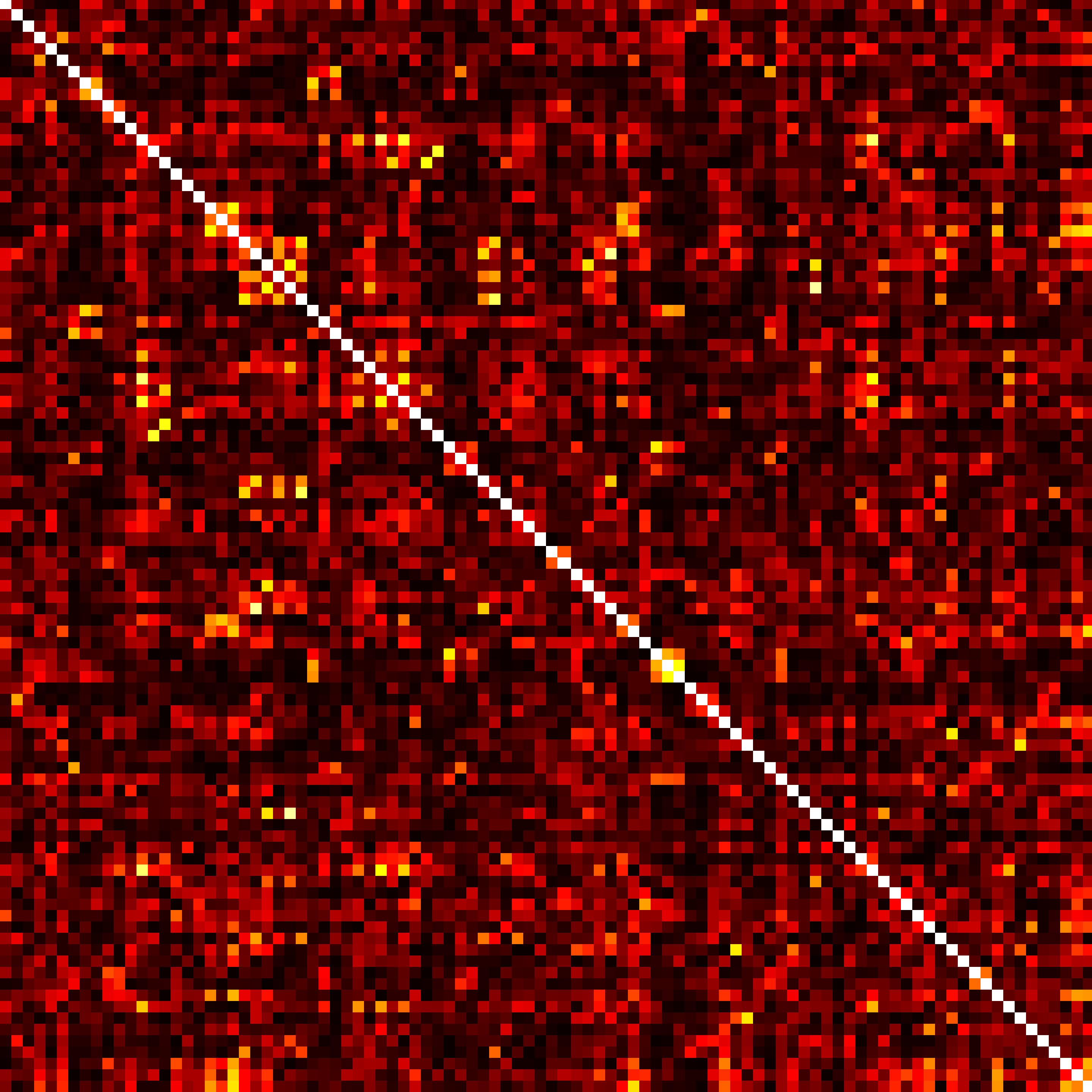}}
~
    \covarlabels{conv2a}{192}{conv2a}{192}{\includegraphics[width=0.11\textwidth]{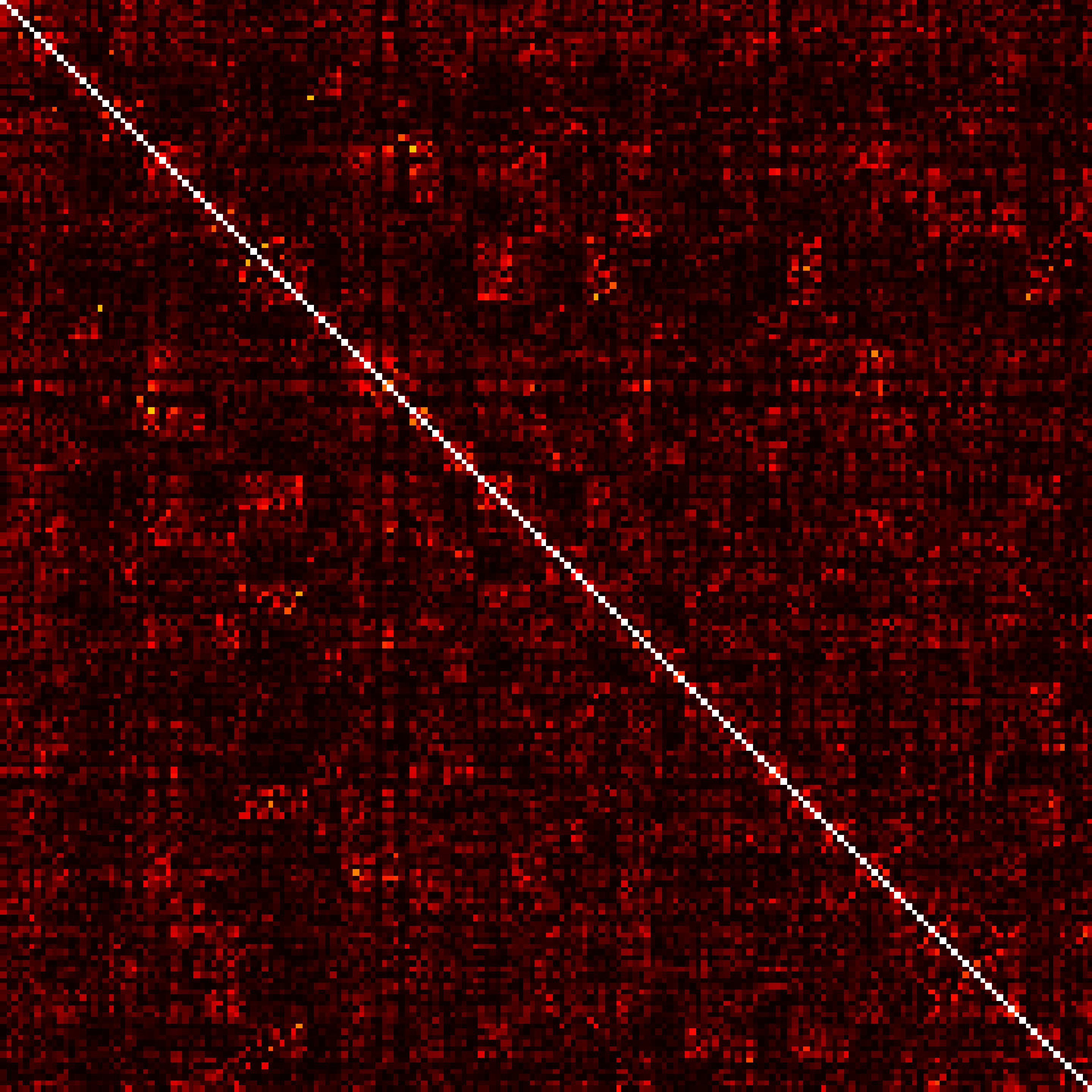}}
~
    \covarlabels{conv2b}{192}{conv2b}{192}{\includegraphics[width=0.11\textwidth]{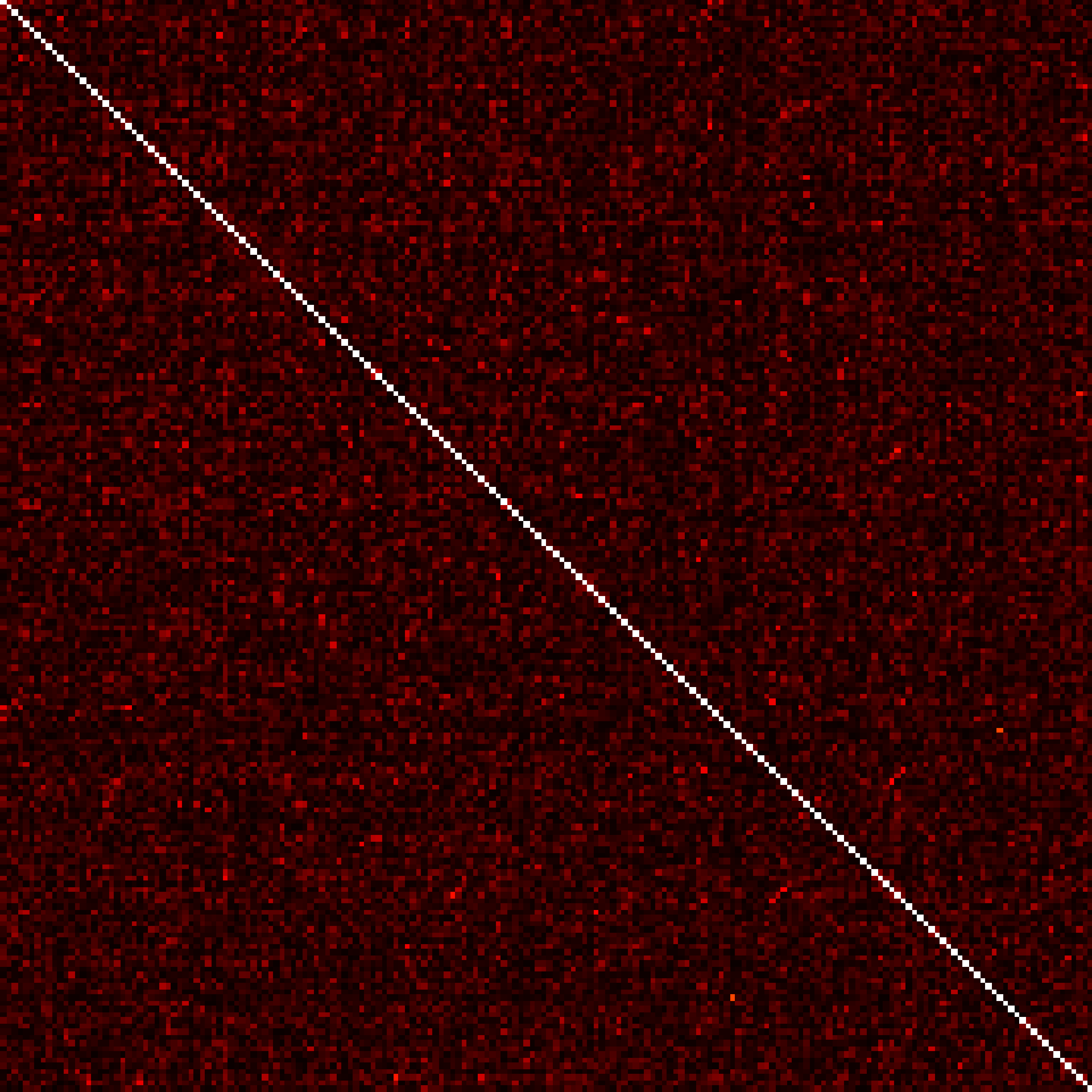}}
~
    \covarlabels{conv2c}{192}{conv2c}{192}{\includegraphics[width=0.11\textwidth]{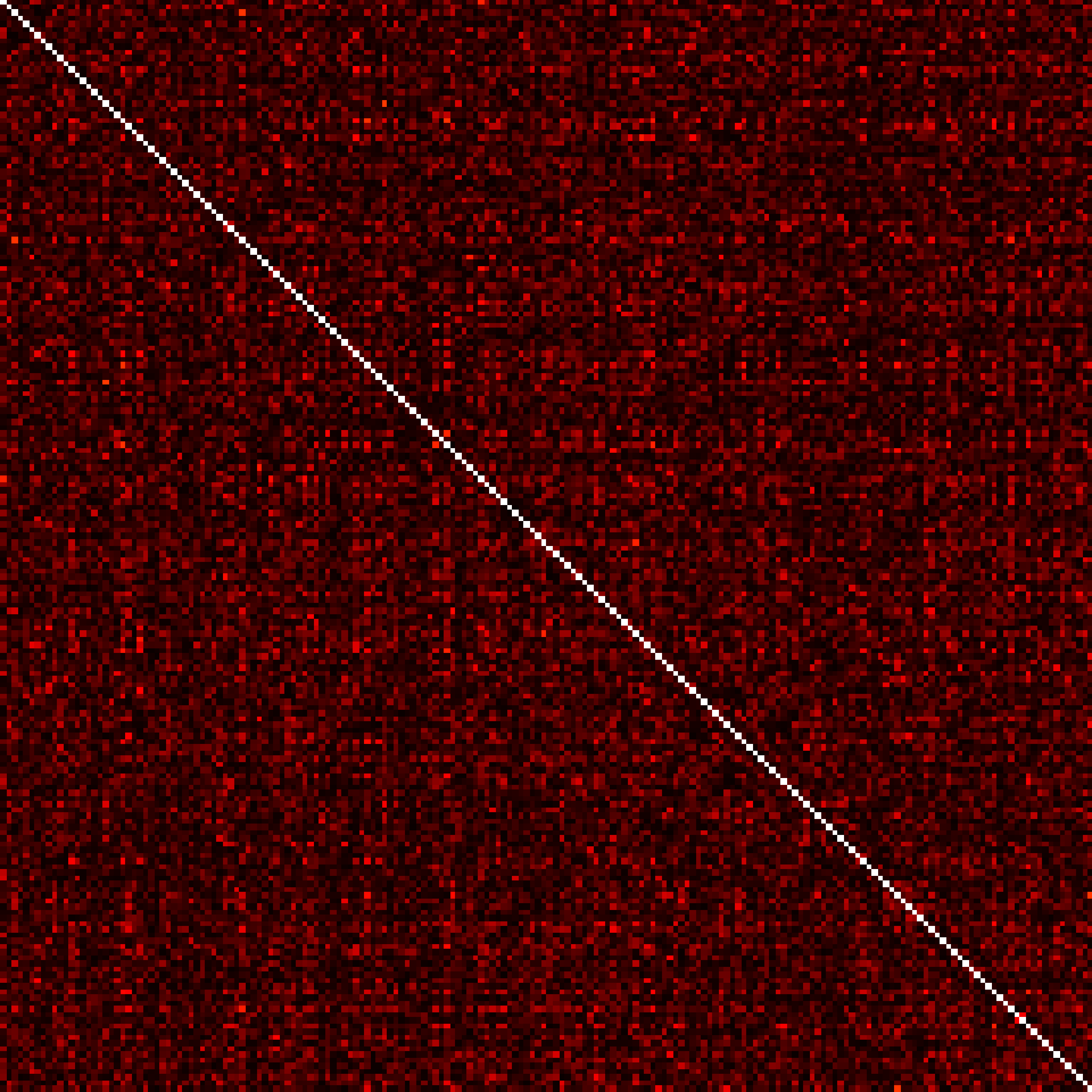}}
~
    \covarlabels{conv3a}{192}{conv3a}{192}{\includegraphics[width=0.11\textwidth]{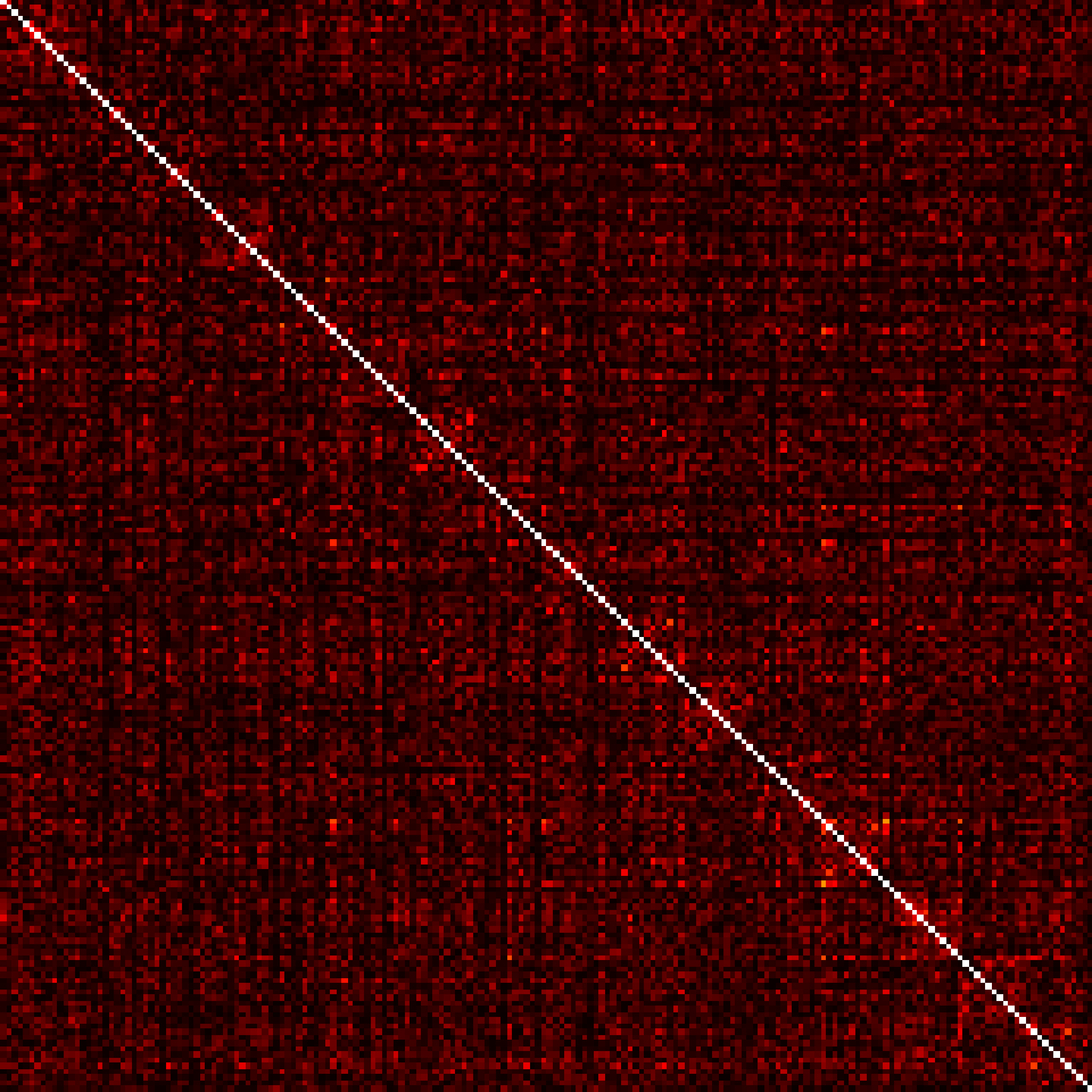}}
~
    \covarlabels{conv3b}{192}{conv3b}{192}{\includegraphics[width=0.11\textwidth]{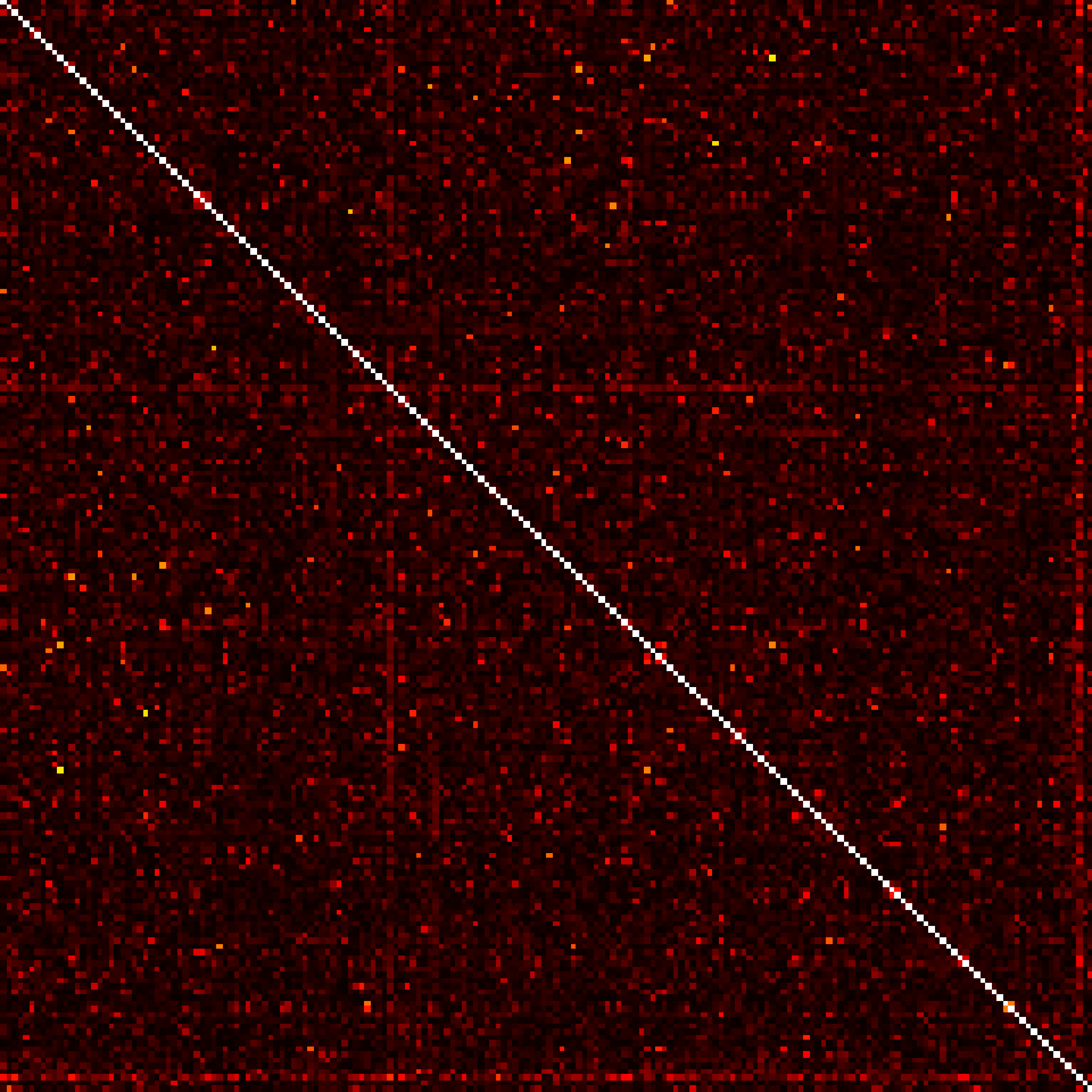}}
\caption{\textbf{Root-32.}}
\label{fig:corrroot32}
\end{subfigure}
\centering
\includegraphics[width=0.4\linewidth]{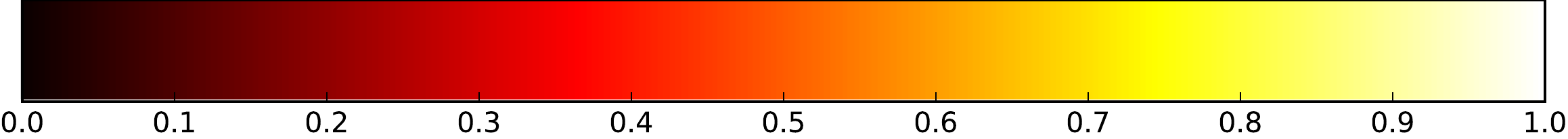}
\caption{\textbf{Network-in-Network Intra-Layer Correlation.} Absolute Correlation of filters within each layer of a NiN model variant.}
\label{fig:nincorr}
\end{figure}
\end{landscape}
\begin{landscape}
\begin{figure}[p]
\centering
\begin{subfigure}[b]{0.32\linewidth}
\centering
    \covarlabels{conv2c}{192}{conv3a}{192}{\includegraphics[width=\textwidth]{figs/msrc-cifar-nin-4pad-conv8-corr}}
    \caption{Standard}
    \label{fig:normalcovartestfull}
\end{subfigure}
~
\begin{subfigure}[b]{0.32\linewidth}
\centering
    \covarlabels{conv2c}{192}{conv3a}{192}{\includegraphics[width=\linewidth]{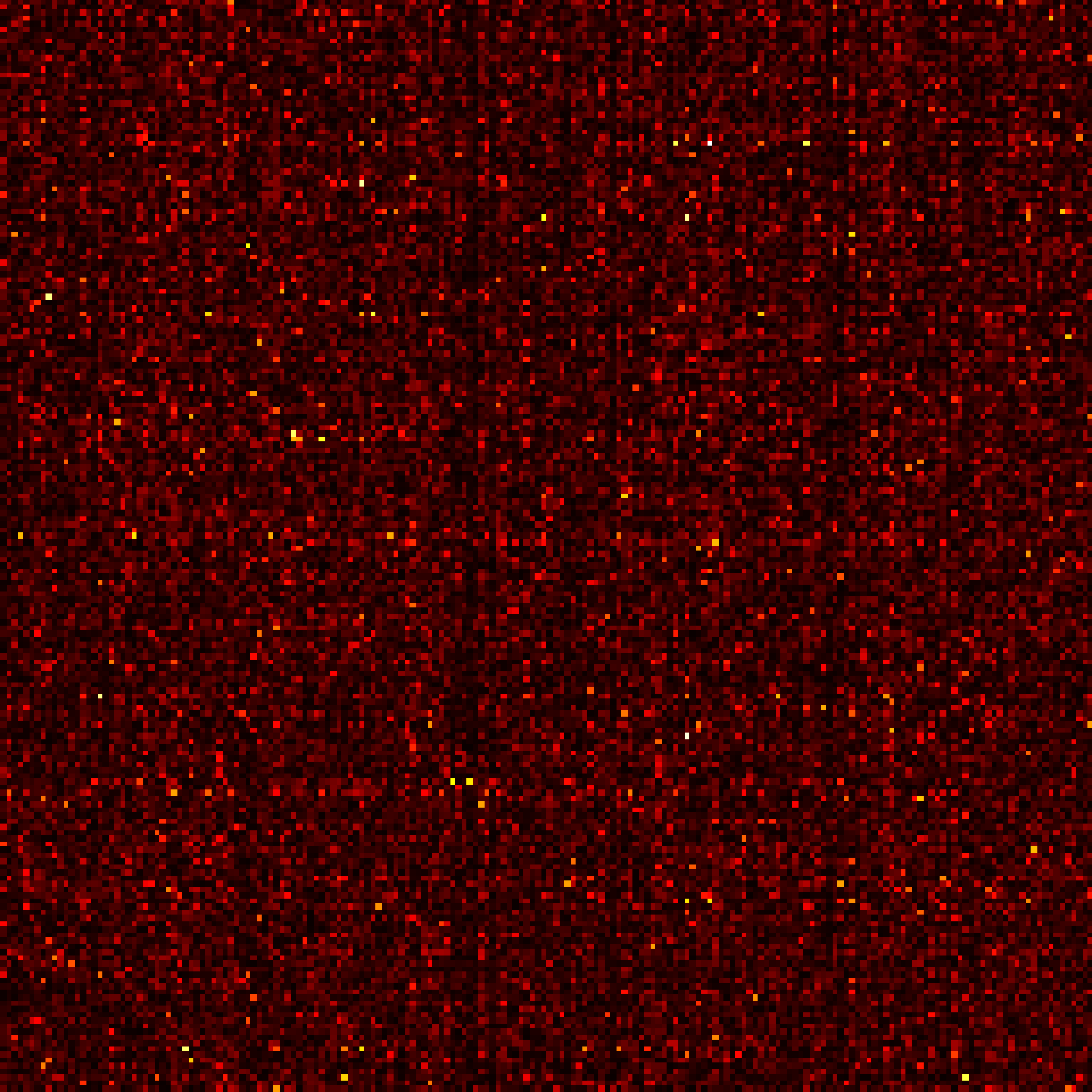}}
    \caption{Root-2}
    \label{fig:root2corrfull}
\end{subfigure}
~
\begin{subfigure}[b]{0.32\linewidth}
\centering
    \covarlabels{conv2c}{192}{conv3a}{192}{\includegraphics[width=\linewidth]{figs/msrc-cifar-nin-4pad-funnel4-convonly-conv8-corr}}
    \caption{Root-4}
    \label{fig:root4corrfull}
\end{subfigure}
~
\begin{subfigure}[b]{0.32\linewidth}
\centering
    \vspace*{0.6em}
    \covarlabels{conv2c}{192}{conv3a}{192}{\includegraphics[width=\linewidth]{figs/msrc-cifar-nin-4pad-funnel8-convonly-conv8-corr}}
    \caption{Root-8}
    \label{fig:root8corrfull}
\end{subfigure}
~
\begin{subfigure}[b]{0.32\linewidth}
\centering
    \vspace*{0.6em}
    \covarlabels{conv2c}{192}{conv3a}{192}{\includegraphics[width=\linewidth]{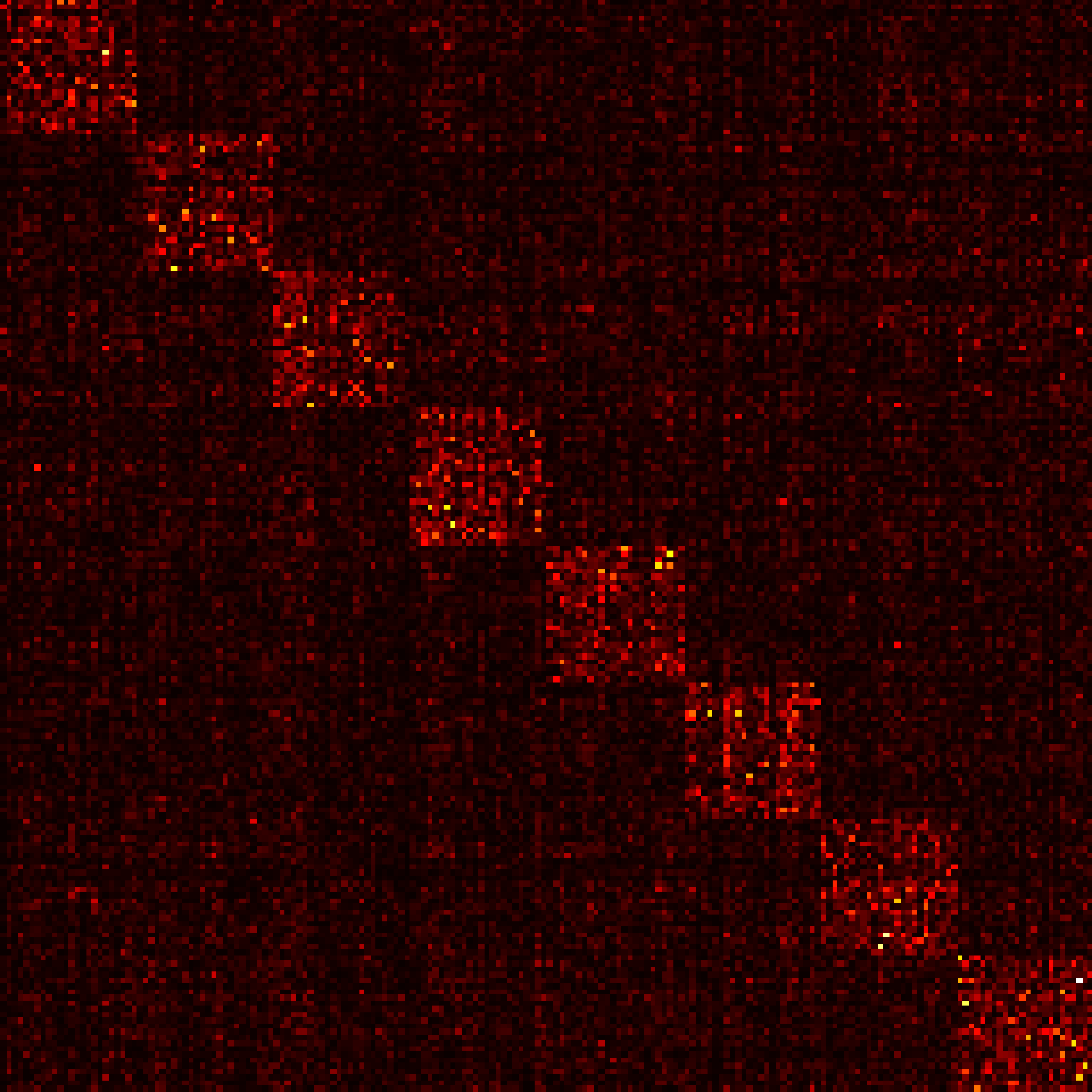}}
    \caption{Root-16}
    \label{fig:root16corrfull}
\end{subfigure}
~
\begin{subfigure}[b]{0.32\linewidth}
\centering
    \vspace*{0.6em}
    \covarlabels{conv2c}{192}{conv3a}{192}{\includegraphics[width=\linewidth]{figs/msrc-cifar-nin-4pad-funnel32-convonly-conv8-corr}}
    \caption{Root-32}
    \label{fig:root32corrfull}
\end{subfigure}
\caption{\textbf{Filter Inter-layer Covariance conv2c--conv3a.} The block-diagonal sparsity learned by a root-unit is visible in the correlation of filters on layers \texttt{conv3a} and \texttt{conv2c} in the NiN network.}
\label{fig:covarfull}
\end{figure}
\end{landscape}

\begin{landscape}
\begin{figure}[p]
\begin{subfigure}[c]{\paperwidth}
\centering
    \covarlabels{conv1a}{192}{conv1b}{160}{\includegraphics[width=0.1\textwidth]{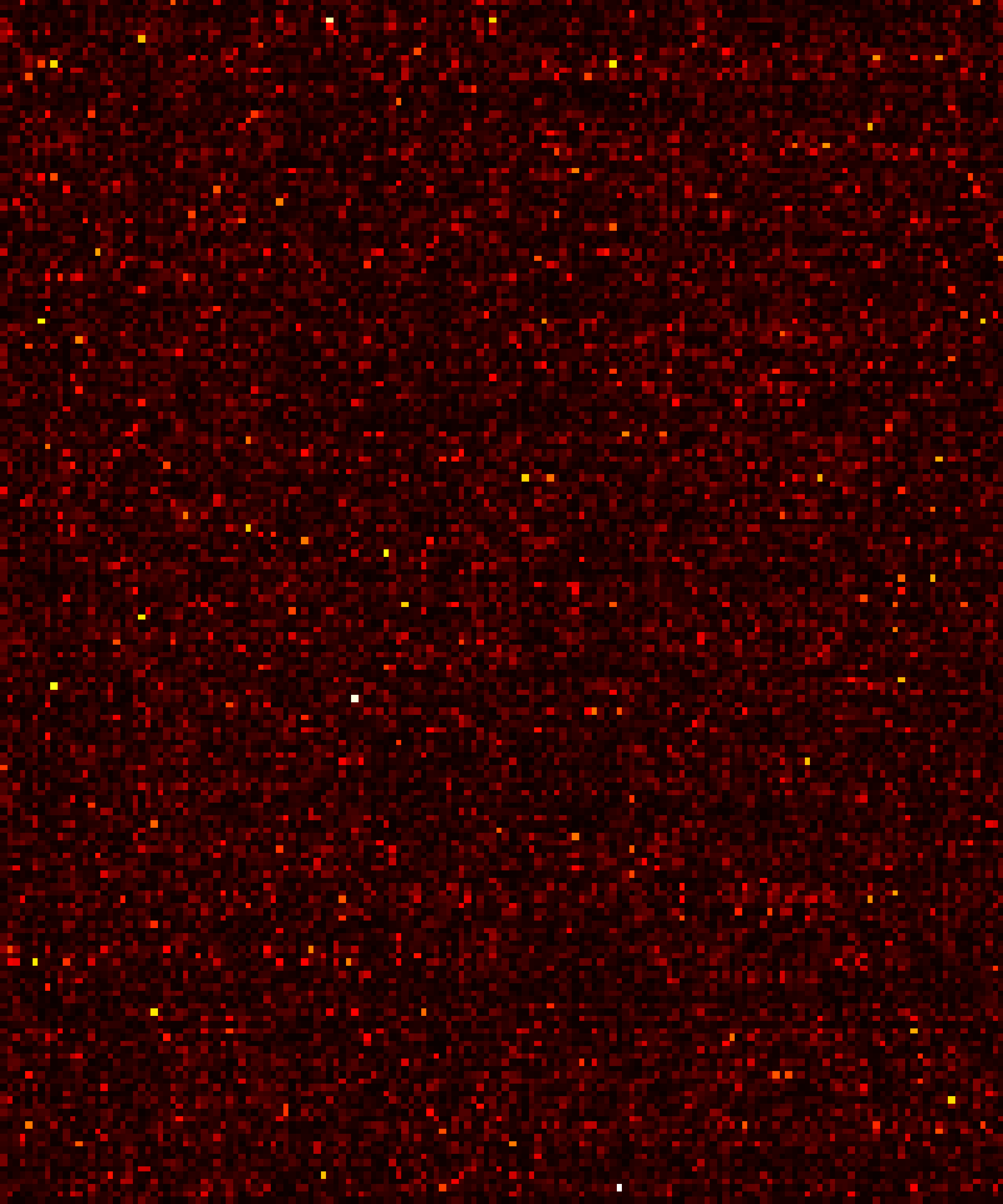}}
~
    \covarlabels{conv1b}{160}{conv1c}{96}{\includegraphics[width=0.05\textwidth]{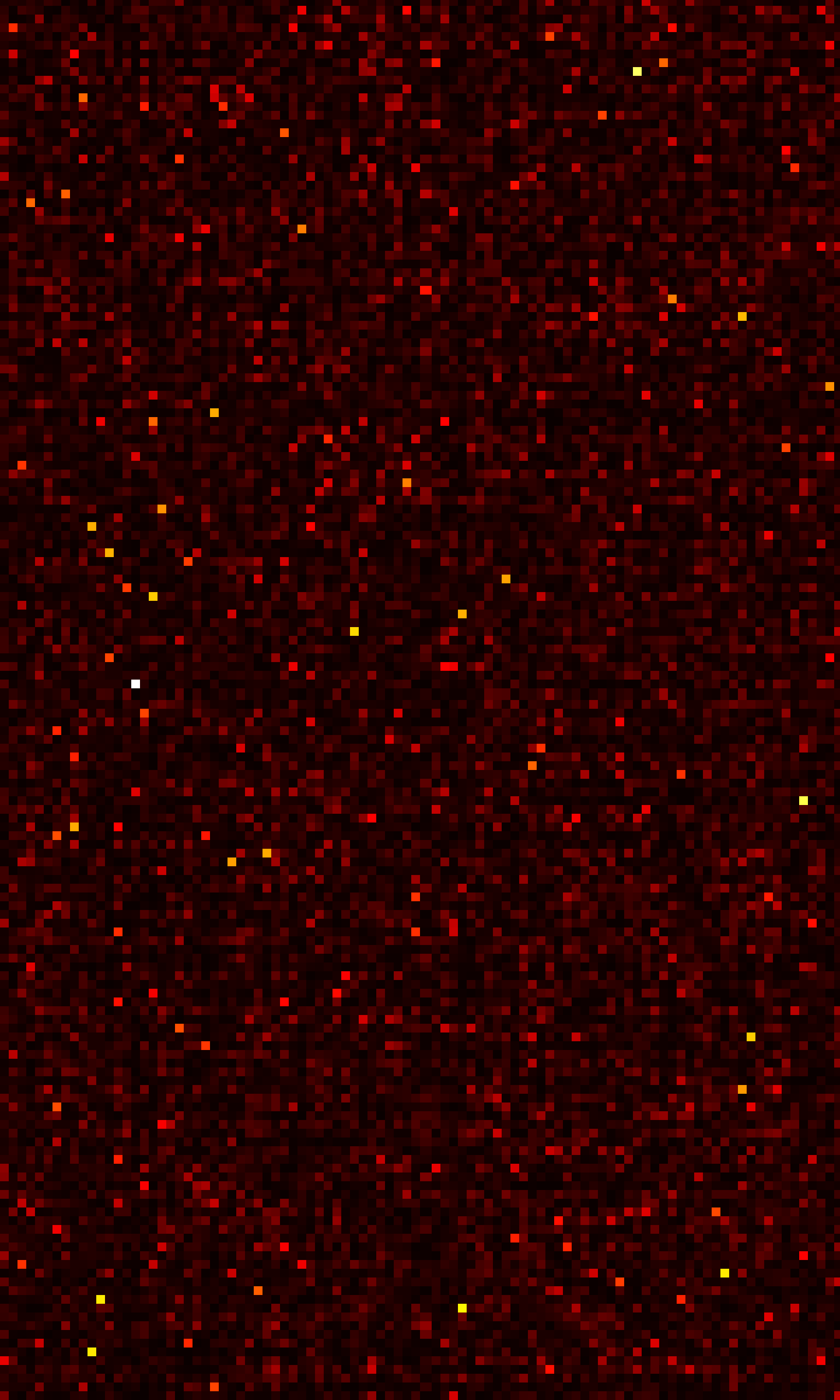}}
~
    \covarlabels{conv1c}{96}{conv2a}{192}{\includegraphics[width=0.12\textwidth]{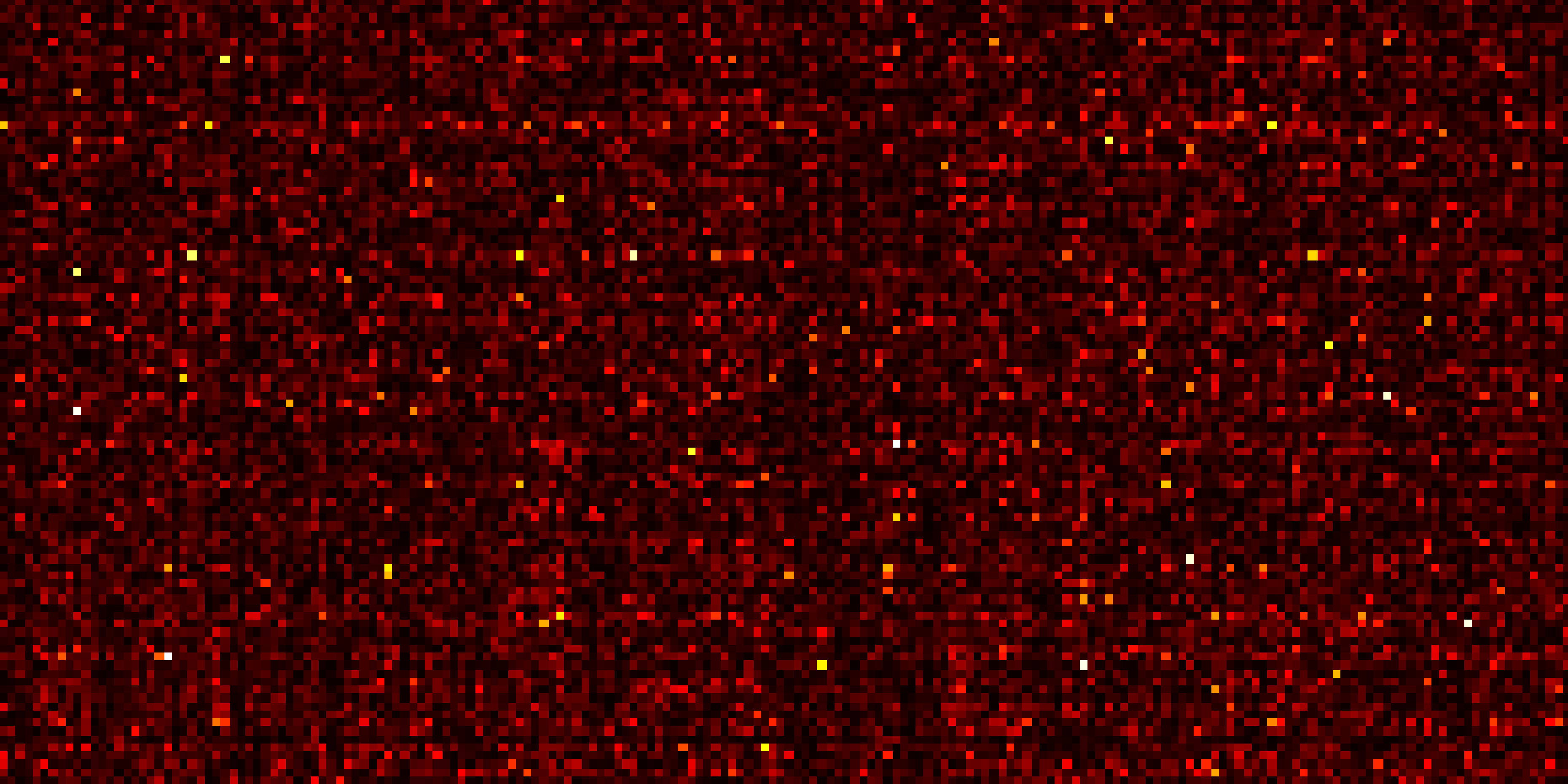}}
~
    \covarlabels{conv2a}{192}{conv2b}{192}{\includegraphics[width=0.12\textwidth]{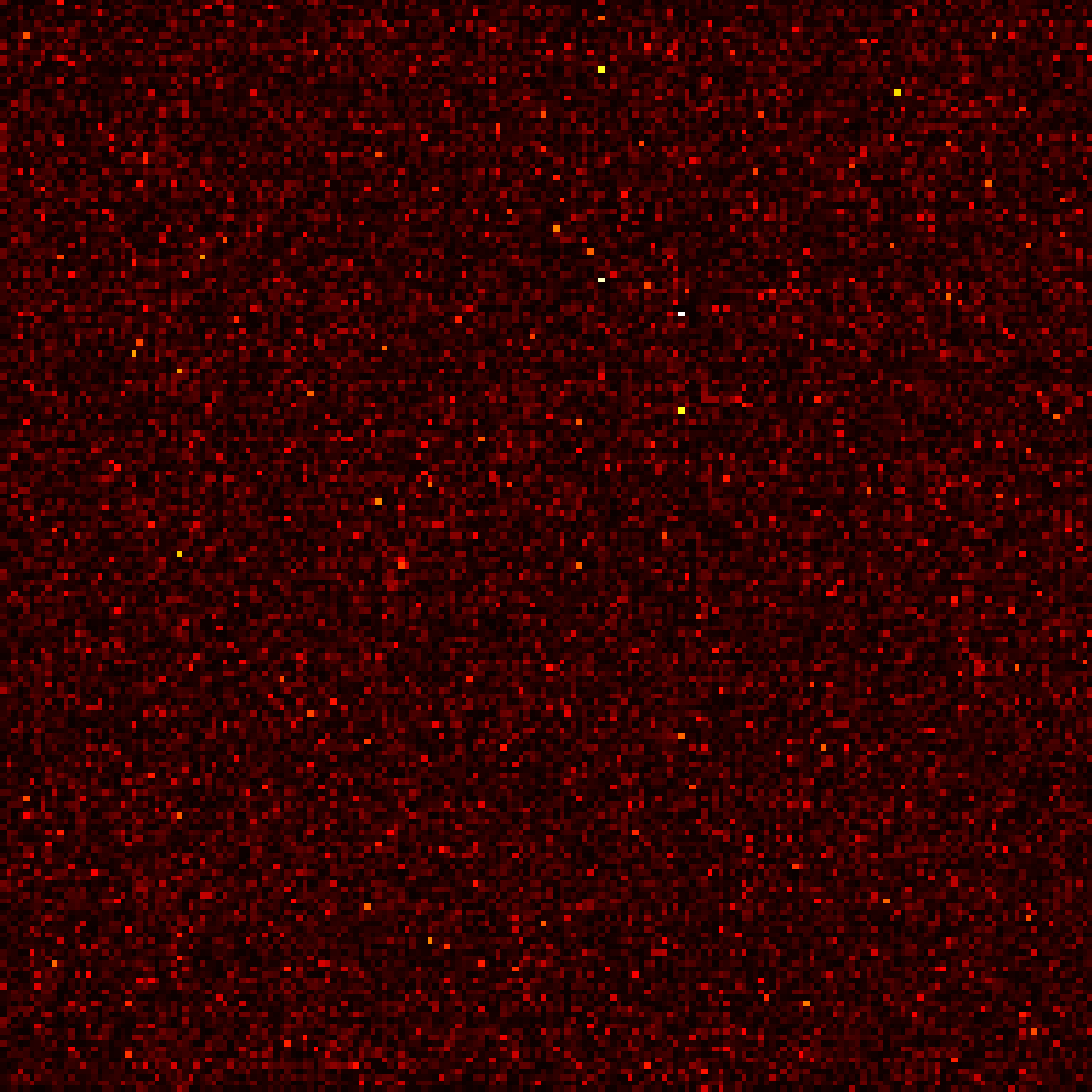}}
~
    \covarlabels{conv2b}{192}{conv2c}{192}{\includegraphics[width=0.12\textwidth]{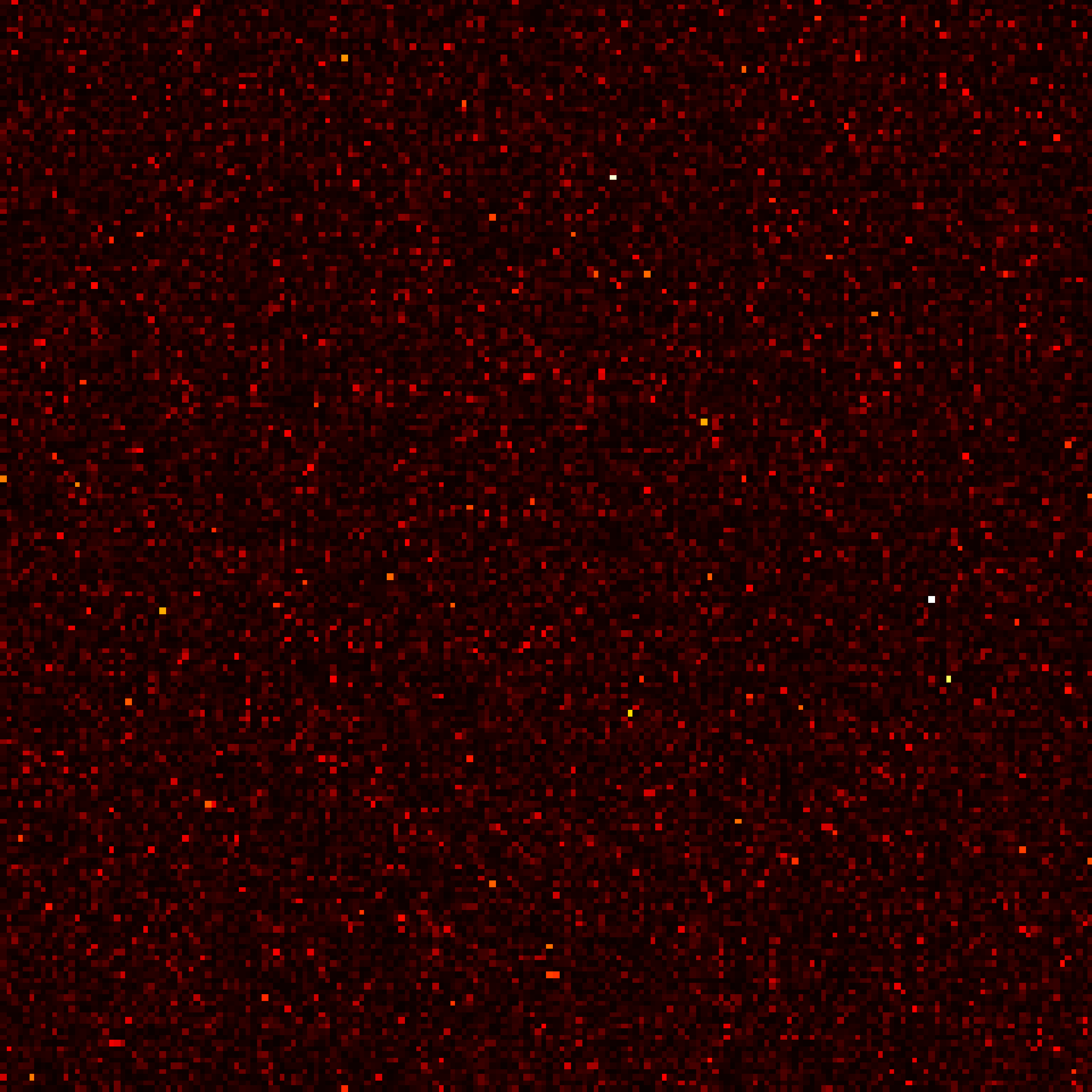}}
~
    \covarlabels{conv2c}{192}{conv3a}{192}{\includegraphics[width=0.12\textwidth]{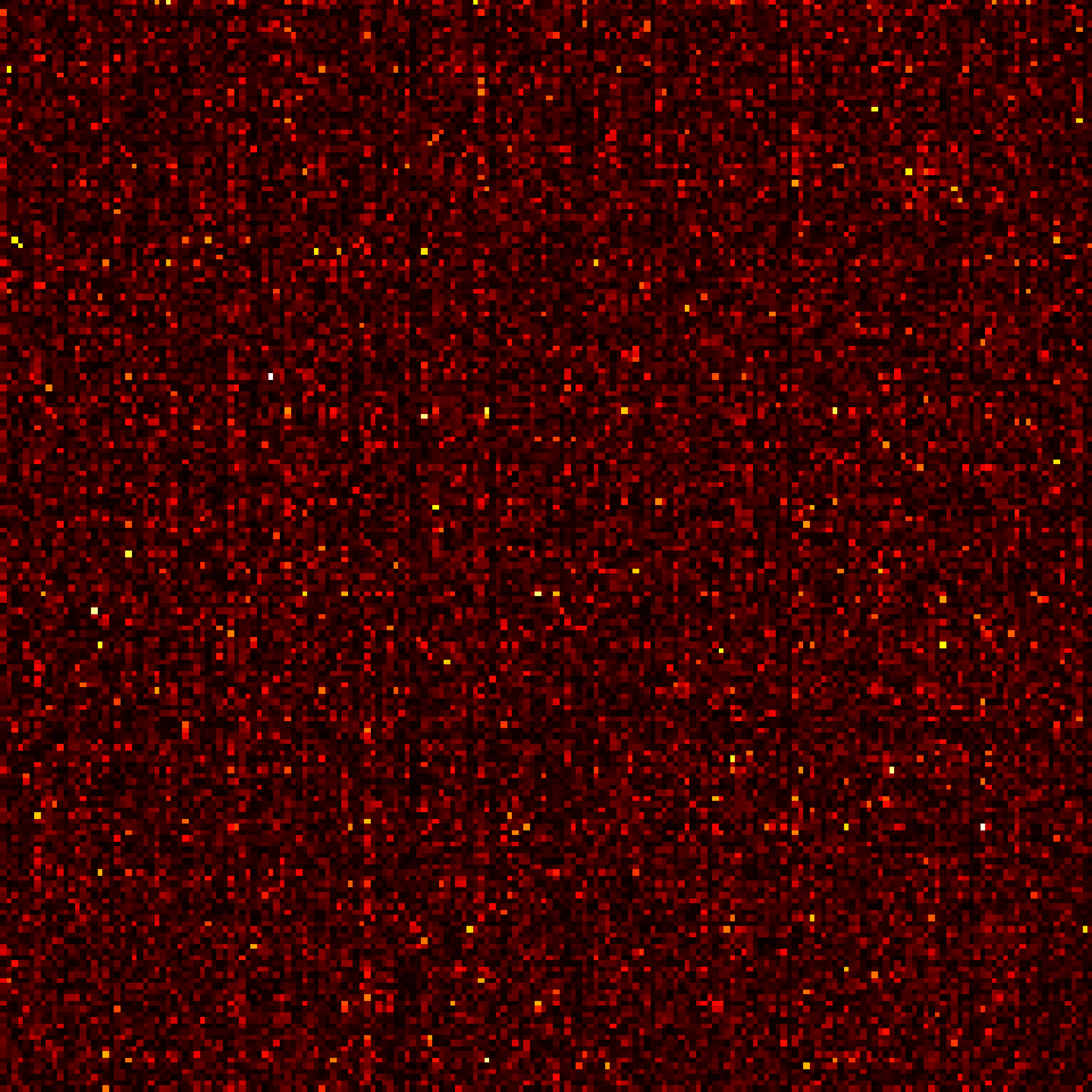}}
~
    \covarlabels{conv3a}{192}{conv3b}{192}{\includegraphics[width=0.12\textwidth]{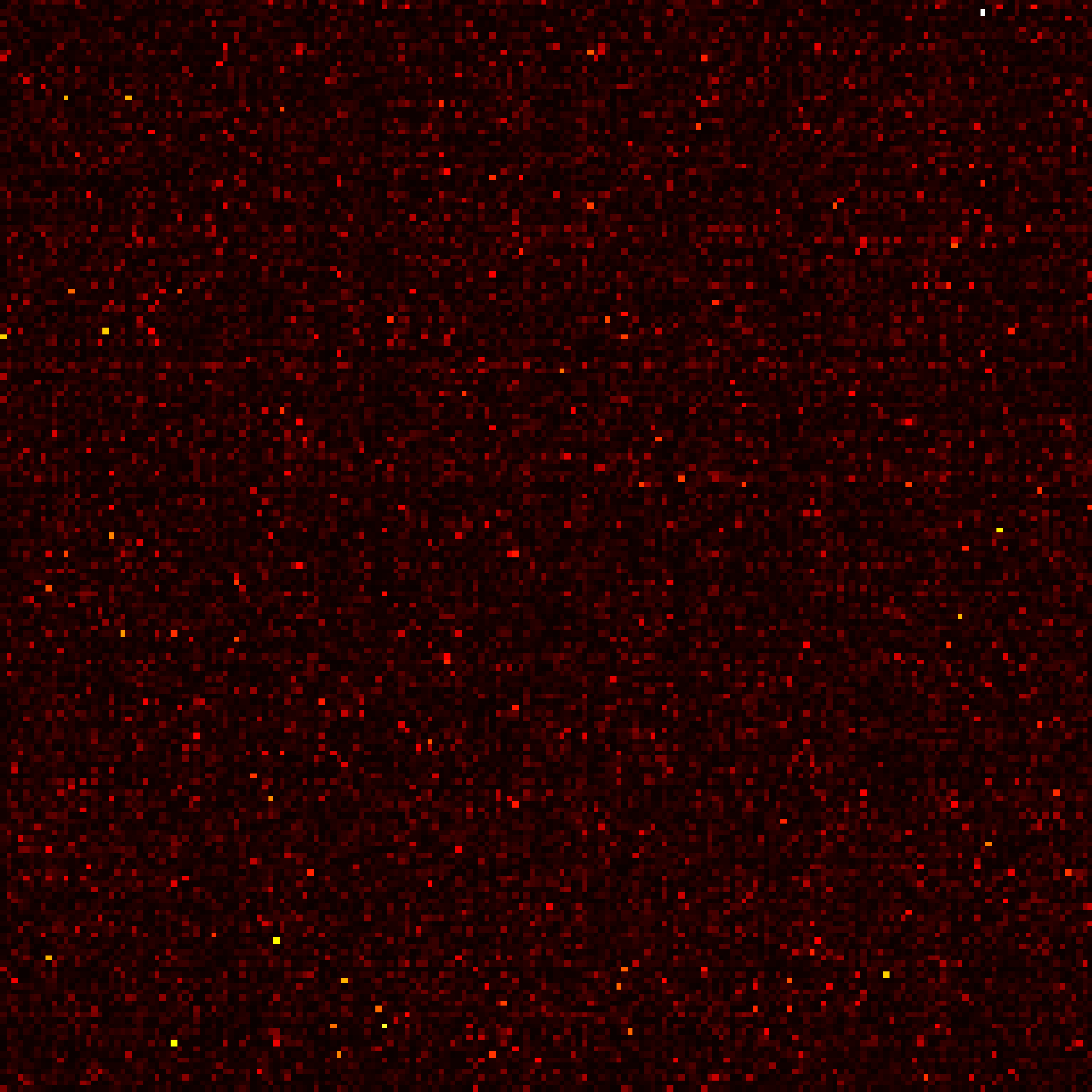}}
~
    \covarlabels{}{192}{}{10}{\includegraphics[width=0.00624996\textwidth]{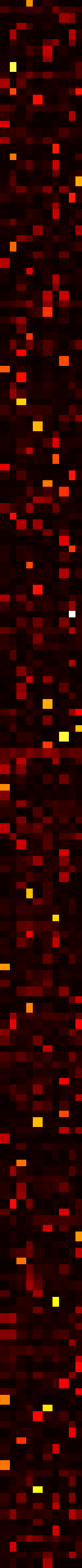}}
\caption{\textbf{Network-in-Network.}}
\vspace*{0.6em}
\label{fig:covaroot1}
\end{subfigure}
~
\begin{subfigure}[c]{\paperwidth}
\centering
    \covarlabels{conv1a}{192}{conv1b}{160}{\includegraphics[width=0.1\textwidth]{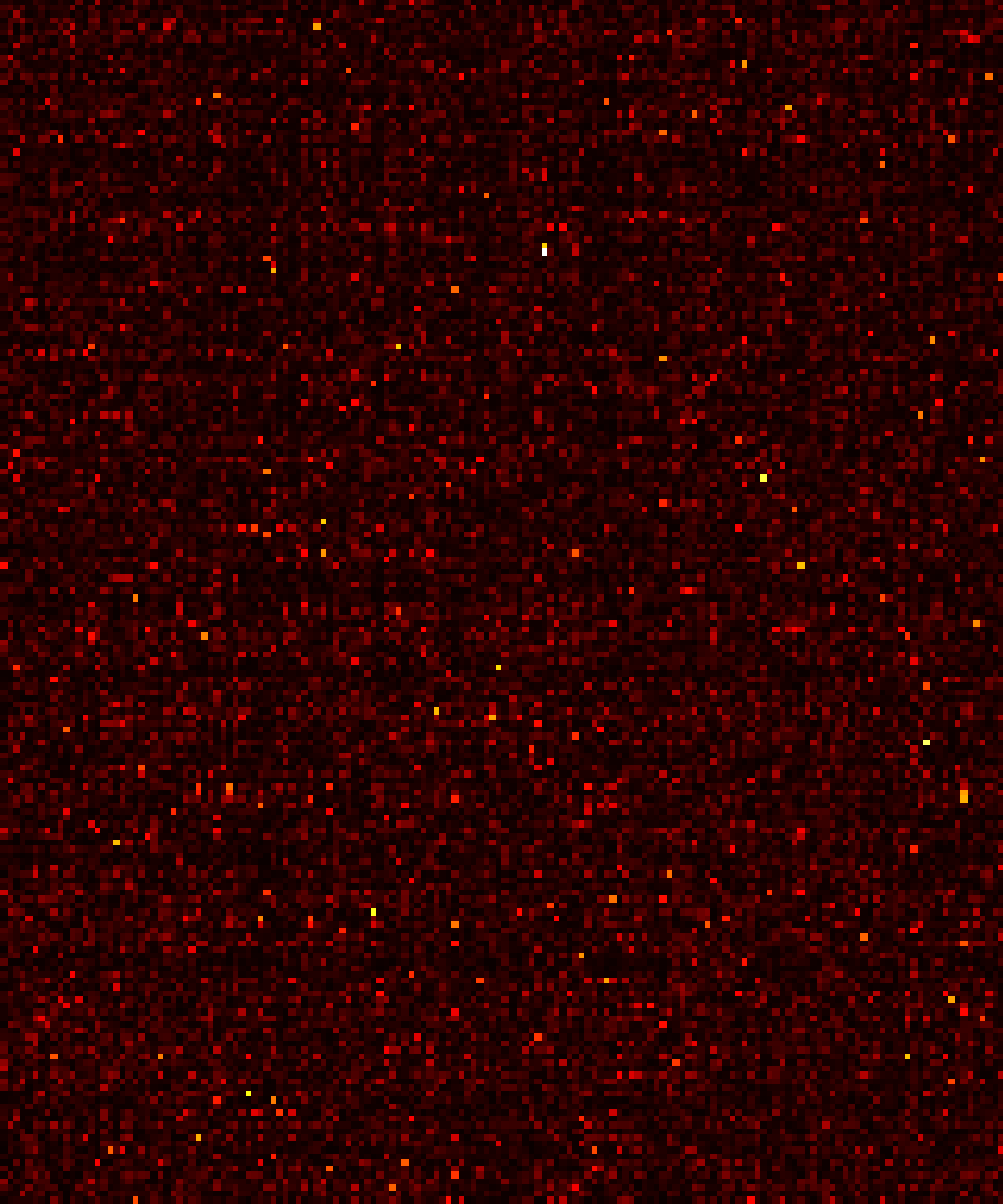}}
~
    \covarlabels{conv1b}{160}{conv1c}{96}{\includegraphics[width=0.05\textwidth]{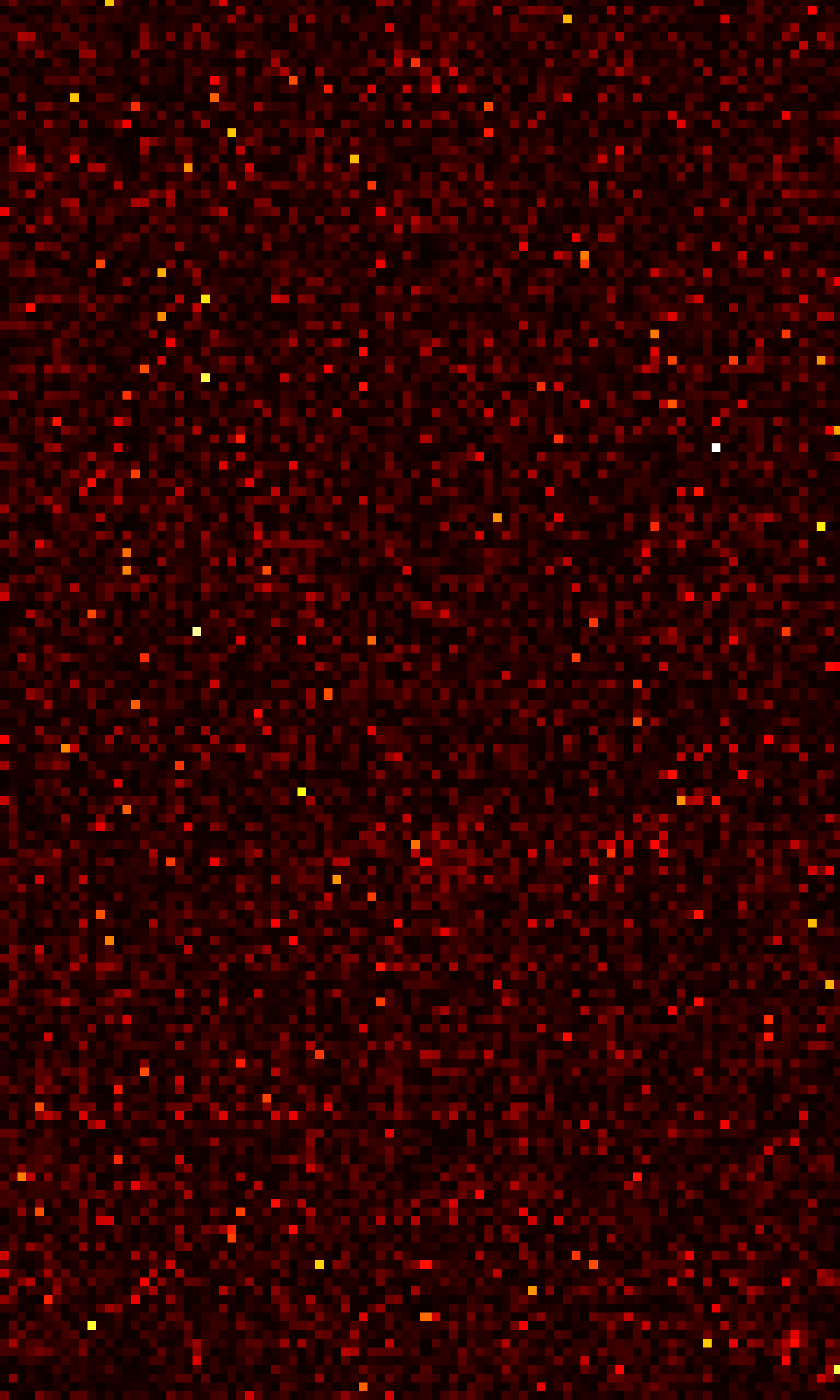}}
~
    \covarlabels{conv1c}{96}{conv2a}{192}{\includegraphics[width=0.12\textwidth]{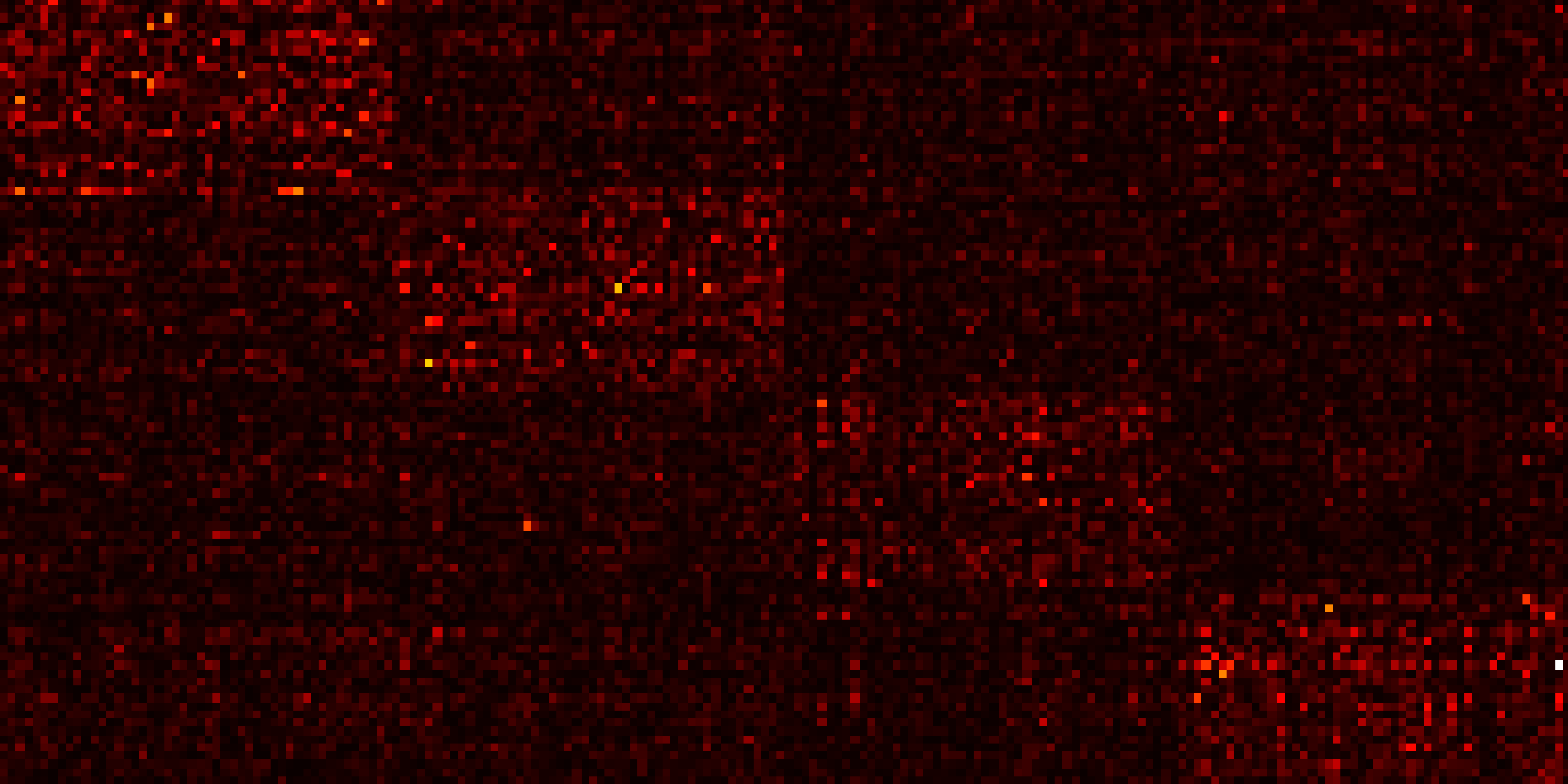}}
~
    \covarlabels{conv2a}{192}{conv2b}{192}{\includegraphics[width=0.12\textwidth]{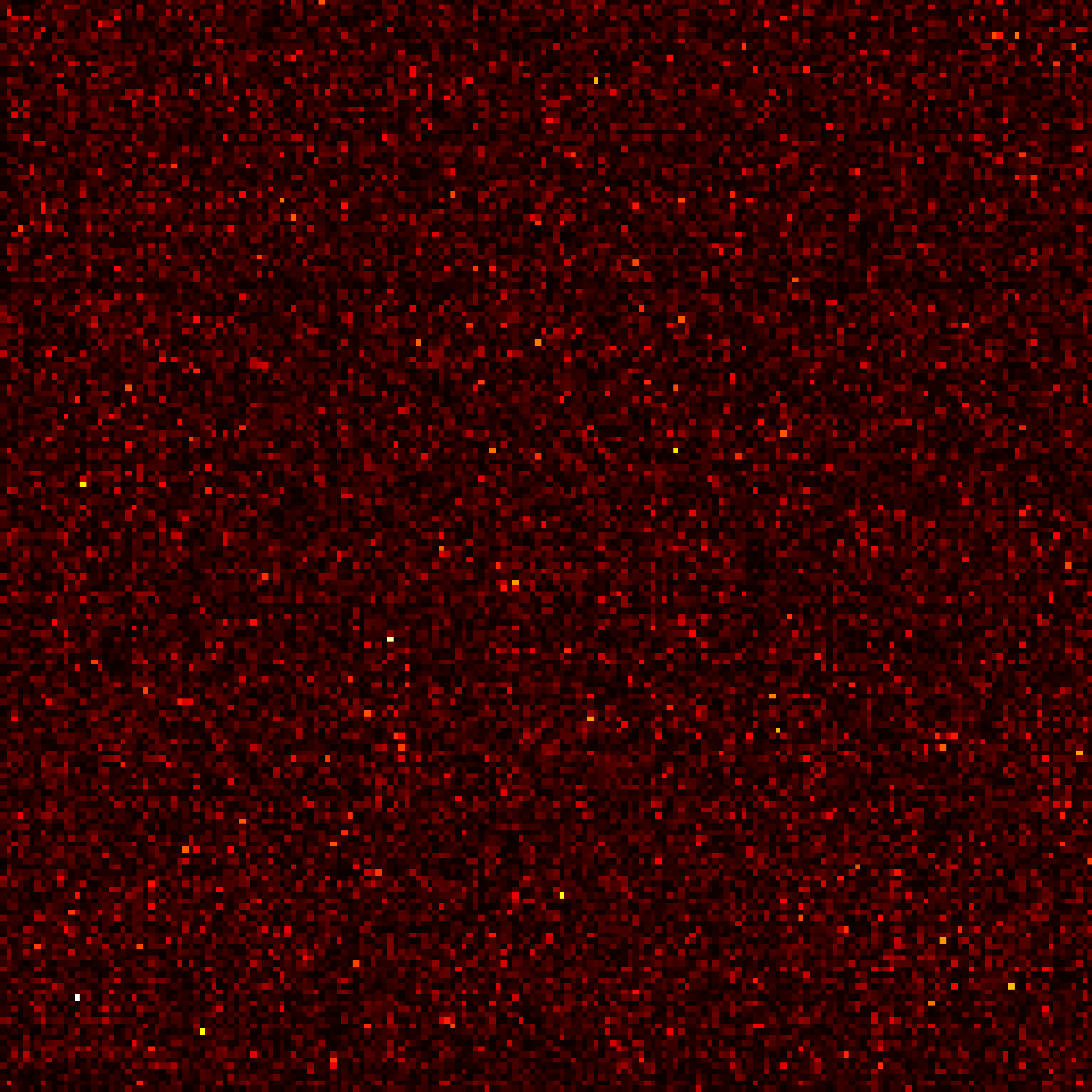}}
~
    \covarlabels{conv2b}{192}{conv2c}{192}{\includegraphics[width=0.12\textwidth]{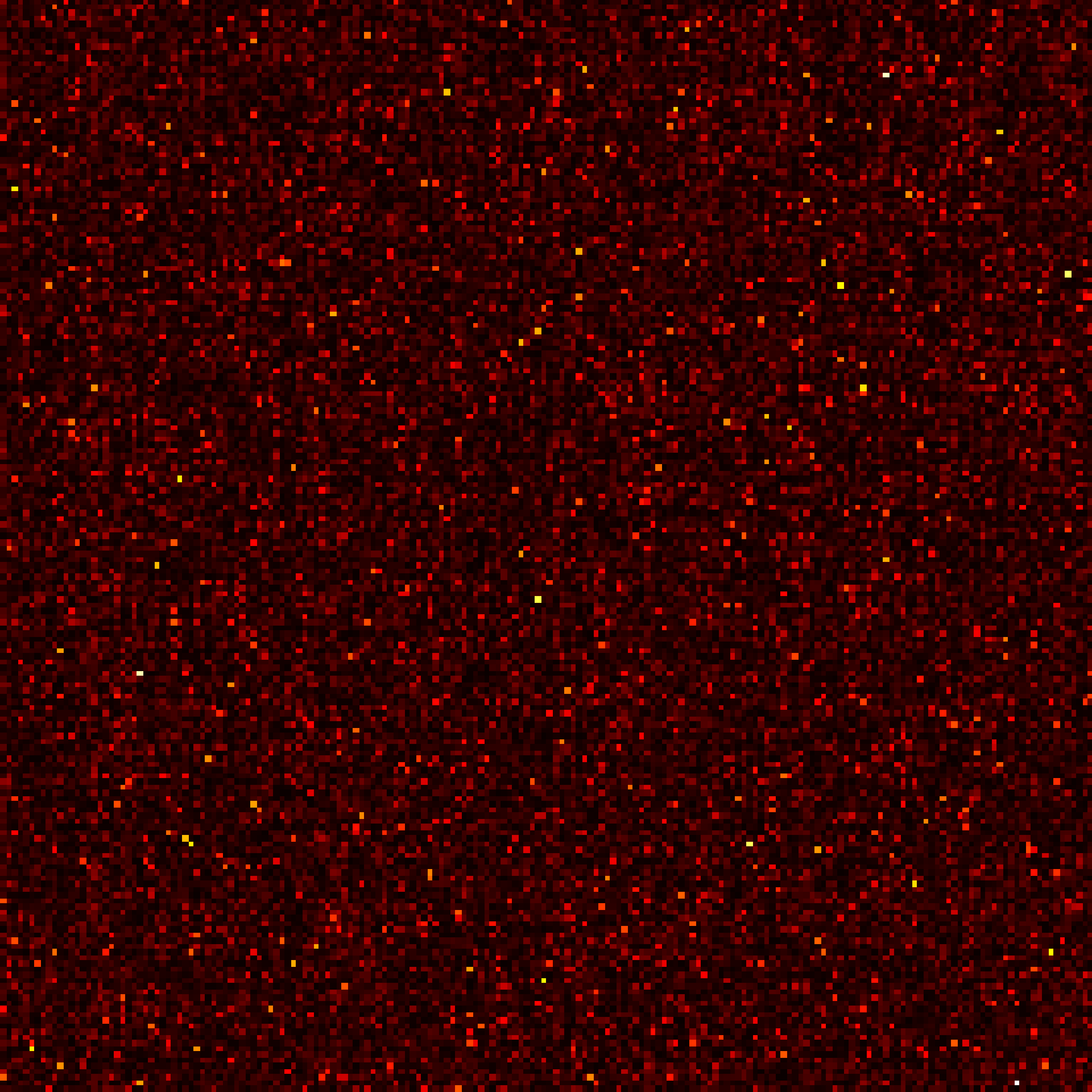}}
~
    \covarlabels{conv2c}{192}{conv3a}{192}{\includegraphics[width=0.12\textwidth]{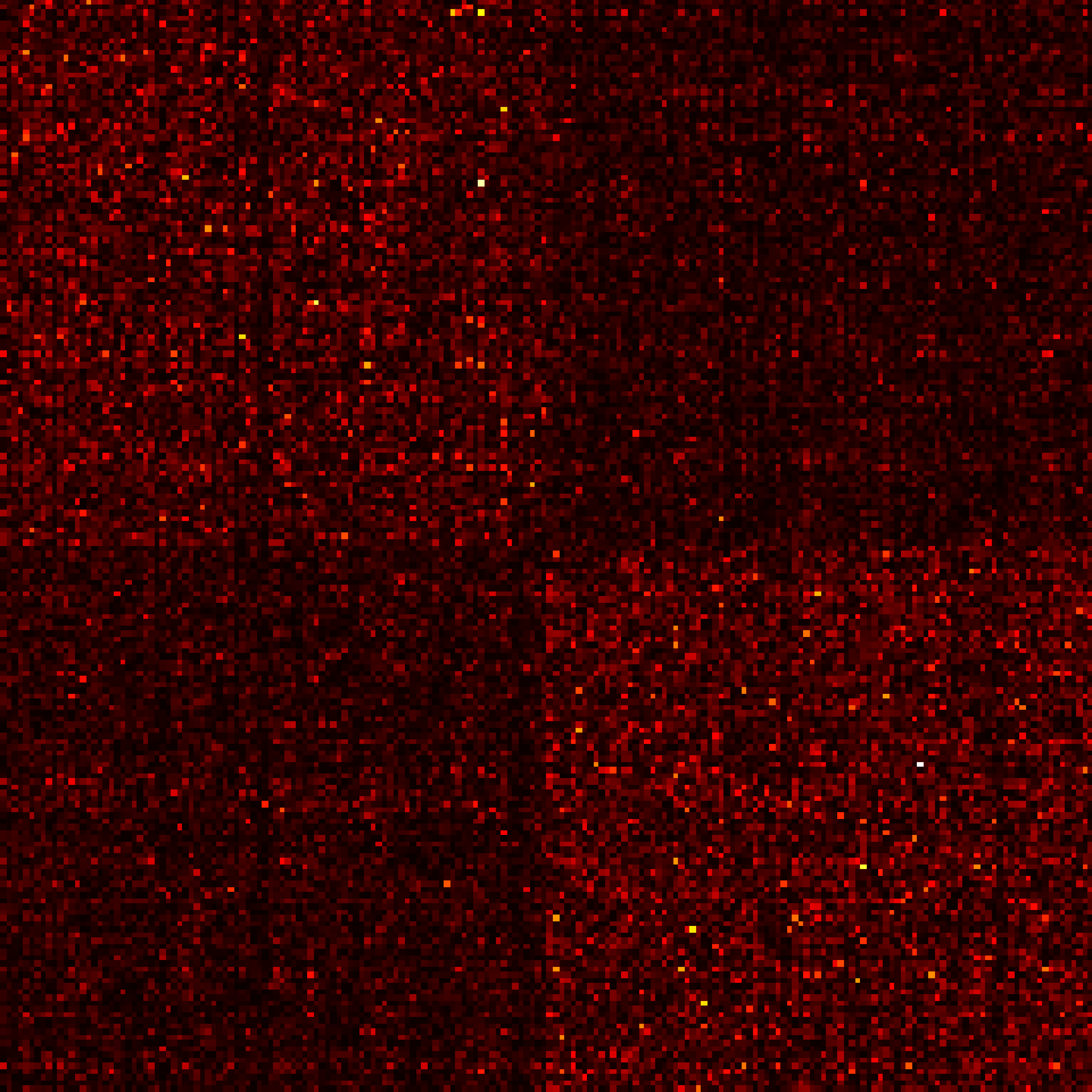}}
~
    \covarlabels{conv3a}{192}{conv3b}{192}{\includegraphics[width=0.12\textwidth]{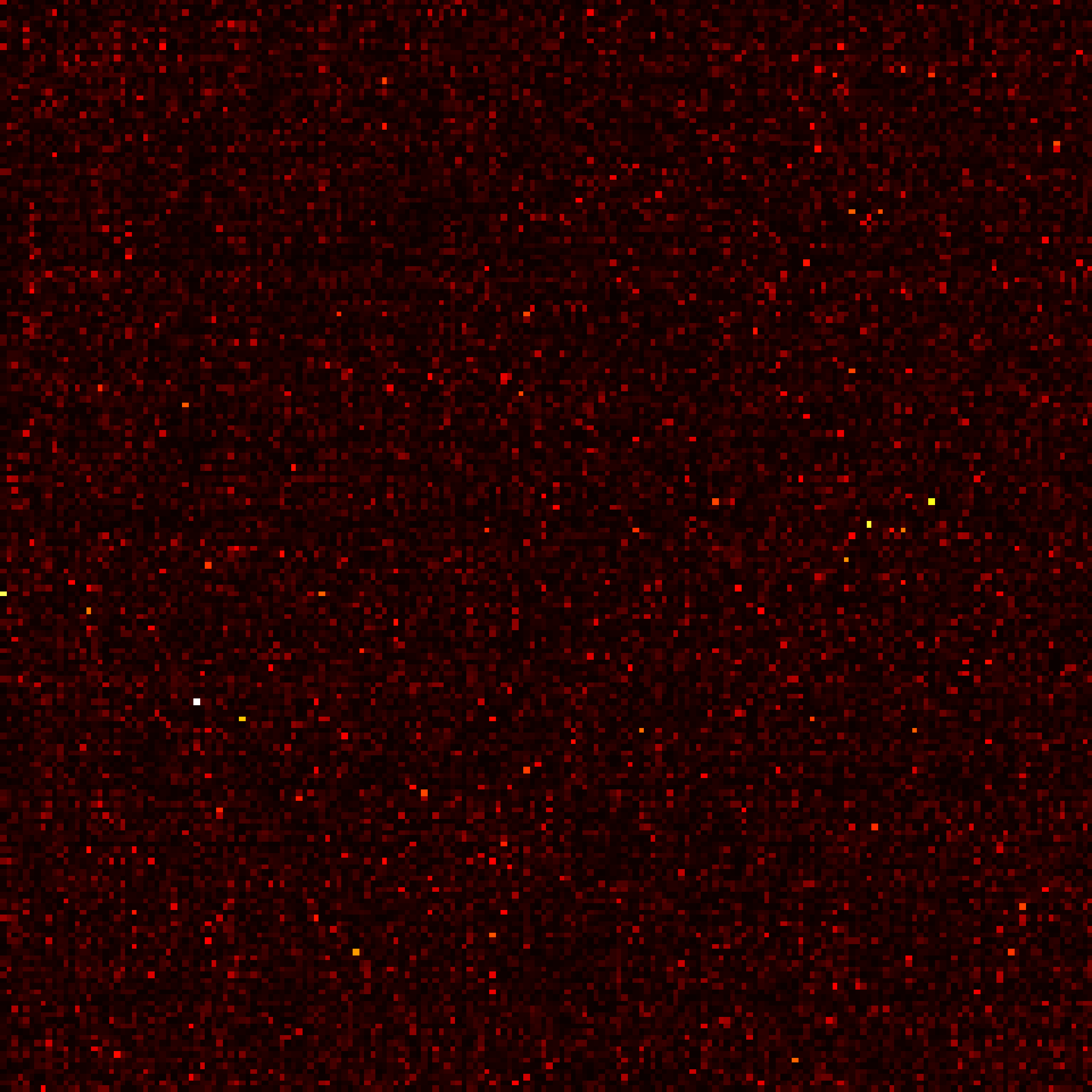}}
~
    \covarlabels{}{192}{}{10}{\includegraphics[width=0.00624996\textwidth]{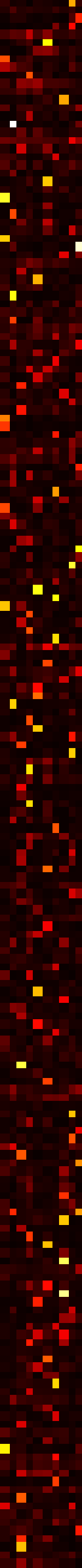}}
\caption{\textbf{Root-4.}}
\vspace*{0.6em}
\label{fig:covarroot4}
\end{subfigure}
\begin{subfigure}[c]{\paperwidth}
\centering
    \covarlabels{conv1a}{192}{conv1b}{160}{\includegraphics[width=0.1\textwidth]{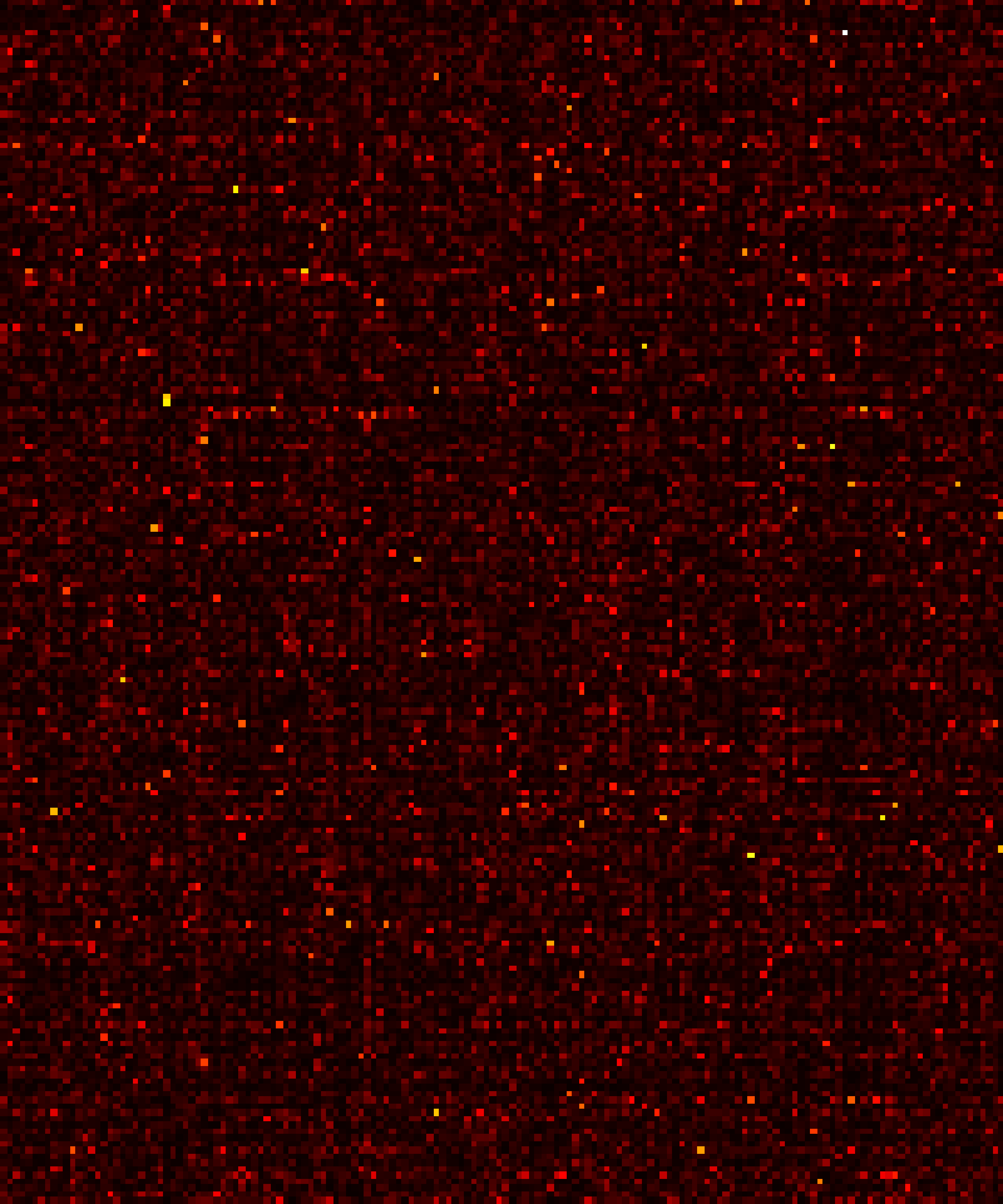}}
~
    \covarlabels{conv1b}{160}{conv1c}{96}{\includegraphics[width=0.05\textwidth]{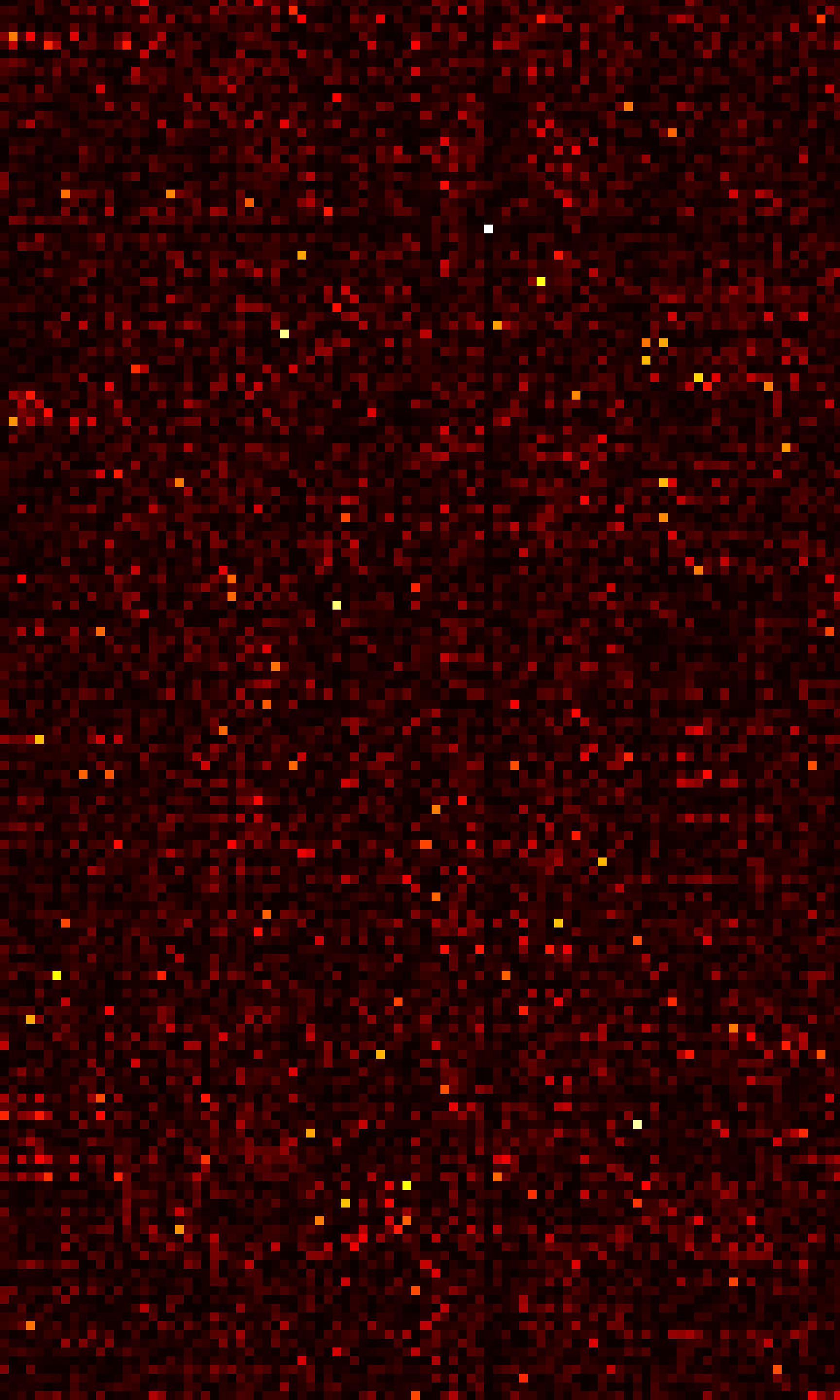}}
~
    \covarlabels{conv1c}{96}{conv2a}{192}{\includegraphics[width=0.12\textwidth]{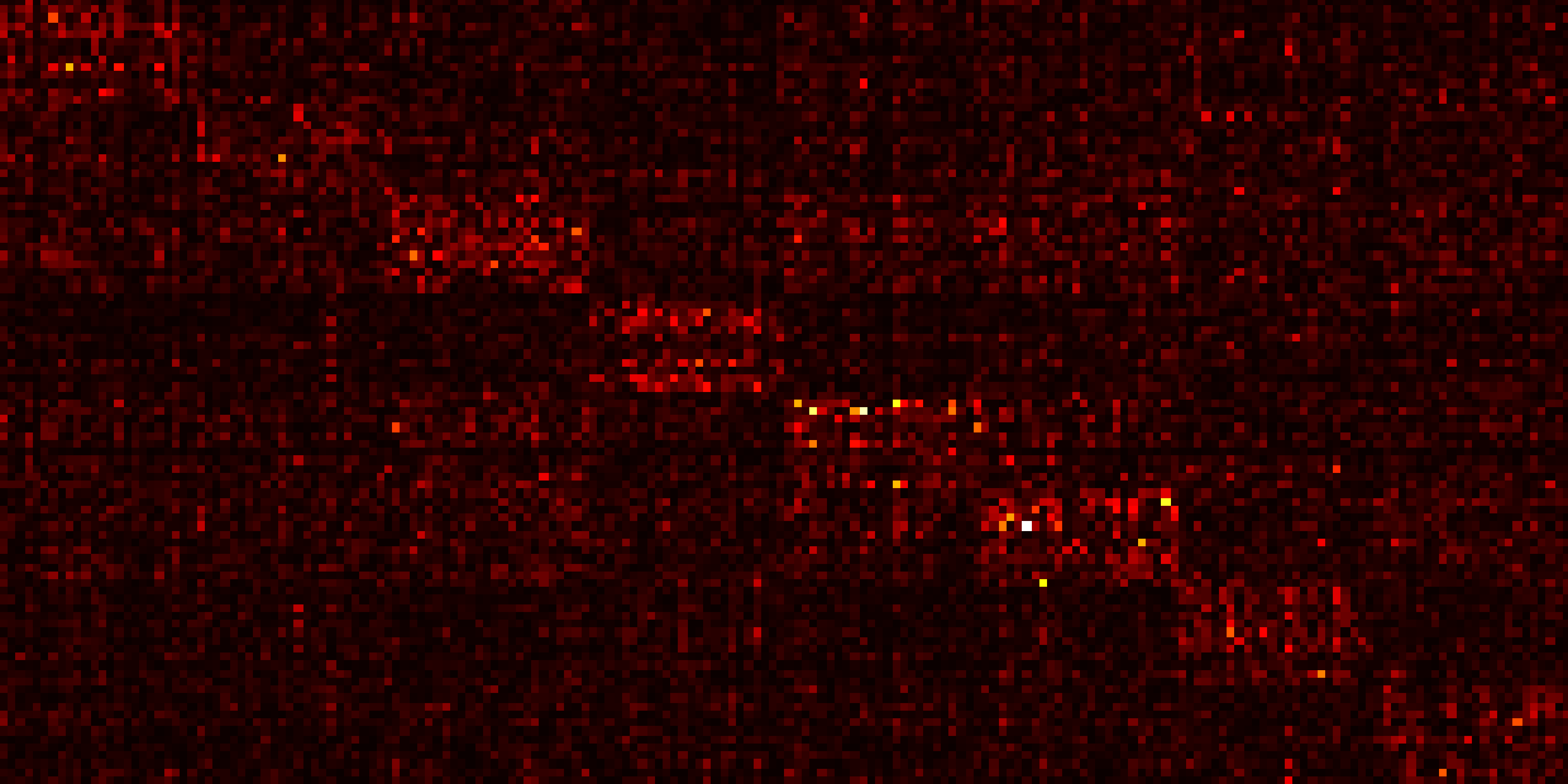}}
~
    \covarlabels{conv2a}{192}{conv2b}{192}{\includegraphics[width=0.12\textwidth]{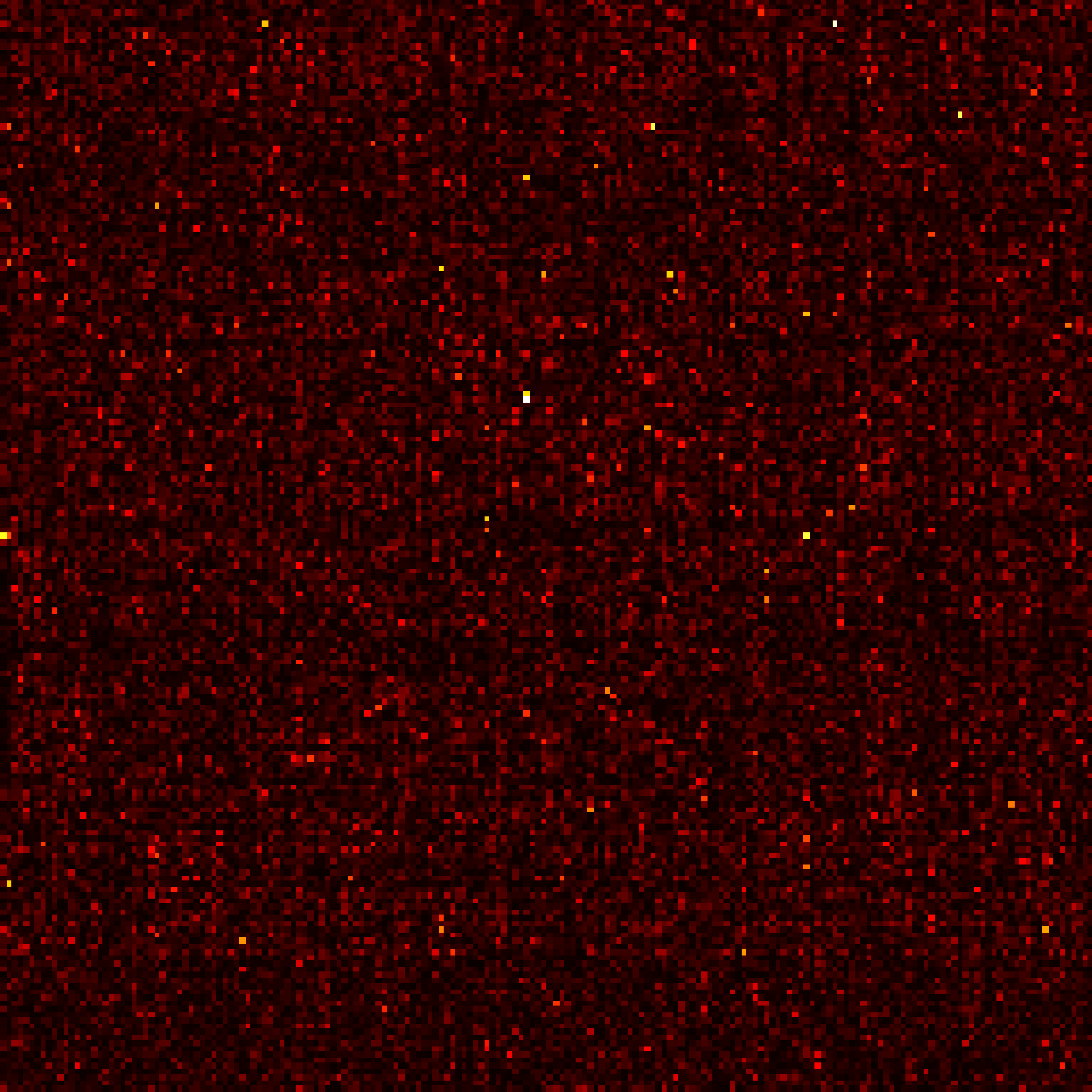}}
~
    \covarlabels{conv2b}{192}{conv2c}{192}{\includegraphics[width=0.12\textwidth]{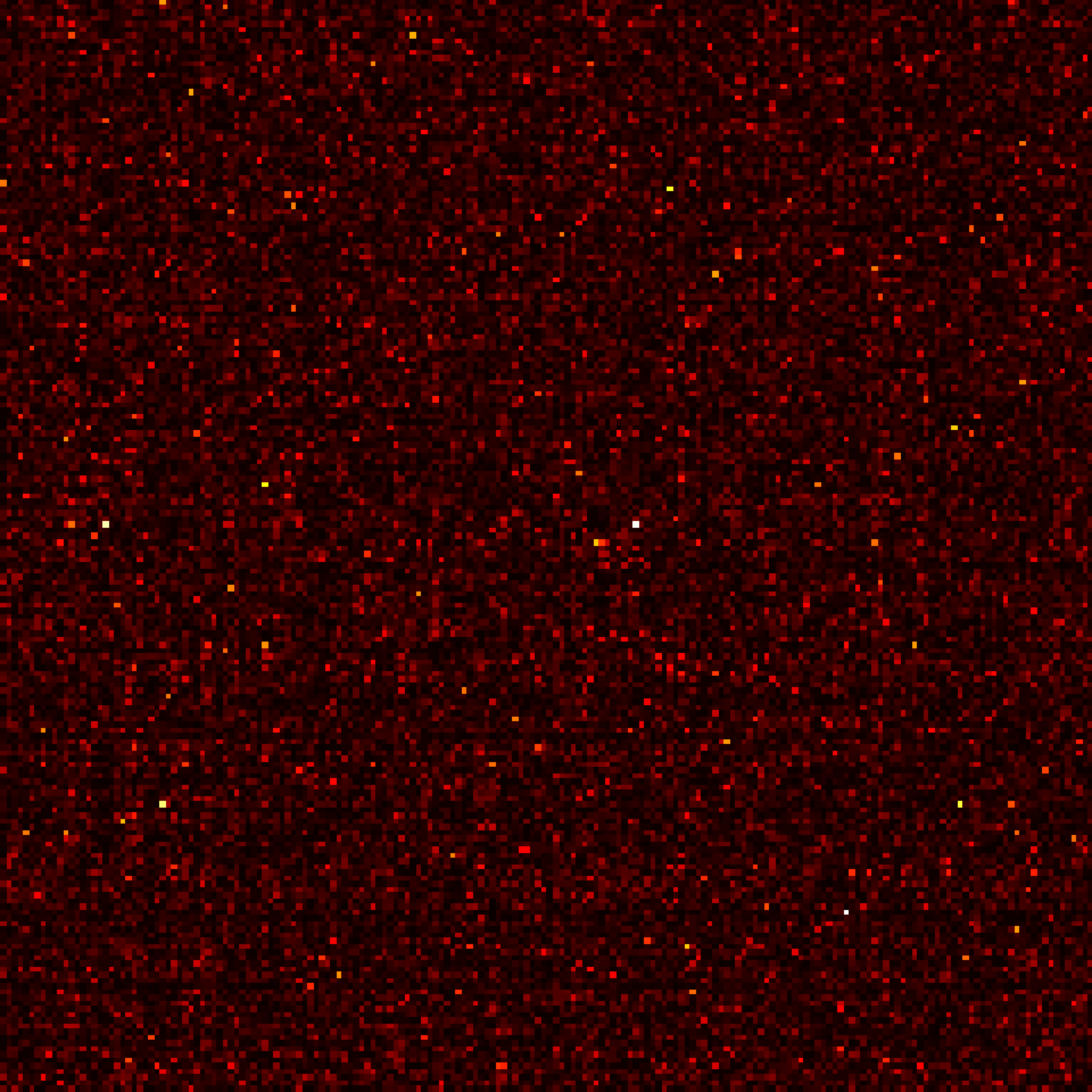}}
~
    \covarlabels{conv2c}{192}{conv3a}{192}{\includegraphics[width=0.12\textwidth]{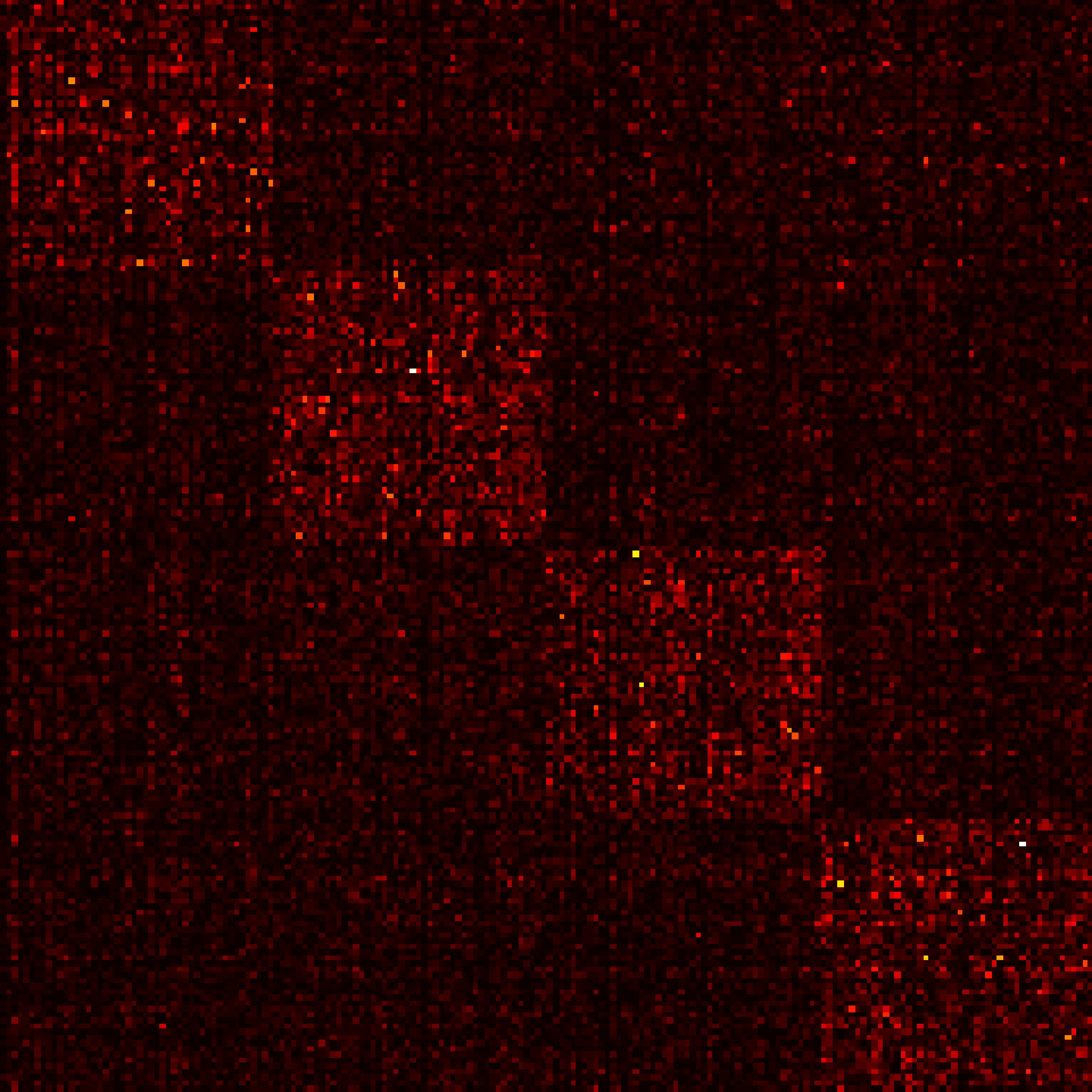}}
~
    \covarlabels{conv3a}{192}{conv3b}{192}{\includegraphics[width=0.12\textwidth]{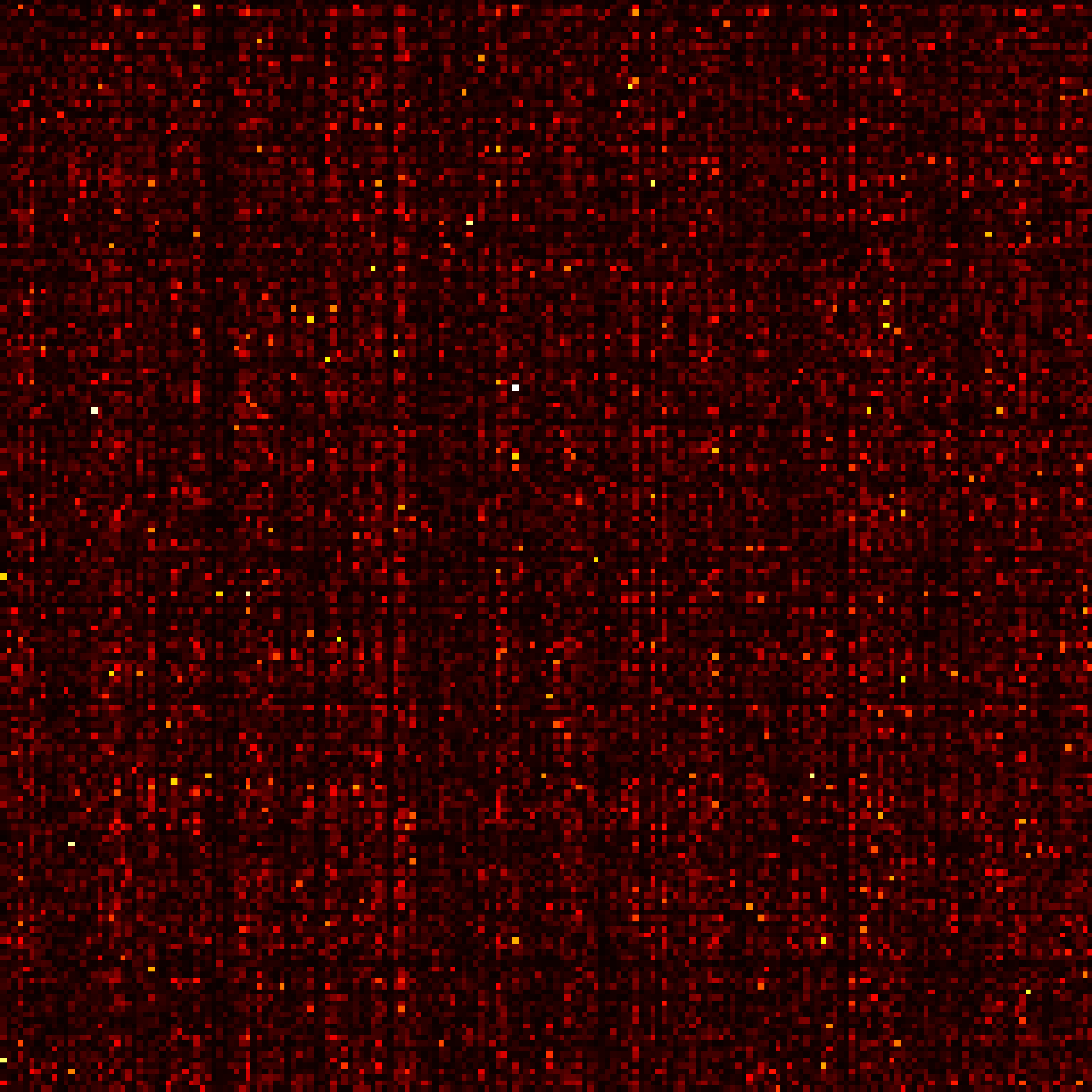}}
~
    \covarlabels{}{192}{}{10}{\includegraphics[width=0.00624996\textwidth]{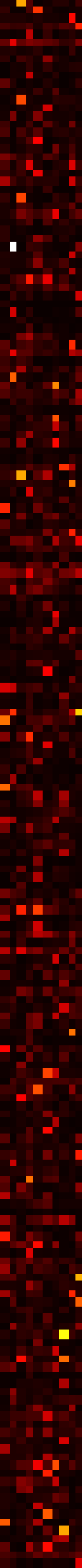}}
\caption{\textbf{Root-8.}}
\vspace*{0.6em}
\label{fig:covarroot8}
\end{subfigure}
\begin{subfigure}[c]{\paperwidth}
\centering
    \covarlabels{conv1a}{192}{conv1b}{160}{\includegraphics[width=0.1\textwidth]{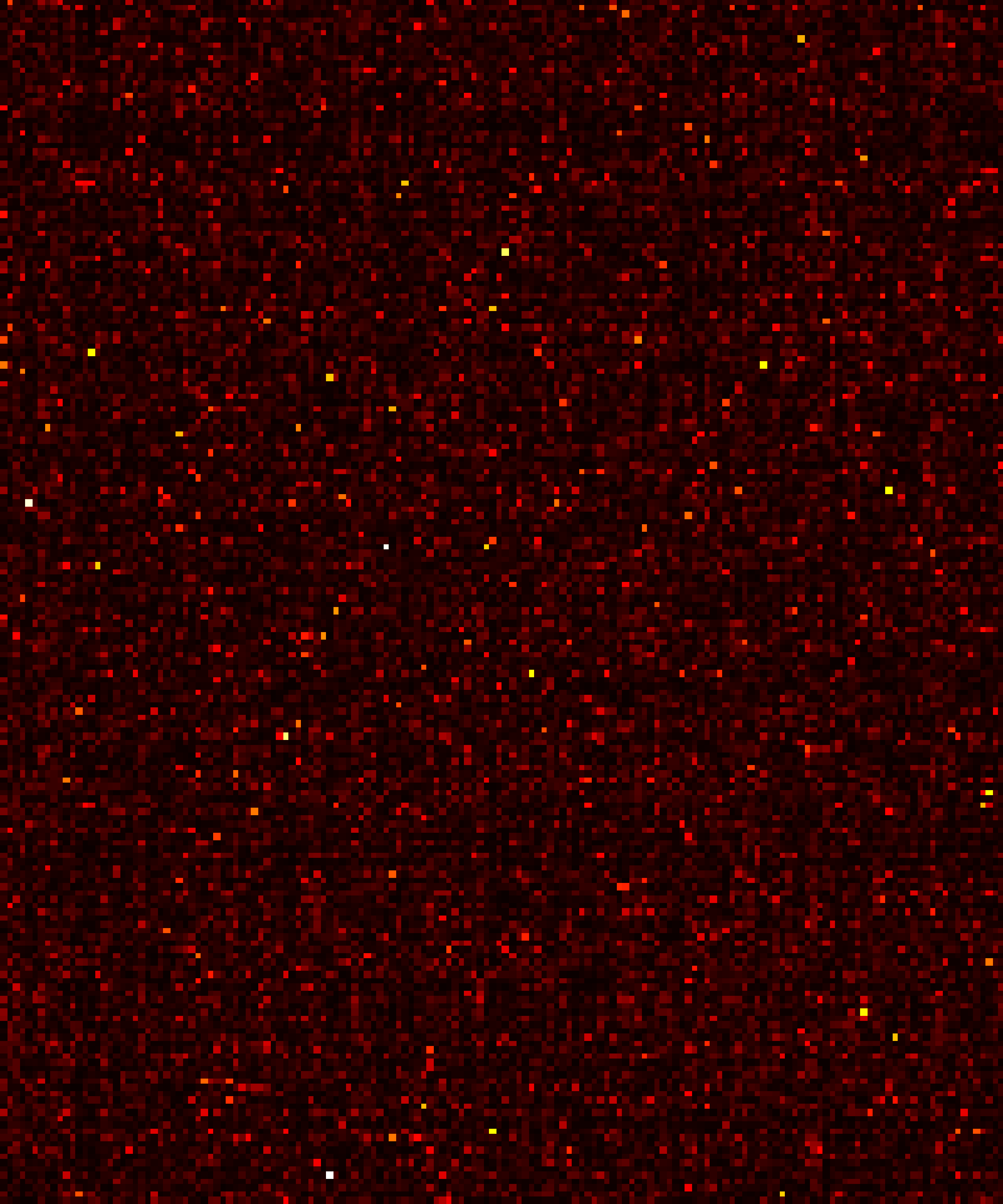}}
~
    \covarlabels{conv1b}{160}{conv1c}{96}{\includegraphics[width=0.05\textwidth]{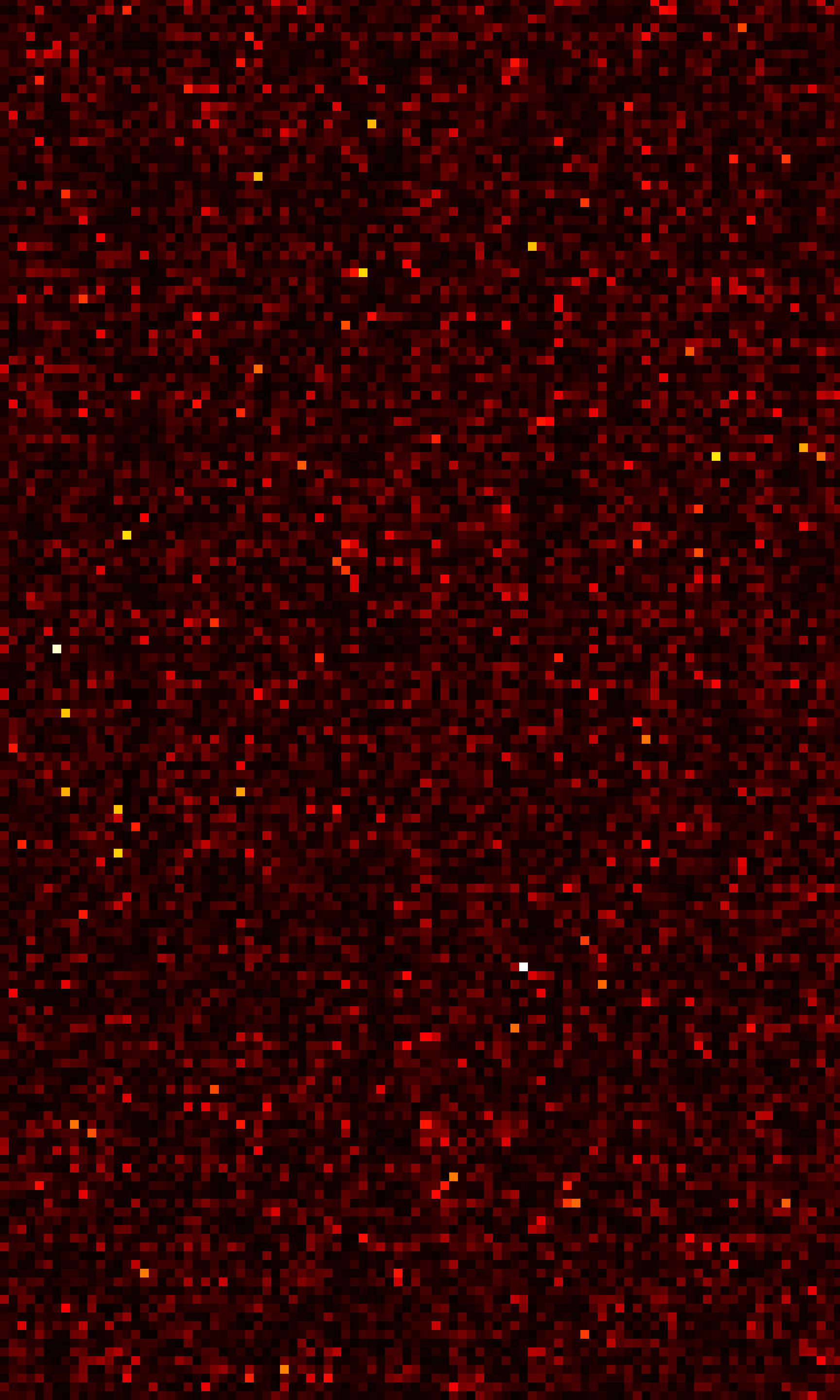}}
~
    \covarlabels{conv1c}{96}{conv2a}{192}{\includegraphics[width=0.12\textwidth]{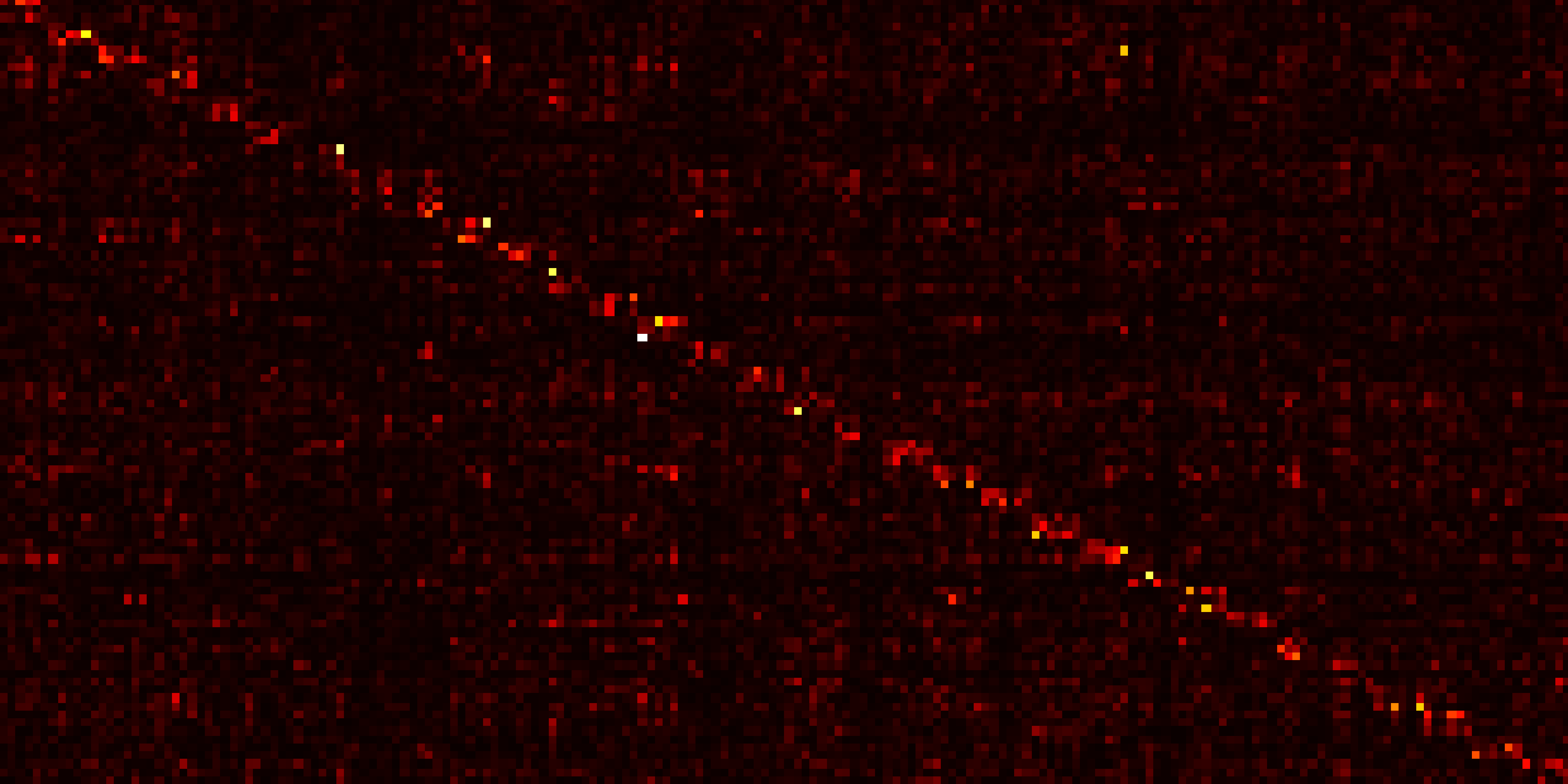}}
~
    \covarlabels{conv2a}{192}{conv2b}{192}{\includegraphics[width=0.12\textwidth]{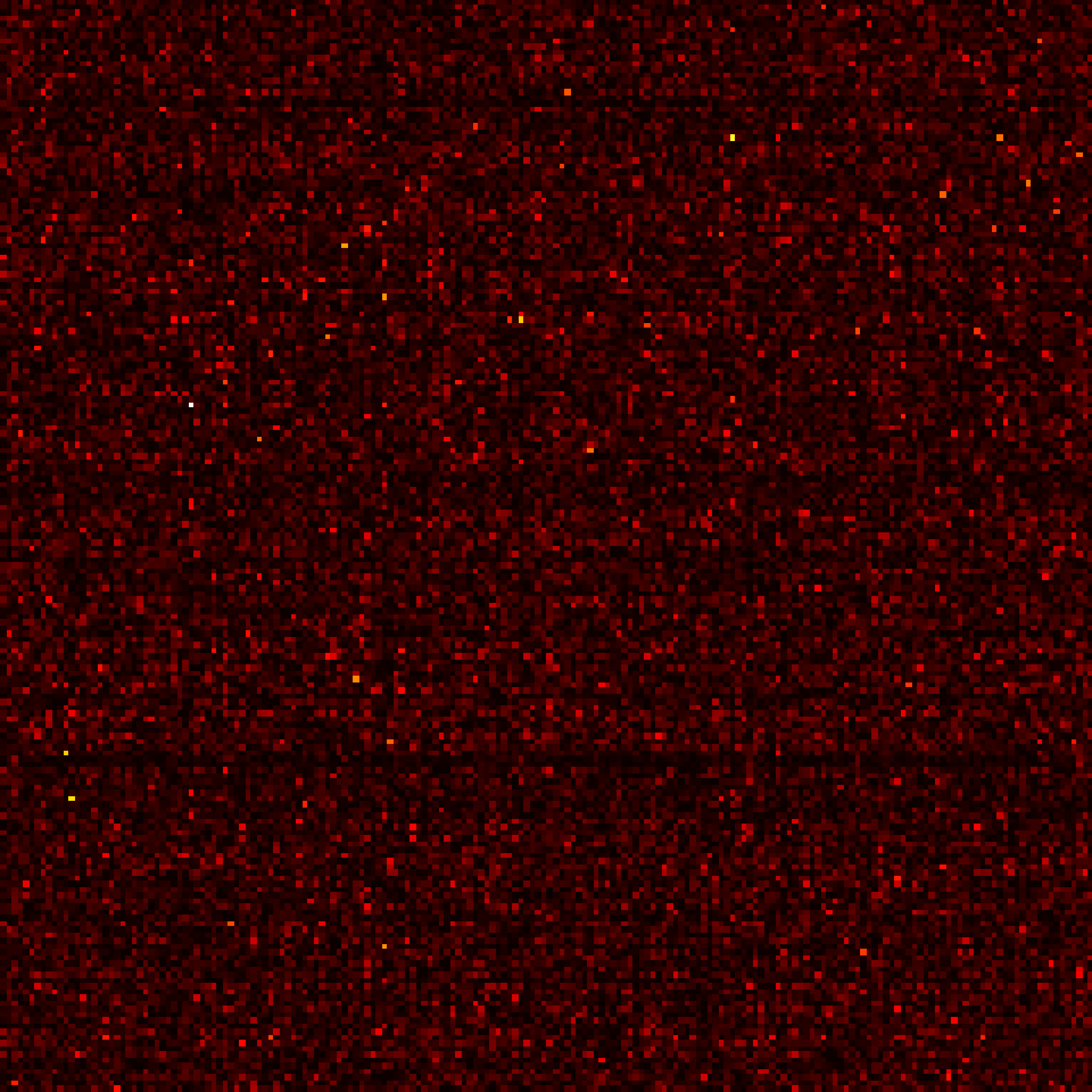}}
~
    \covarlabels{conv2b}{192}{conv2c}{192}{\includegraphics[width=0.12\textwidth]{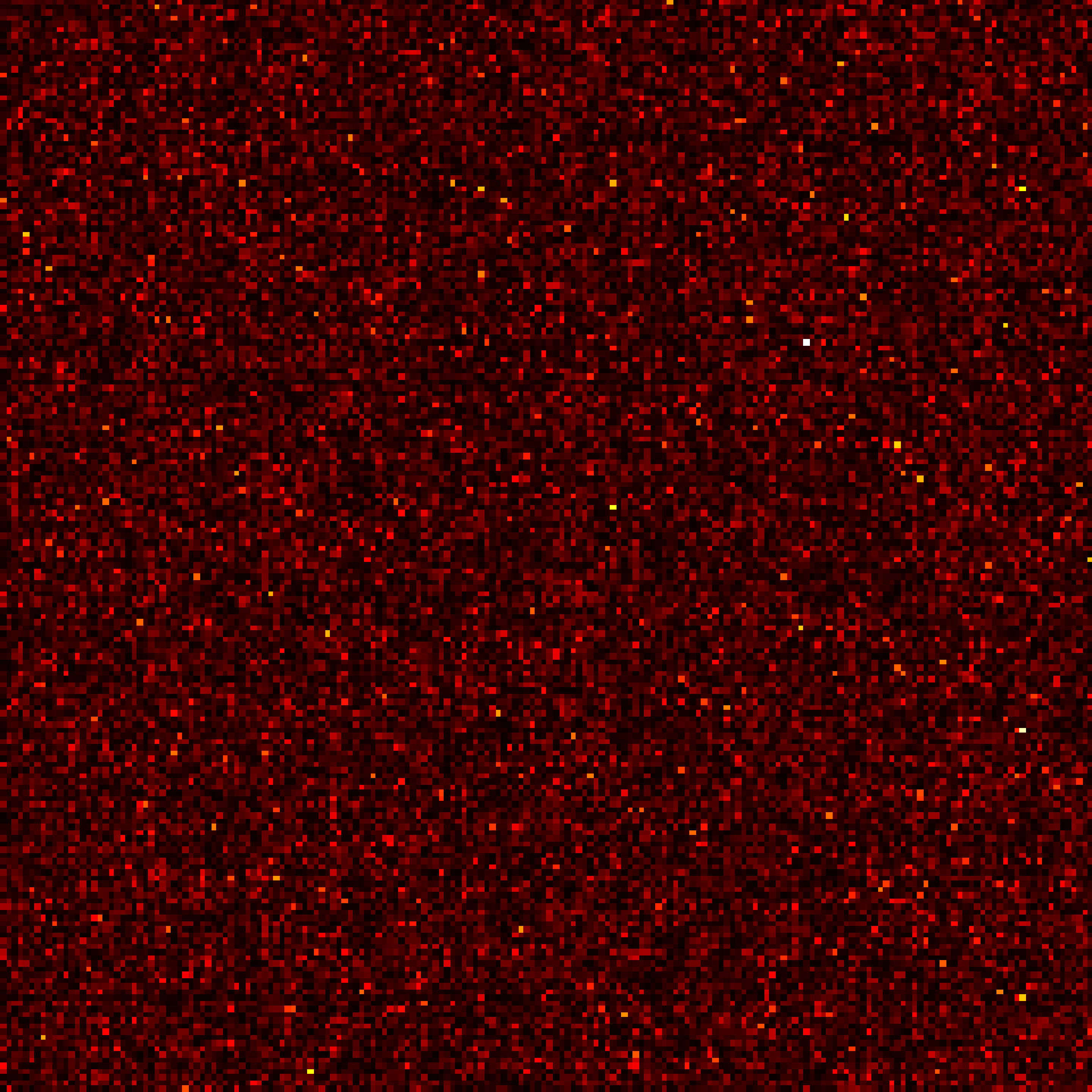}}
~
    \covarlabels{conv2c}{192}{conv3a}{192}{\includegraphics[width=0.12\textwidth]{figs/ninroot32/layercovarwhite_conv8.png}}
~
    \covarlabels{conv3a}{192}{conv3b}{192}{\includegraphics[width=0.12\textwidth]{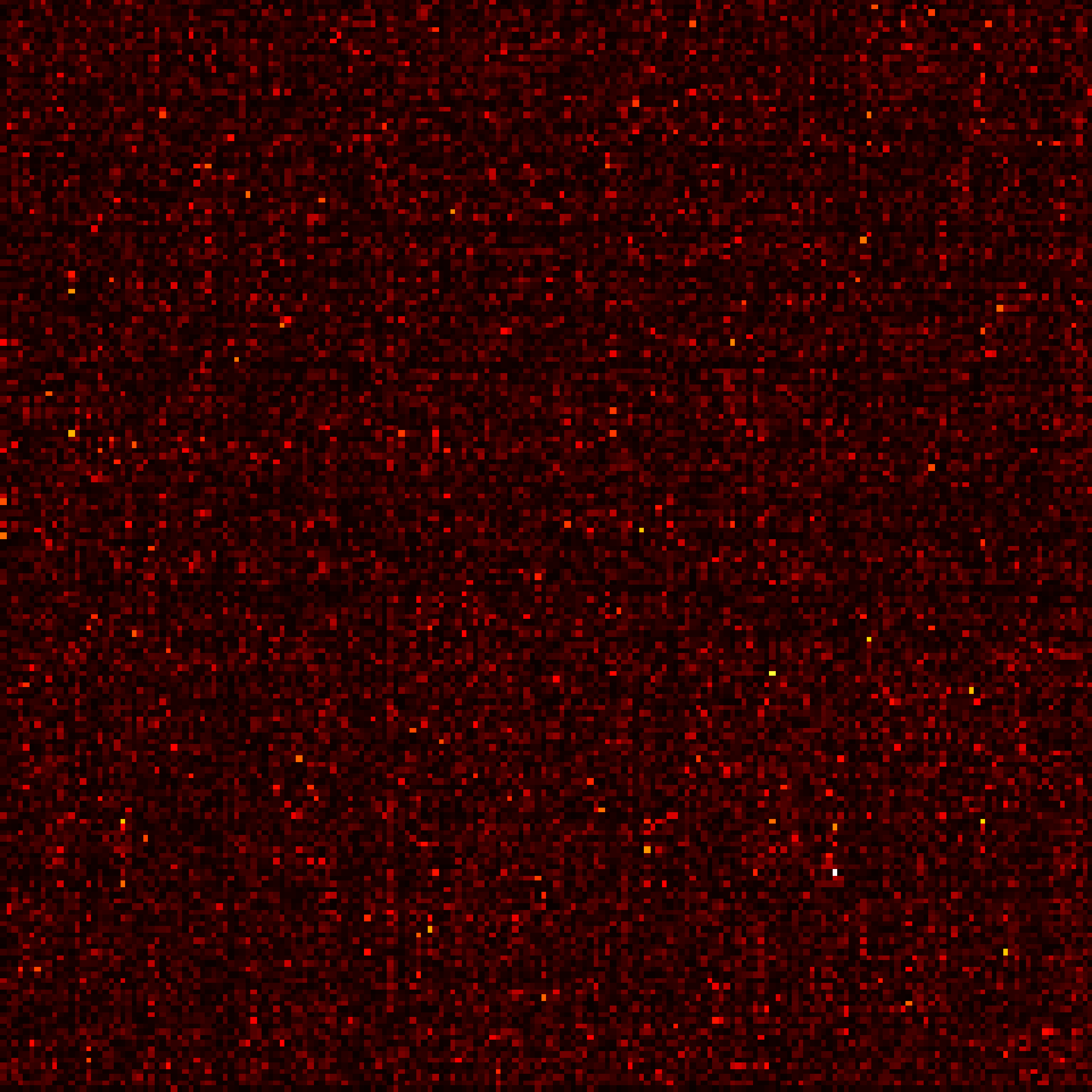}}
~
    \covarlabels{}{192}{}{10}{\includegraphics[width=0.00624996\textwidth]{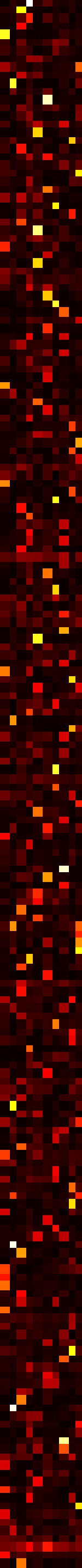}}
\caption{\textbf{Root-32.}}
\label{fig:covarroot32}
\end{subfigure}
\caption{\textbf{Network-in-Network Inter-layer Absolute Covariance.} The inter-layer covariance for all layers in variants of the NiN network.}
\label{fig:suppcovariances}
\end{figure}
\end{landscape}
}

Figure \ref{fig:nincorr} shows the per-layer (intra-layer) filter correlation. This shows the correlation of filters is more structured in root-networks, filters are learned to be linearly combined into useful filters by the root module, and thus filters are often grouped together with other filters they correlate strongly with.

Figure \ref{fig:covarfull} shows the complete, enlarged version of Fig.~\ref{fig:covar}, showing the inter-layer filter covariances between layers \texttt{conv3a} and \texttt{conv2c}. Figure \ref{fig:suppcovariances} shows the full set of inter-layer covariances between all convolutional layers in the NiN models. Block-diagonal sparsity is visible on the layers with filter groups, \texttt{conv2a} and \texttt{conv3a}. This block-diagonal is shown for all variants in more detail in Fig.~\ref{fig:suppcovariances}.

\subsection{The Affect on Image-level Filters of Root Modules}
\begin{figure}[tb]
\centering
\begin{subfigure}[b]{0.45\linewidth}
\centering
    \includegraphics[width=\linewidth]{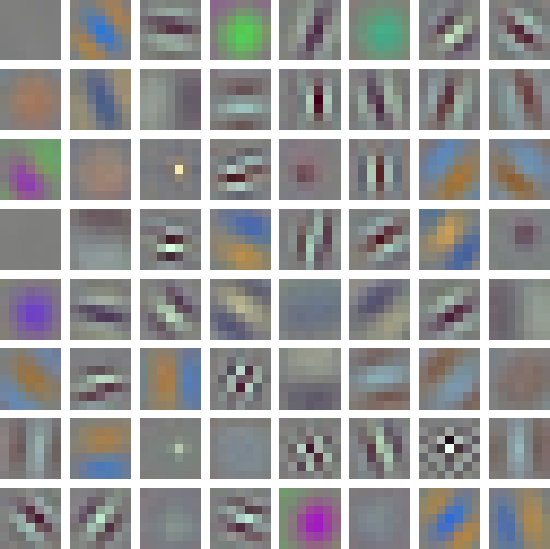}
    \caption{Standard}
    \label{fig:resnet50normalconv0}
\end{subfigure}
~
\begin{subfigure}[b]{0.45\linewidth}
\centering
    \includegraphics[width=\linewidth]{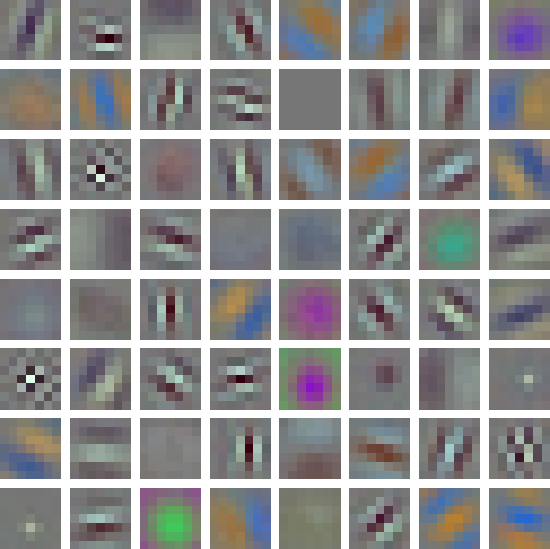}
    \caption{Root-2}
    \label{fig:resnet50root2conv0}
\end{subfigure}
~
\begin{subfigure}[b]{0.45\linewidth}
\centering
    \includegraphics[width=\linewidth]{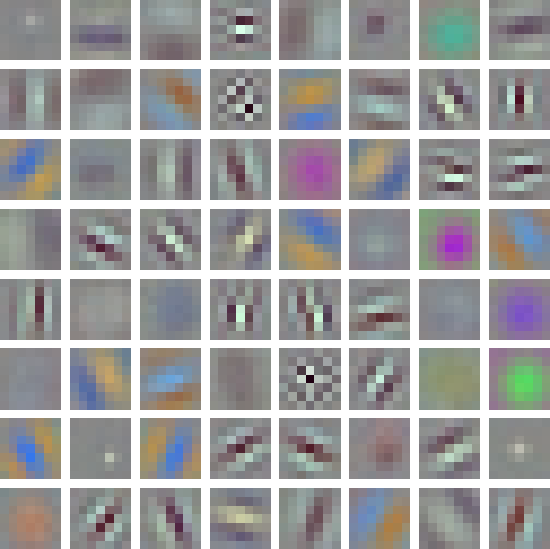}
    \caption{Root-4}
    \label{fig:resnet50root4conv0}
\end{subfigure}
~
\begin{subfigure}[b]{0.45\linewidth}
\centering
    \includegraphics[width=\linewidth]{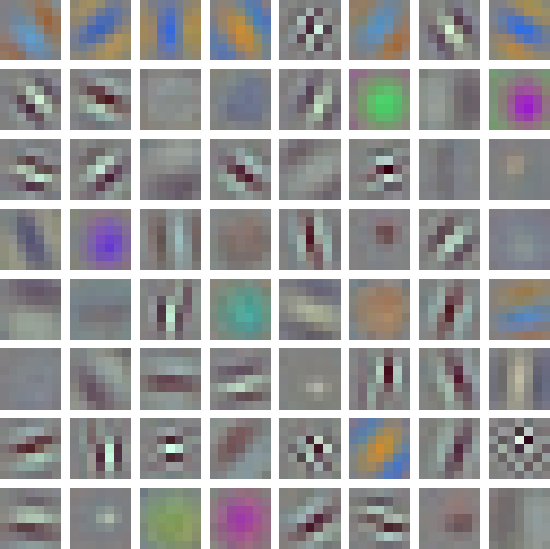}
    \caption{Root-8}
    \label{fig:resnet50root8conv0}
\end{subfigure}
~
\begin{subfigure}[b]{0.45\linewidth}
\centering
    \includegraphics[width=\linewidth]{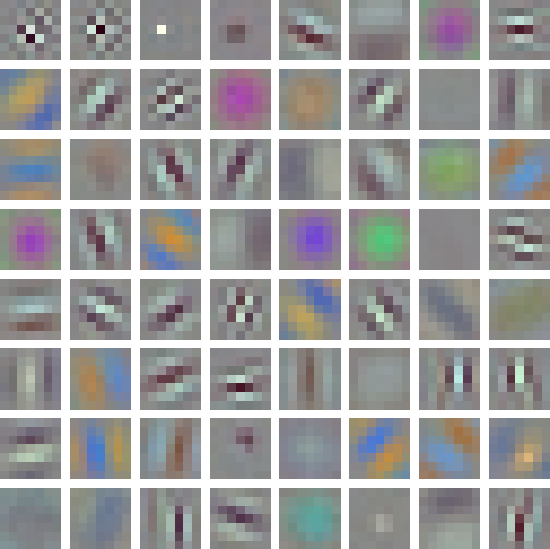}
    \caption{Root-16}
    \label{fig:resnet50root16conv0}
\end{subfigure}
~
\begin{subfigure}[b]{0.45\linewidth}
\centering
    \includegraphics[width=\linewidth]{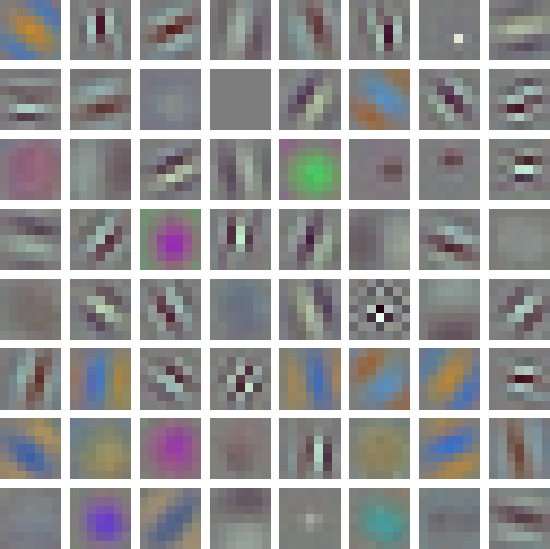}
    \caption{Root-32}
    \label{fig:resnet50root32conv0}
\end{subfigure}
\caption{\textbf{ResNet 50 \texttt{conv1} filters.} With filter groups directly after \texttt{conv1}, in \texttt{conv2}, some of the organization of filters can be directly observed, and give us intuition as to what is happening in root networks.}
\label{fig:resnet50conv0}
\end{figure}
In the ResNet root models, filter groups are used in \texttt{conv2},  directly after the image level filters of \texttt{conv1} some of the organization of filters can be directly observed, and give us intuition as to what is happening in root networks. Figure \ref{fig:resnet50conv0} shows the \texttt{conv0} filters learned for each of the ResNet 50 models. It is apparent that the filters learned in these networks are very similar to those learned in the original model, although sometimes inverted or with a different ordering. This ordering is somewhat consistent in models with filter groups however, even with different random initializations. This is because filter groups cause filters with strong mutual information to be grouped adjacent to each other.

For example, in the root-8 network (Fig.~\ref{fig:resnet50root8conv0}), each row of filters corresponds to the input of an independent filter group in \texttt{conv2}. We can see that the first row primarily is composed of filters giving various directions of the same color gradient. These filters can be combined in the next layer to produce color edges easily. Due to the shortcut layer and the learned combinations of filters however, not all filter groupings are so obvious.

\subsection{Layer-wise Compute/Parameter Savings}
\begin{figure}[tbp]
\centering
\begin{subfigure}[b]{\linewidth}

\pgfplotstableread[col sep=comma]{data/resnet50-layerma.csv}\datatable
\pgfplotstableread[col sep=comma]{data/resnet50-root64-layerma.csv}\datatablea
\begin{tikzpicture}
\begin{axis}[
    axis x line=bottom,
    axis y line=left,
    ybar=0pt,
    bar width=0.15em,
    width=1.05\linewidth,
    height=0.4\linewidth,
    enlarge x limits=0.02,
    ylabel=FLOPS,
    y label style={at={(axis description cs:0.09,.5)},anchor=south},
    y tick label style={
        /pgf/number format/.cd,
            fixed,
            fixed zerofill,
            precision=1,
        /tikz/.cd
    },
    ymin=0,
    xmajorticks=false,
    legend style={at={(0.5,1.2)},
      draw=none, anchor=north,legend columns=-1},
    area legend,
]
\addplot[ybar, draw=none, fill=red!40] table [x expr=\coordindex,y=MA]{\datatable};
\addplot[ybar, draw=none, fill=blue!40] table [x expr=\coordindex,y=MA]{\datatablea};
\legend{ResNet 50, Root-64};
\end{axis}
\end{tikzpicture}

\end{subfigure}

\begin{subfigure}[b]{\linewidth}

\pgfplotstableread[col sep=comma]{data/resnet50-layerparam.csv}\datatable
\pgfplotstableread[col sep=comma]{data/resnet50-root64-layerparam.csv}\datatablea
\begin{tikzpicture}
\begin{axis}[
    axis x line=bottom,
    axis y line=left,
    ybar=0pt,
    bar width=0.15em,
    width=1.05\linewidth,
    height=0.4\linewidth,
    enlarge x limits=0.02,
    ylabel=Parameters,
    y label style={at={(axis description cs:0.09,.5)},anchor=south},
    y tick label style={
        /pgf/number format/.cd,
            fixed,
            fixed zerofill,
            precision=1,
        /tikz/.cd
    },
    ymin=0,
    xticklabels from table={\datatable}{layer},
    xticklabel style = {rotate = 90, xshift = -0.8ex, anchor = mid east, font=\tiny},
    xtick=data,
]
\addplot[ybar, draw=none, fill=red!40] table [x expr=\coordindex,y=param]{\datatable};
\addplot[ybar, draw=none, fill=blue!40] table [x expr=\coordindex,y=param]{\datatablea};
\end{axis}
\end{tikzpicture}

\end{subfigure}
\caption{\textbf{ResNet 50 Layer-wise FLOPS/Parameters.} 
}

\label{fig:resnet50layerwisema}
\end{figure}
Figure \ref{fig:resnet50layerwisema} shows the difference in compute and parameters for each layer in a standard ResNet-50 model and a root-64 variant. The layers in the original networks with the highest computational complexity are clearly the spatial convolutional layers, \ie layers with 3$\times$3 spatial filters. When instead a root-module is used, the computational complexity of these layers is reduced dramatically. While the low dimensional embedding layers (1$\times$1) are not changed, these have less than half the compute of the spatial convolution layers. The number of parameters in spatial convolution layers with large numbers of input channels, which increase towards the end of the network, are similarly reduced.
\end{document}

%% file: ACsettings.tex
\definecolor{accolornotes}{rgb}{0.7,0.3,0.2}

\definecolor{changes}{rgb}{0.45,0,0}

\definecolor{yicolornotes}{rgb}{0.3,0.7,0.2}

\definecolor{jscolornotes}{rgb}{0.7,0.3,0.7}

\definecolor{acgray}{rgb}{0.8,0.8,0.8}

\definecolor{myred}{rgb}{0.6, 0, 0}
\definecolor{myblue}{rgb}{0.3, 0.1, 0.9}

\newcommand{\be}{\begin{equation}}
\newcommand{\ee}{\end{equation}}
\newcommand{\bea}{\begin{eqnarray}}
\newcommand{\eea}{\end{eqnarray}}
\newcommand{\beas}{\begin{eqnarray*}}
\newcommand{\eeas}{\end{eqnarray*}}

\renewcommand{\eqref}[2][\reflabel]{(\ref{eq:#1-#2})} 

\newcommand{\reflabel}{dummy} 

\DeclareMathAlphabet{\mathcal}{OMS}{cmsy}{m}{n}

%% file: pgftablehighlight.tex
\newcommand{\findmax}[3]{
    \pgfplotstablevertcat{\datatable}{#1}
    \pgfplotstablecreatecol[
    create col/expr={%
    \pgfplotstablerow
    }]{rownumber}\datatable
    \pgfplotstablesort[sort key={#2},sort cmp={float >}]{\sorted}{\datatable}%
    \pgfplotstablegetelem{0}{rownumber}\of{\sorted}%
    \pgfmathtruncatemacro#3{\pgfplotsretval}
    \pgfplotstableclear{\datatable}
}

\newcommand{\findmin}[3]{
    \pgfplotstablevertcat{\datatable}{#1}
    \pgfplotstablecreatecol[
      create col/expr={%
    \pgfplotstablerow
    }]{rownumber}\datatable
    \pgfplotstablesort[sort key={#2},sort cmp={float <}]{\sorted}{\datatable}%
    \pgfplotstablegetelem{0}{rownumber}\of{\sorted}%
    \pgfmathtruncatemacro#3{\pgfplotsretval}
    \pgfplotstableclear{\datatable}
}

\pgfplotstableset{
    highlight col max/.code 2 args={
        \findmax{#1}{#2}{\maxval}
        \edef\setstyles{\noexpand\pgfplotstableset{
                every row \maxval\noexpand\space column #2/.style={
                    postproc cell content/.append style={
                        /pgfplots/table/@cell content/.add={$\noexpand\bf}{$}
                    },
                }
            }
        }\setstyles
    },
    highlight col min/.code 2 args={
        \findmin{#1}{#2}{\minval}
        \edef\setstyles{\noexpand\pgfplotstableset{
                every row \minval\noexpand\space column #2/.style={
                    postproc cell content/.append style={
                        /pgfplots/table/@cell content/.add={$\noexpand\bf}{$}
                    },
                }
            }
        }\setstyles
    },
    highlight row max/.code 2 args={
        \pgfmathtruncatemacro\rowindex{#2-1}
        \pgfplotstabletranspose{\transposed}{#1}
        \findmax{\transposed}{\rowindex}{\maxval}
        \edef\setstyles{\noexpand\pgfplotstableset{
                every row \rowindex\space column \maxval\noexpand/.style={
                    postproc cell content/.append style={
                        /pgfplots/table/@cell content/.add={$\noexpand\bf}{$}
                    },
                }
            }
        }\setstyles
    },
    highlight row min/.code 2 args={
        \pgfmathtruncatemacro\rowindex{#2-1}
        \pgfplotstabletranspose{\transposed}{#1}
        \findmin{\transposed}{\rowindex}{\maxval}
        \edef\setstyles{\noexpand\pgfplotstableset{
                every row \rowindex\space column \maxval\noexpand/.style={
                    postproc cell content/.append style={
                        /pgfplots/table/@cell content/.add={$\noexpand\bf}{$}
                    },
                }
            }
        }\setstyles
    },
}

\makeatletter
\long\def\pgfplotstabletypeset@opt@collectarg[#1]#2{%

    \pgfplotstable@isloadedtable{#2}%
        {\pgfplotstabletypeset@opt@[#1]{#2}}%
        {\pgfplotstabletypesetfile@opt@[#1]{#2}}%
}
\makeatother

%% file: alexnetmaplot.tex
\begin{figure}[tb]
\centering
\pgfplotstableread[col sep=comma]{data/alexnetma.csv}\datatable
\pgfplotsset{major grid style={dotted,red}}

\begin{tikzpicture}
\begin{axis}[
  width=\linewidth,
  height=0.4\linewidth,
  axis x line=bottom,
  ylabel=Top-5 Val.\ Error,
  xlabel=Model Parameters (\# floats),
  axis lines=left,
  enlarge x limits=0.12,
  enlarge y limits=0.1,
  grid=major,
  ytick={0.01,0.02,...,0.21},
  ymin=0.18,ymax=0.2,
  yticklabel={\pgfmathparse{\tick*100}\pgfmathprintnumber{\pgfmathresult}\%},style={
        /pgf/number format/fixed,
        /pgf/number format/precision=1
  },
  legend style={at={(0.98,0.98)}, anchor=north east, column sep=0.5em},
  legend columns=2,
]
\addplot[mark=*,mark options={fill=black},nodes near coords,only marks,
   point meta=explicit symbolic,
] table[meta=Network,x=Param.,y expr={1 - \thisrow{Top-5 Acc.} },]{\datatable};
\end{axis}
\end{tikzpicture}
\caption{\textbf{AlexNet Filter Groups.} Model Parameters \vs top-5 error for variants of the AlexNet model on ILSVRC image classification dataset. Models with moderate numbers of filter groups have far fewer parameters, yet surprisingly maintain comparable error.}
\label{fig:alexnetplots}
\end{figure}

%% file: cifarninmatable.tex
\begin{table}[tbp]
\caption[Network-in-Network CIFAR10]{\textbf{Network-in-Network CIFAR10}}
\label{table:nincifarresults}
\centering
\pgfplotstableread[col sep=comma]{data/nincifar.csv}\data
\pgfplotstableread[col sep=comma]{data/nincifar_root_s.csv}\codata
\pgfplotstablevertcat{\data}{\codata}
\pgfplotstabletypeset[
    every head row/.style={
    before row=\toprule,after row=\midrule},
    every last row/.style={
    after row=\bottomrule},
    every first row/.style={
    after row=\midrule}, 
    fixed zerofill,     
    columns={full name, ma, param, accuracy, cpu, gpu},
    columns/full name/.style={
        column name=Model,
        string type
    },
    columns/ma/.style={
        column name=FLOPS {\small $\times 10^{8}$},
        preproc/expr={{##1/1e8}}
    },
    columns/param/.style={
        column name=Param. {\small $\times 10^{5}$},
        preproc/expr={{##1/1e5}}
    },
    columns/accuracy/.style={
        column name=Accuracy,
        precision=4
    },
    columns/gpu/.style={
        column name=GPU (ms),
        precision=3
    },
    columns/cpu/.style={
        column name=CPU (ms),
        precision=1
    },
    column type/.add={@{}lp{3em}p{3em}p{4em}p{3em}p{3em}p{3em}@{}}{},
    highlight col max ={\data}{accuracy},
    highlight col min ={\data}{param}, 
    highlight col min ={\data}{ma}, 
    col sep=comma]{\data}
\end{table}

%% file: cifarninmaplot.tex
\begin{figure}[tbp]
\centering
\begin{subfigure}[b]{\linewidth}
\pgfplotstableread[col sep=comma]{data/nincifar.csv}\datatable
\pgfplotstableread[col sep=comma]{data/nincifar_root_s.csv}\rdatatable
\pgfplotstableread[col sep=comma]{data/nincifar_tree_s.csv}\tdatatable
\pgfplotstableread[col sep=comma]{data/nincifar_col_s.csv}\cdatatable
\pgfplotsset{major grid style={dotted,red}}

\centering
\begin{tikzpicture}
\begin{axis}[
  width=\linewidth,
  height=0.66\linewidth,
  axis x line=bottom,
  ylabel=Error,
  xlabel=Model Parameters,
  axis lines=left,
  enlarge x limits=0.05,
  enlarge y limits=0.05,
  grid=major,
  ytick={0.002,0.004,...,1.0},
  xticklabel style={
        /pgf/number format/fixed,
        /pgf/number format/precision=1
  },
  yticklabel={\pgfmathparse{\tick*1}\pgfmathprintnumber{\pgfmathresult}\%},style={
        /pgf/number format/fixed zerofill,
        /pgf/number format/precision=1
  },
  legend style={at={(1,1.1)}, anchor=south east, column sep=0.2em, font=\small},
  legend columns=4,
]
\addplot[mark=*,mark options={fill=red},
   only marks,
   point meta=explicit symbolic,
   error bars/y dir=both,
   error bars/y fixed=0.00131497782,
] table[meta=name,x=param,y expr={1 - \thisrow{accuracy} },]{\datatable};
\addplot[mark=square*,mark options={fill=green},
   nodes near coords, only marks,
   every node near coord/.append style={inner sep=4pt},
   point meta=explicit symbolic,
] table[meta=name,x=param,y expr={1 - \thisrow{accuracy} },]{\rdatatable};
\addplot[mark=triangle*,mark options={fill=blue},
   nodes near coords, nodes near coords align = {below}, only marks,
   every node near coord/.append style={inner sep=4pt},
   point meta=explicit symbolic,
] table[meta=name,x=param,y expr={1 - \thisrow{accuracy} },]{\tdatatable};
\addplot[mark=diamond*,mark options={fill=yellow},
   nodes near coords, nodes near coords align = {below}, only marks,
   every node near coord/.append style={inner sep=4pt},
   point meta=explicit symbolic,
] table[meta=name,x=param,y expr={1 - \thisrow{accuracy} },]{\cdatatable};
\legend{NiN, Root, Tree, Column}
\end{axis}
\end{tikzpicture}
\end{subfigure}
~
\begin{subfigure}[b]{\linewidth}
\pgfplotstableread[col sep=comma]{data/nincifar.csv}\datatable
\pgfplotstableread[col sep=comma]{data/nincifar_root_s.csv}\rdatatable
\pgfplotstableread[col sep=comma]{data/nincifar_tree_s.csv}\tdatatable
\pgfplotstableread[col sep=comma]{data/nincifar_col_s.csv}\cdatatable
\pgfplotsset{major grid style={dotted,red}}

\centering
\begin{tikzpicture}
\begin{axis}[
  width=\linewidth,
  height=0.66\linewidth,
  axis x line=bottom,
  ylabel=Error,
  xlabel=FLOPS (Multiply-Add),
  axis lines=left,
  enlarge x limits=0.05,
  enlarge y limits=0.05,
  grid=major,
  ytick={0.002,0.004,...,1.0},
  xticklabel style={
        /pgf/number format/fixed zerofill,
        /pgf/number format/precision=1
  },
  yticklabel={\pgfmathparse{\tick*1}\pgfmathprintnumber{\pgfmathresult}\%},style={
        /pgf/number format/fixed zerofill,
        /pgf/number format/precision=1
  },
]
\addplot[mark=*,mark options={fill=red},
   only marks,
   point meta=explicit symbolic,
   error bars/y dir=both,
   error bars/y fixed=0.00131497782,
] table[meta=name,x=ma,y expr={1 - \thisrow{accuracy} },]{\datatable};
\addplot[mark=square*,mark options={fill=green},
   nodes near coords, only marks,
   every node near coord/.append style={inner sep=4pt},
   point meta=explicit symbolic,
] table[meta=name,x=ma,y expr={1 - \thisrow{accuracy} },]{\rdatatable};
\addplot[mark=triangle*,mark options={fill=blue},
   nodes near coords, nodes near coords align = {below}, only marks,
   every node near coord/.append style={inner sep=4pt},
   point meta=explicit symbolic,
] table[meta=name,x=ma,y expr={1 - \thisrow{accuracy} },]{\tdatatable};
\addplot[mark=diamond*,mark options={fill=yellow},
   nodes near coords, nodes near coords align = {below}, only marks,
   every node near coord/.append style={inner sep=4pt},
   point meta=explicit symbolic,
] table[meta=name,x=ma,y expr={1 - \thisrow{accuracy} },]{\cdatatable};
\end{axis}
\end{tikzpicture}
\end{subfigure}

\caption{\textbf{Network-in-Network CIFAR10 Results.} Spatial filters (3$\times$3, 5$\times$5) are grouped hierarchically. The best models are closest to the origin. For the standard network, the mean and standard deviation (error bars) are shown over 5 different random initializations.
}
\label{fig:nincifarplotsconvonly}
\end{figure}

%% file: resnet50matable.tex
\begin{table}[tbp]
\caption[ResNet 50 ILSVRC Results]{\textbf{ResNet 50 Results.}}
\label{table:resnet50imagenetresults}
\centering
\pgfplotstableread[col sep=comma]{data/resnet50ma.csv}\data
\pgfplotstableread[col sep=comma]{data/resnet50maconvonly.csv}\codata
\pgfplotstablevertcat{\data}{\codata}
\pgfplotstableset{
    create on use/singlegpu/.style={
        create col/expr={\thisrow{GPU Forward} / \thisrow{Batch Size}}},
}
\pgfplotstableset{
    create on use/singlecpu/.style={
        create col/expr={\thisrow{CPU Forward} / \thisrow{Batch Size}}},
}
\pgfplotstabletypeset[
    every head row/.style={
    before row=\toprule,after row=\midrule},
    every last row/.style={
    after row=\bottomrule},
    every first row/.style={
    after row=\midrule}, 
    fixed zerofill,     
    columns={Full Name, Multiply-Acc., Param., Top-1 Acc., Top-5 Acc., singlecpu, singlegpu},
    columns/Full Name/.style={
        column name=Model,
        string type
    },
    columns/singlegpu/.style={
        column name=GPU (ms),
        precision=1
    },
    columns/singlecpu/.style={
        column name=CPU (ms),
        precision=0
    },
    columns/Multiply-Acc./.style={
        column name=FLOPS {\small $\times 10^{9}$},
        preproc/expr={{##1/1e9}}
    },
    columns/Param./.style={
        column name=Param. {\small $\times 10^{7}$},
        preproc/expr={{##1/1e7}}
    },
    columns/Top-1 Acc./.style={precision=3},
    columns/Top-5 Acc./.style={precision=3},
    highlight col max ={\data}{Top-1 Acc.},
    highlight col max ={\data}{Top-5 Acc.}, 
    highlight col min ={\data}{Param.}, 
    highlight col min ={\data}{Multiply-Acc.}, 
    column type/.add={@{}lp{3em}p{3em}p{3em}p{3em}p{3em}p{3em}@{}}{},
    col sep=comma]{\data}
\end{table}

%% file: resnet50maplot.tex
\begin{figure}[tbp]
\centering
\begin{subfigure}[b]{\linewidth}
\pgfplotstableread[col sep=comma]{data/resnet50ma.csv}\gdatatable
\pgfplotstableread[col sep=comma]{data/resnet50maconvonly.csv}\codatatable
\pgfplotsset{major grid style={dotted,red}}

\pgfplotstableset{
    create on use/singlecpu/.style={
        create col/expr={\thisrow{CPU Forward} / \thisrow{Batch Size}}},
}

\centering
\begin{tikzpicture}
\begin{axis}[
  width=\linewidth,
  height=0.45\linewidth,
  axis x line=bottom,
  ylabel=Top-5 Error,
  xlabel=Model Parameters (\# Floats),
  axis lines=left,
  enlarge x limits=0.05,
  grid=major,
  ytick={0.01,0.02,...,0.2},
  ymin=0.07,ymax=0.1,
  xticklabel style={
        /pgf/number format/fixed,
        /pgf/number format/precision=3
  },
  yticklabel={\pgfmathparse{\tick*100}\pgfmathprintnumber{\pgfmathresult}\%},style={
        /pgf/number format/fixed,
        /pgf/number format/precision=1
  },
  legend style={at={(0.98,0.98)}, anchor=north east, column sep=0.5em},
  legend columns=3,
]
\addplot[mark=*,mark options={fill=red},
   only marks,
   point meta=explicit symbolic,
] table[meta=Network,x=Param.,y expr={1 - \thisrow{Top-5 Acc.} },]{\gdatatable};
\addplot[mark=square*,mark options={fill=green},
   nodes near coords, nodes near coords align = {below}, only marks,
   every node near coord/.append style={inner sep=4pt},
   only marks,
   point meta=explicit symbolic,
] table[meta=Network,x=Param.,y expr={1 - \thisrow{Top-5 Acc.} },]{\codatatable};
\legend{ResNet 50, All Filters, Spatial Filters, LDE Half}
\end{axis}
\end{tikzpicture}
\caption{\textbf{Model Parameters \vs Top-5 Error.}}
\label{fig:resnet5050param}
\end{subfigure}
~
\begin{subfigure}[b]{\linewidth}
\pgfplotstableread[col sep=comma]{data/resnet50ma.csv}\gdatatable
\pgfplotstableread[col sep=comma]{data/resnet50maconvonly.csv}\codatatable
\pgfplotsset{major grid style={dotted,red}}

\centering
\begin{tikzpicture}
\begin{axis}[
  width=\linewidth,
  height=0.45\linewidth,
  axis x line=bottom,
  ylabel=Top-5 Error,
  xlabel=FLOPS (Multiply-Add),
  axis lines=left,
  enlarge x limits=0.05,
  grid=major,
  ytick={0.01,0.02,...,0.2},
  ymin=0.07,ymax=0.1,
  xticklabel style={
        /pgf/number format/fixed,
        /pgf/number format/precision=3
  },
  yticklabel={\pgfmathparse{\tick*100}\pgfmathprintnumber{\pgfmathresult}\%},style={
        /pgf/number format/fixed,
        /pgf/number format/precision=1
  },
  legend style={at={(0.98,0.98)}, anchor=north east, column sep=0.5em},
  legend columns=3,
]
\addplot[mark=*,mark options={fill=red},
   only marks,
   point meta=explicit symbolic,
] table[meta=Network,x=Multiply-Acc.,y expr={1 - \thisrow{Top-5 Acc.} },]{\gdatatable};
\addplot[mark=square*,mark options={fill=green},
   nodes near coords, nodes near coords align = {below}, only marks,
   every node near coord/.append style={inner sep=4pt},
   only marks,
   point meta=explicit symbolic,
] table[meta=Network,x=Multiply-Acc.,y expr={1 - \thisrow{Top-5 Acc.} },]{\codatatable};
\end{axis}
\end{tikzpicture}
\caption{\textbf{FLOPS (Multiply-Add) \vs Top-5 Error.}}
\label{fig:resnet50ma}
\end{subfigure}
~
\begin{subfigure}[b]{\linewidth}
\pgfplotstableread[col sep=comma]{data/resnet50ma.csv}\gdatatable
\pgfplotstableread[col sep=comma]{data/resnet50maconvonly.csv}\codatatable
\pgfplotsset{major grid style={dotted,red}}

\centering
\begin{tikzpicture}
\begin{axis}[
  width=\linewidth,
  height=0.45\linewidth,
  axis x line=bottom,
  ylabel=Top-5 Error,
  xlabel=GPU Forward (ms),
  axis lines=left,
  enlarge x limits=0.05,
  grid=major,
  ytick={0.01,0.02,...,0.2},
  ymin=0.07,ymax=0.1,
  xticklabel style={
        /pgf/number format/fixed,
        /pgf/number format/precision=3
  },
  yticklabel={\pgfmathparse{\tick*100}\pgfmathprintnumber{\pgfmathresult}\%},style={
        /pgf/number format/fixed,
        /pgf/number format/precision=1
  },
  legend style={at={(0.98,0.98)}, anchor=north east, column sep=0.5em},
  legend columns=3,
]
\addplot[mark=*,mark options={fill=red},
   only marks,
   point meta=explicit symbolic,
] table[meta=Network,
    x expr={\thisrow{GPU Forward} / \thisrow{Batch Size}},
    y expr={1 - \thisrow{Top-5 Acc.} }
]{\gdatatable};
\addplot[mark=square*,mark options={fill=green},
   nodes near coords, nodes near coords align = {below}, only marks,
   every node near coord/.append style={inner sep=4pt},
   only marks,
   point meta=explicit symbolic,
] table[meta=Network,
    x expr={\thisrow{GPU Forward} / \thisrow{Batch Size}},
    y expr={1 - \thisrow{Top-5 Acc.} },
]{\codatatable};
\end{axis}
\end{tikzpicture}
\caption{\textbf{GPU Forward Time \vs Top-5 Error.}}
\label{fig:resnet5050gpuforward}
\end{subfigure}
~
\begin{subfigure}[b]{\linewidth}
\pgfplotstableread[col sep=comma]{data/resnet50ma.csv}\gdatatable
\pgfplotstableread[col sep=comma]{data/resnet50maconvonly.csv}\codatatable
\pgfplotsset{major grid style={dotted,red}}

\centering
\begin{tikzpicture}
\begin{axis}[
  width=\linewidth,
  height=0.45\linewidth,
  axis x line=bottom,
  ylabel=Top-5 Error,
  xlabel=CPU Forward (ms),
  axis lines=left,
  enlarge x limits=0.05,
  grid=major,
  ytick={0.01,0.02,...,0.2},
  ymin=0.07,ymax=0.1,
  xticklabel style={
        /pgf/number format/fixed,
        /pgf/number format/precision=3
  },
  yticklabel={\pgfmathparse{\tick*100}\pgfmathprintnumber{\pgfmathresult}\%},style={
        /pgf/number format/fixed,
        /pgf/number format/precision=1
  },
  legend style={at={(0.98,0.98)}, anchor=north east, column sep=0.5em},
  legend columns=3,
]
\addplot[mark=*,mark options={fill=red},
   only marks,
   point meta=explicit symbolic,
] table[meta=Network,
    x expr={\thisrow{CPU Forward} / \thisrow{Batch Size}},
    y expr={1 - \thisrow{Top-5 Acc.} },
]{\gdatatable};
\addplot[mark=square*,mark options={fill=green},
   nodes near coords, nodes near coords align = {below}, only marks,
   every node near coord/.append style={inner sep=4pt},
   only marks,
   point meta=explicit symbolic,
] table[meta=Network,
    x expr={\thisrow{CPU Forward} / \thisrow{Batch Size}},
    y expr={1 - \thisrow{Top-5 Acc.} },
]{\codatatable};
\end{axis}
\end{tikzpicture}
\caption{\textbf{CPU Forward Time \vs Top-5 Error.}}
\label{fig:resnet5050cpuforward}
\end{subfigure}

\caption{\textbf{ResNet-50 Results.} Models with filter groups have fewer parameters, and less floating point operations, while maintaining error comparable to the baseline.}
\label{fig:resnet50plots}
\end{figure}

%% file: resnet200matable.tex
\begin{table}[tb]
\caption[ResNet 200 ILSVRC Results]{\textbf{ResNet-200 Results}}
\label{table:resnet200imagenetresults}
\centering
\pgfplotstableread[col sep=comma]{data/resnet200.csv}\data
\pgfplotstabletypeset[
    every head row/.style={
    before row=\toprule,after row=\midrule},
    every last row/.style={
    after row=\bottomrule},
    every first row/.style={
    after row=\midrule}, 
    columns={full name, ma, param, top1, top5},
    columns/full name/.style={
        column name=Model,
        string type
    },
    columns/ma/.style={
        column name=FLOPS~{\small$\times 10^{12}$},
        fixed zerofill,
        preproc/expr={{##1/1e12}},
    },
    columns/param/.style={
        column name=Param.~{\small$\times 10^{7}$},
        fixed zerofill,
        preproc/expr={{##1/1e7}},
    },
    columns/top1/.style={
        precision=4,
        column name=Top-1 Err.,
        fixed zerofill,
    },
    columns/top5/.style={
        precision=4,
        column name=Top-5 Err.,
        fixed zerofill,
    },
    column type/.add={@{}lrrrrrr@{}}{},
    highlight col min ={\data}{top1},
    highlight col min ={\data}{top5}, 
    highlight col min ={\data}{param}, 
    highlight col min ={\data}{ma}, 
    col sep=comma]{\data}
\end{table}

%% file: googlenetmatable.tex
\begin{table}[tbp]
\caption[GoogLeNet ILSVRC Results]{\textbf{GoogLeNet Results.}}
\label{table:googlenetimagenetresults}
\centering
\pgfplotstableread[col sep=comma]{data/googlenetma.csv}\data
\pgfplotstableread[col sep=comma]{data/googlenetmaconvonly.csv}\codata
\pgfplotstablevertcat{\data}{\codata}
\pgfplotstableset{
    create on use/singlegpu/.style={
        create col/expr={\thisrow{GPU Forward} / \thisrow{Batch Size}}},
}
\pgfplotstableset{
    create on use/singlecpu/.style={
        create col/expr={\thisrow{CPU Forward} / \thisrow{Batch Size}}},
}
\pgfplotstabletypeset[
    every head row/.style={
    before row=\toprule,after row=\midrule},
    every last row/.style={
    after row=\bottomrule},
    every first row/.style={
    after row=\midrule}, 
    fixed zerofill,     
    columns={Full Name, Multiply-Acc., Param., Top-1 Acc., Top-5 Acc., singlecpu, singlegpu},
    columns/Full Name/.style={
        column name=Model,
        string type
    },
    columns/singlegpu/.style={
        column name=GPU (ms),
        precision=2
    },
    columns/singlecpu/.style={
        column name=CPU (ms),
        precision=0
    },
    columns/Multiply-Acc./.style={
        column name=FLOPS {\small $\times 10^{9}$},
        preproc/expr={{##1/1e9}}
    },
    columns/Param./.style={
        column name=Param. {\small $\times 10^{7}$},
        preproc/expr={{##1/1e7}}
    },
    columns/Top-1 Acc./.style={precision=3},
    columns/Top-5 Acc./.style={precision=3},
    highlight col max ={\data}{Top-1 Acc.},
    highlight col max ={\data}{Top-5 Acc.}, 
    highlight col min ={\data}{Param.}, 
    highlight col min ={\data}{Multiply-Acc.}, 
    column type/.add={@{}lp{3em}p{3em}p{3em}p{3em}p{3em}p{3em}@{}}{},
    col sep=comma]{\data}
\end{table}

%% file: googlenetmaplot.tex
\begin{figure}[tbp]
\centering
\begin{subfigure}[b]{\linewidth}
\pgfplotstableread[col sep=comma]{data/googlenetma.csv}\gdatatable
\pgfplotstableread[col sep=comma]{data/googlenetmaconvonly.csv}\codatatable
\pgfplotsset{major grid style={dotted,red}}

\centering
\begin{tikzpicture}
\begin{axis}[
  width=\linewidth,
  height=0.4\linewidth,
  axis x line=bottom,
  ylabel=Top-5 Error,
  xlabel=Model Parameters (\# Floats),
  axis lines=left,
  enlarge x limits=0.05,
  grid=major,
  ytick={0.01,0.02,...,0.2},
  ymin=0.095,ymax=0.125,
  xticklabel style={
        /pgf/number format/fixed,
        /pgf/number format/precision=3
  },
  yticklabel={\pgfmathparse{\tick*100}\pgfmathprintnumber{\pgfmathresult}\%},style={
        /pgf/number format/fixed,
        /pgf/number format/precision=1
  },
  legend style={at={(0.98,0.98)}, anchor=north east, column sep=0.5em},
  legend columns=3,
]
\addplot[mark=*,mark options={fill=red},
   only marks,
   point meta=explicit symbolic,
] table[meta=Network,x=Param.,y expr={1 - \thisrow{Top-5 Acc.} },]{\gdatatable};
\addplot[mark=square*,mark options={fill=green},
   nodes near coords, nodes near coords align = {below}, only marks,
   every node near coord/.append style={inner sep=4pt},
   only marks,
   point meta=explicit symbolic,
] table[meta=Network,x=Param.,y expr={1 - \thisrow{Top-5 Acc.} },]{\codatatable};
\legend{GoogLeNet, All Filters, Spatial Filters}
\end{axis}
\end{tikzpicture}
\caption{\textbf{Model Parameters \vs Top-5 Error.}}
\label{fig:googlenet50param}
\end{subfigure}
~
\begin{subfigure}[b]{\linewidth}
\pgfplotstableread[col sep=comma]{data/googlenetma.csv}\gdatatable
\pgfplotstableread[col sep=comma]{data/googlenetmaconvonly.csv}\codatatable
\pgfplotsset{major grid style={dotted,red}}

\centering
\begin{tikzpicture}
\begin{axis}[
  width=\linewidth,
  height=0.4\linewidth,
  axis x line=bottom,
  ylabel=Top-5 Error,
  xlabel=FLOPS (Multiply-Add),
  axis lines=left,
  enlarge x limits=0.05,
  grid=major,
  ytick={0.01,0.02,...,0.2},
  ymin=0.095,ymax=0.125,
  xticklabel style={
        /pgf/number format/fixed,
        /pgf/number format/precision=3
  },
  yticklabel={\pgfmathparse{\tick*100}\pgfmathprintnumber{\pgfmathresult}\%},style={
        /pgf/number format/fixed,
        /pgf/number format/precision=1
  },
  legend style={at={(0.98,0.98)}, anchor=north east, column sep=0.5em},
  legend columns=3,
]
\addplot[mark=*,mark options={fill=red},
   only marks,
   point meta=explicit symbolic,
] table[meta=Network,x=Multiply-Acc.,y expr={1 - \thisrow{Top-5 Acc.} },]{\gdatatable};
\addplot[mark=square*,mark options={fill=green},
   nodes near coords, nodes near coords align = {below}, only marks,
   every node near coord/.append style={inner sep=4pt},
   only marks,
   point meta=explicit symbolic,
] table[meta=Network,x=Multiply-Acc.,y expr={1 - \thisrow{Top-5 Acc.} },]{\codatatable};
\end{axis}
\end{tikzpicture}
\caption{\textbf{FLOPS (Multiply-Add) \vs Top-5 Error.}}
\label{fig:googlenet50ma}
\end{subfigure}
~
\begin{subfigure}[b]{\linewidth}
\pgfplotstableread[col sep=comma]{data/googlenetma.csv}\gdatatable
\pgfplotstableread[col sep=comma]{data/googlenetmaconvonly.csv}\codatatable
\pgfplotsset{major grid style={dotted,red}}

\centering
\begin{tikzpicture}
\begin{axis}[
  width=\linewidth,
  height=0.4\linewidth,
  axis x line=bottom,
  ylabel=Top-5 Error,
  xlabel=GPU Forward (ms),
  axis lines=left,
  enlarge x limits=0.05,
  grid=major,
  ytick={0.01,0.02,...,0.2},
  ymin=0.095,ymax=0.125,
  xticklabel style={
        /pgf/number format/fixed,
        /pgf/number format/precision=3
  },
  yticklabel={\pgfmathparse{\tick*100}\pgfmathprintnumber{\pgfmathresult}\%},style={
        /pgf/number format/fixed,
        /pgf/number format/precision=1
  },
  legend style={at={(0.98,0.98)}, anchor=north east, column sep=0.5em},
  legend columns=3,
]
\addplot[mark=*,mark options={fill=red},
   only marks,
   point meta=explicit symbolic,
] table[meta=Network,
    x expr={\thisrow{GPU Forward} / \thisrow{Batch Size}},
    y expr={1 - \thisrow{Top-5 Acc.} },]{\gdatatable};
\addplot[mark=square*,mark options={fill=green},
   nodes near coords, nodes near coords align = {below}, only marks,
   every node near coord/.append style={inner sep=4pt},
   only marks,
   point meta=explicit symbolic,
] table[meta=Network,
    x expr={\thisrow{GPU Forward} / \thisrow{Batch Size}},
    y expr={1 - \thisrow{Top-5 Acc.} },]{\codatatable};
\end{axis}
\end{tikzpicture}
\caption{\textbf{GPU Forward Time \vs Top-5 Error.}}
\label{fig:googlenet50gpuforward}
\end{subfigure}
~
\begin{subfigure}[b]{\linewidth}
\pgfplotstableread[col sep=comma]{data/googlenetma.csv}\gdatatable
\pgfplotstableread[col sep=comma]{data/googlenetmaconvonly.csv}\codatatable
\pgfplotsset{major grid style={dotted,red}}

\centering
\begin{tikzpicture}
\begin{axis}[
  width=\linewidth,
  height=0.4\linewidth,
  axis x line=bottom,
  ylabel=Top-5 Error,
  xlabel=CPU Forward (ms),
  axis lines=left,
  enlarge x limits=0.05,
  grid=major,
  ytick={0.01,0.02,...,0.2},
  ymin=0.095,ymax=0.125,
  xticklabel style={
        /pgf/number format/fixed,
        /pgf/number format/precision=3
  },
  yticklabel={\pgfmathparse{\tick*100}\pgfmathprintnumber{\pgfmathresult}\%},style={
        /pgf/number format/fixed,
        /pgf/number format/precision=1
  },
  legend style={at={(0.98,0.98)}, anchor=north east, column sep=0.5em},
  legend columns=3,
]
\addplot[mark=*,mark options={fill=red},
   only marks,
   point meta=explicit symbolic,
] table[meta=Network,
    x expr={\thisrow{CPU Forward} / \thisrow{Batch Size}},
    y expr={1 - \thisrow{Top-5 Acc.} },]{\gdatatable};
\addplot[mark=square*,mark options={fill=green},
   nodes near coords, nodes near coords align = {below}, only marks,
   every node near coord/.append style={inner sep=4pt},
   only marks,
   point meta=explicit symbolic,
] table[meta=Network,
    x expr={\thisrow{CPU Forward} / \thisrow{Batch Size}},
    y expr={1 - \thisrow{Top-5 Acc.} },]{\codatatable};
\end{axis}
\end{tikzpicture}
\caption{\textbf{CPU Forward Time \vs Top-5 Error.}}
\label{fig:googlenet50cpuforward}
\end{subfigure}

\caption{\textbf{GoogLeNet Results.} Models with filter groups have fewer parameters, and less floating point operations, while maintaining error comparable to the baseline.}
\label{fig:googlenet50plots}
\end{figure}